\documentclass[runningheads]{llncs}

\usepackage{eccv}

\usepackage{eccvabbrv}

\usepackage{graphicx}
\usepackage{booktabs}

\usepackage[accsupp]{axessibility}  

\usepackage{array, colortbl}
\usepackage{arydshln}

\usepackage[pagebackref,breaklinks,colorlinks,citecolor=eccvblue]{hyperref}

\usepackage{orcidlink}

\usepackage{microtype}

\usepackage[utf8]{inputenc} 
\usepackage[T1]{fontenc}    
\usepackage{hyperref}       
\usepackage{url}            
\usepackage{amsfonts}       
\usepackage{nicefrac}       
\usepackage{xcolor}         
\usepackage{wrapfig}
\usepackage{algorithm}
\usepackage{algorithmic}
\usepackage{bm}
\usepackage{tikz}
\usepackage{pdfpages}
\usepackage{fontawesome}
\usepackage{amsmath}
\usepackage{amssymb}
\usepackage{bbm}
\usepackage{pifont}
\usepackage{fancybox}
\usepackage{tikz}
\usepackage{enumitem}
\usepackage{color, colortbl}
\usepackage{dsfont}
\usepackage{multirow} 
\usepackage[referable]{threeparttablex}

\usetikzlibrary{positioning}
\usetikzlibrary{arrows}
\usetikzlibrary{calc}
\usetikzlibrary{fit, shadows, shapes, backgrounds}
\usetikzlibrary{bayesnet}
\usepackage{url}
\usepackage{xr}

\usetikzlibrary{shapes,decorations,arrows,calc,arrows.meta,fit,positioning}
\tikzset{
    -Latex,auto,node distance =1 cm and 1 cm,semithick,
    state/.style ={ellipse, draw, minimum width = 0.7 cm},
    point/.style = {circle, draw, inner sep=0.04cm,fill,node contents={}},
    bidirected/.style={-Latex,dashed},
    el/.style = {inner sep=2pt, align=left, sloped},
   box/.style={rectangle,draw,node distance=1cm,text width=15em,text centered,rounded corners,minimum height=2em,thick},
    arrow/.style={draw,-latex',thick,dashed},
}
\definecolor{Gray}{gray}{0.9}
\definecolor{cvprblue}{rgb}{0.21,0.49,0.74}
\newcommand{\gr}{\color{green!50!black}}

\definecolor{newcolor}{rgb}{.8,.349,.1}

\renewlist{tablenotes}{enumerate}{1}
\makeatletter
\setlist[tablenotes]{label=\tnote{\alph*},ref=\alph*,itemsep=\z@,topsep=\z@skip,partopsep=\z@skip,parsep=\z@,itemindent=\z@,labelindent=\tabcolsep,labelsep=.1em,leftmargin=*,align=left,before={\scriptsize}}
\makeatother

\newcommand{\colortext}{\color{black}}

\begin{document}

\title{SelEx: Self-Expertise in Fine-Grained Generalized Category Discovery} 

\titlerunning{SelEx: Self-Expertise in Fine-Grained Generalized Category Discovery}

\author{Sarah Rastegar\inst{1}\orcidlink{0000-0002-4542-7388} \and
Mohammadreza Salehi\inst{1}\orcidlink{0000-0002-9247-9439} \and
Yuki M. Asano\inst{1}\orcidlink{0000-0002-8533-4020}\and
Hazel~Doughty\inst{2}\orcidlink{0000-0002-3670-3897}\and
Cees G. M. Snoek\inst{1}\orcidlink{0000-0001-9092-1556}}

\authorrunning{S.~Rastegar et al.}

\institute{University of Amsterdam \and
Leiden University
}
\maketitle

\vspace{-0.5em}
\begin{abstract}
\vspace{-0.5em}
In this paper, we address Generalized Category Discovery, aiming to simultaneously uncover novel categories and accurately classify known ones. Traditional methods, which lean heavily on self-supervision and contrastive learning, often fall short when distinguishing between fine-grained categories. To address this, we introduce a novel concept called `self-expertise', which enhances the model's ability to recognize subtle differences and uncover unknown categories. Our approach combines unsupervised and supervised self-expertise strategies to refine the model's discernment and generalization. Initially, hierarchical pseudo-labeling is used to provide `soft supervision', improving the effectiveness of self-expertise. Our supervised technique differs from traditional methods by utilizing more abstract positive and negative samples, aiding in the formation of clusters that can generalize to novel categories. Meanwhile, our unsupervised strategy encourages the model to sharpen its category distinctions by considering within-category examples as `hard' negatives. Supported by theoretical insights, our empirical results showcase that our method outperforms existing state-of-the-art techniques in Generalized Category Discovery across several fine-grained datasets. {\colortext Our code is available at: }\url{https://github.com/SarahRastegar/SelEx}.
\vspace{-0.5em}
\end{abstract}


\vspace{-3em}
\section{Introduction}
\vspace{-0.5em}

\begin{figure*}[t!]
\vspace{-0.75em}
\begin{tabular}{>{\arrayrulecolor{gray}}c:c}
\arrayrulecolor{gray}
\toprule
\textbf{Traditional Methods}&\textbf{This Paper}\\
\midrule
\begin{subfigure}{.5\textwidth}
  \centering
  \includegraphics[height=0.45\linewidth]{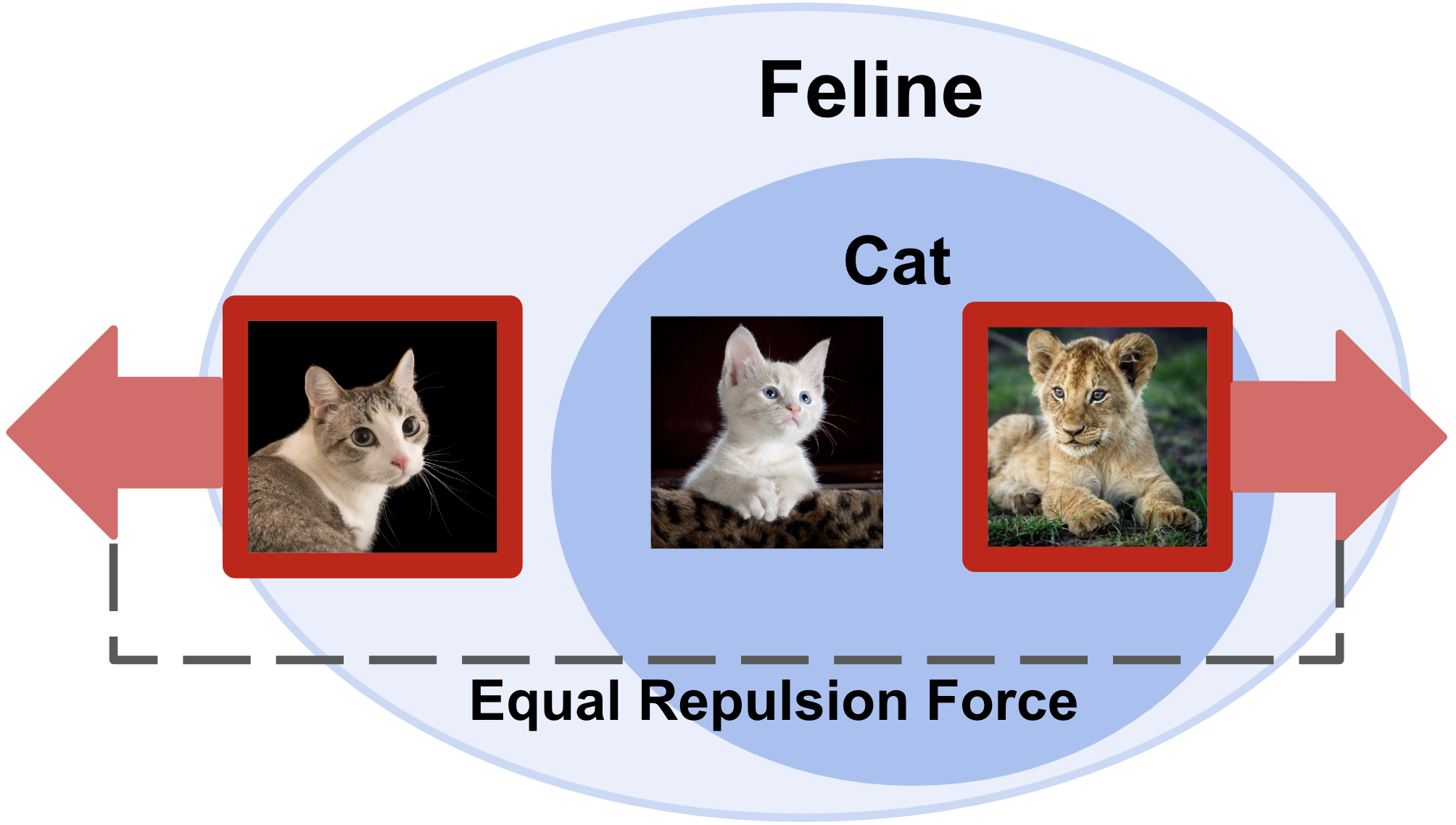}
  \caption{Unsupervised Contrastive Learning}
  \label{fig:figa}
\end{subfigure}
&\begin{subfigure}{.5\textwidth}
  \centering
  \includegraphics[height=0.45\linewidth]{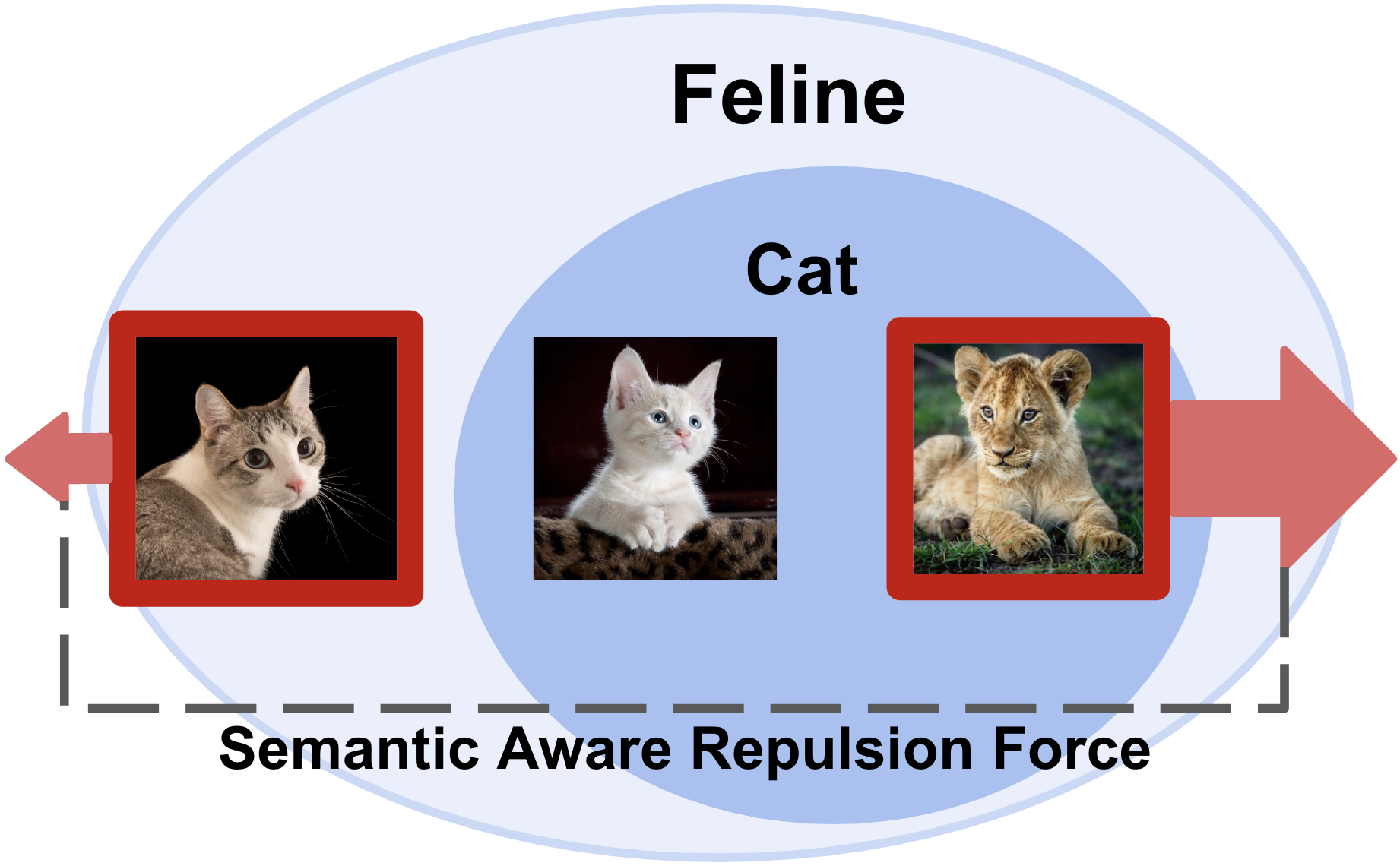}
  \caption{Unsupervised Self-Expertise}
  \label{fig:figb}
\end{subfigure}\\
\begin{subfigure}{.5\textwidth}
  \centering
  \includegraphics[height=0.45\linewidth]{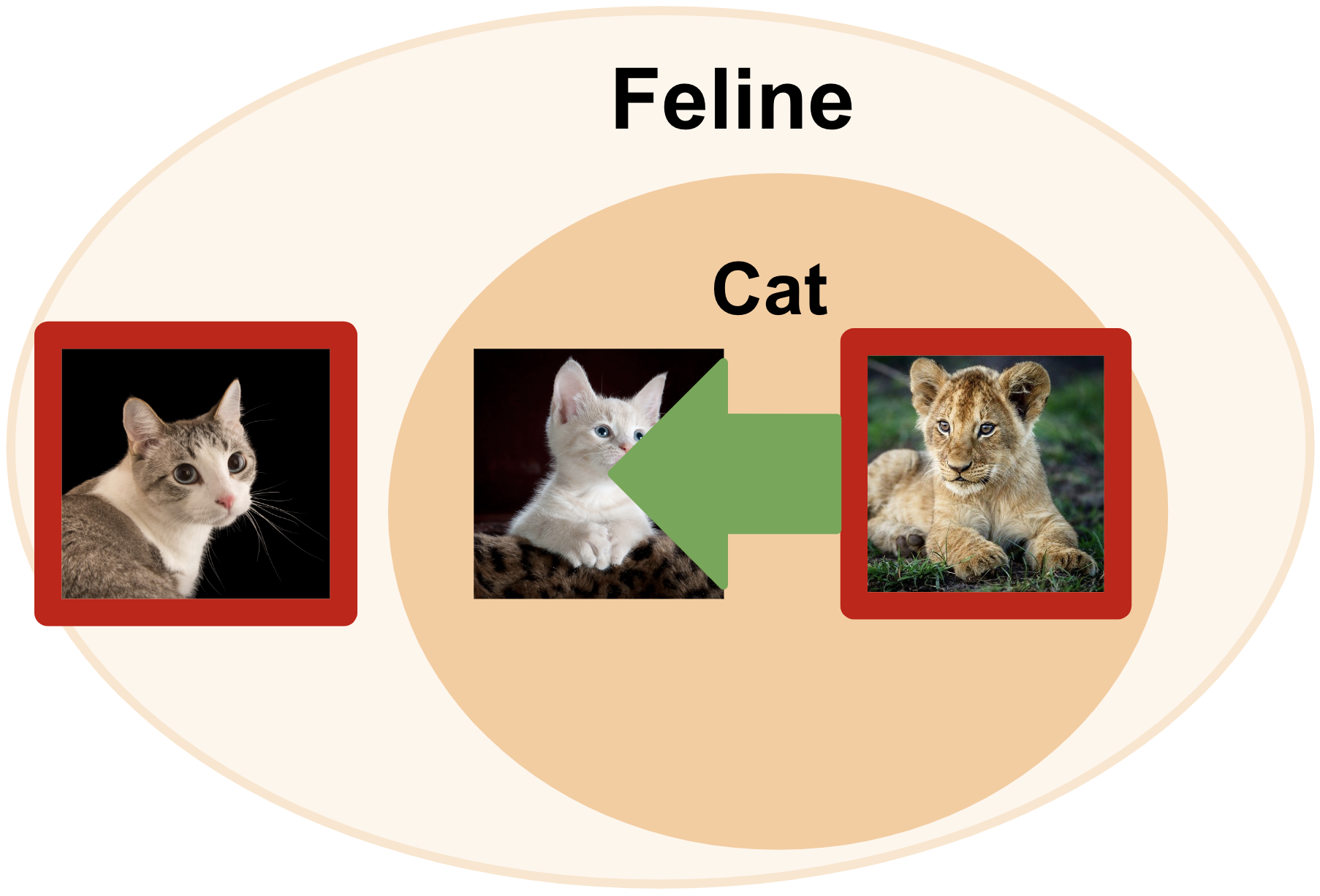}
  \caption{Supervised Contrastive Learning}
  \label{fig:figc}
\end{subfigure}
&\begin{subfigure}{.5\textwidth}
  \centering
  \includegraphics[height=0.45\linewidth]{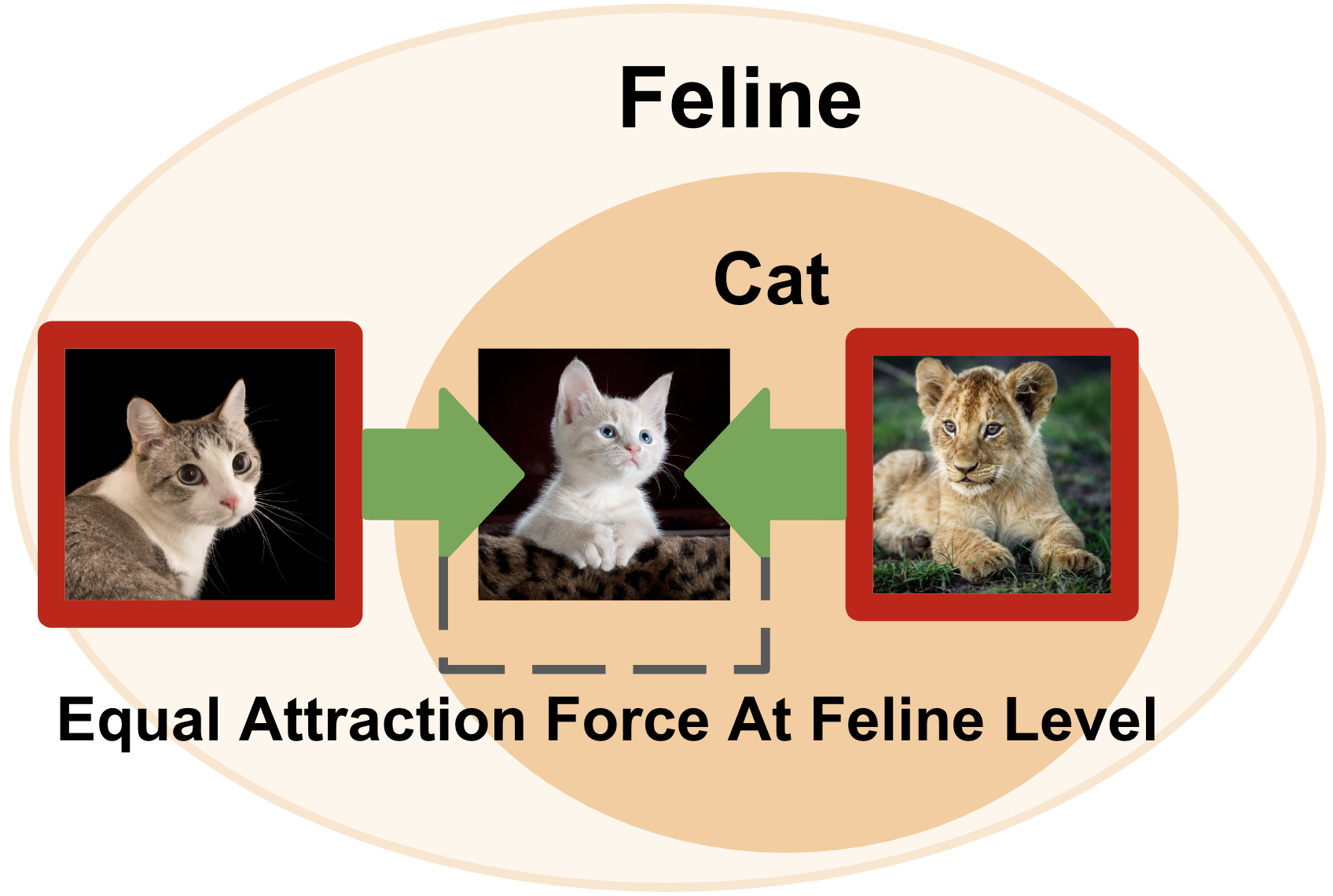}
  \caption{Supervised Self-Expertise}
  \label{fig:figd}
\end{subfigure}\\
\bottomrule
\end{tabular}
\vspace{-1em}
\caption{\textbf{The motivation for self-expertise} \textit{(a) Unsupervised Contrastive Learning.} shows identical repulsion for both misclassified cat and lion. \textit{(b)~Unsupervised Self-Expertise.}  
focuses on distinguishing hard negatives within a category cluster, applying less repulsion to external members, and varying the repulsion for misclassified samples, resulting in milder repulsion for the mislabeled cat and stronger for the lion. \textit{(c)~Supervised Contrastive Learning.} focuses on attracting similar category members, leaving others unaffected. \textit{(d) Supervised Self-Expertise.} graduates the attraction of samples based on semantic similarity. Both a misclassified cat and lion are equally attracted to the cat sample at the feline level. Supervised and unsupervised self-expertise together enhance accuracy by attracting similar samples and repelling dissimilar ones.}

\vspace{-2.7em}
\label{fig:fig}
\end{figure*}

Supervised learning has proven its effectiveness in classifying predefined image categories \cite{he2016deep, krizhevsky2017imagenet, simonyan2014very, huang2017densely, szegedy2015going}. However, it struggles significantly when presented with unknown categories, hindering its real-world applicability \cite{boult2019learning,9723693,salehi2022a,zhu2024open,troisemaine2023novel}. 
 Generalized Category Discovery (GCD) addresses this limitation by automatically identifying both known and novel categories from unlabeled data \cite{cao2021open,vaze2022generalized, zhang2022promptcal, hao2023cipr, chiaroni2023parametric, wen2022simple, an2023generalized,rastegar2023learn,vaze2023clevr4}.
A key approach for handling unknown categories within GCD has been self-supervision through contrastive learning \cite{jaiswal2020survey,zhai2019s4l,liu2021self,chen2020simple, li2021prototypical, noroozi2016unsupervised,caron2018deep, cao2021open, han2020automatically}. However, this method struggles with fragmented clustering and an increased false negative rate, particularly in fine-grained categorization where positive augmented samples may significantly differ from their negative counterparts from the same category, which leads to misclassification \cite{cole2022does,caron2018deep, khorasgani2022slic, huynh2022boosting}. 
Although supervised contrastive learning \cite{khosla2020supervised, vaze2022generalized} improves discrimination among known categories, it struggles with unknown categories due to the absence of supervisory signals. Our work navigates this essential trade-off, aiming to merge the discovery of unknown categories with fine-grained classification through self-supervision.

In this paper, we present a novel Generalized Category Discovery approach that combines contrastive learning with pseudo-labeling to uncover novel categories by enhancing self-expertise. We define `expertise' as the skill to generalize across different abstraction levels, much like an ornithologist distinguishes between species at various levels, a capability that extends beyond ordinary observation. Our method focuses on honing the ability to identify subtle distinctions and achieve broad generalizations. In \cref{fig:fig}, we illustrate how our model improves the detection of fine details and generalization through unsupervised and supervised self-expertise, respectively.  We make four contributions:
\begin{itemize}[noitemsep,nolistsep,leftmargin=*]
\item[$\circ$] We present a hierarchical semi-supervised k-means clustering approach that better initializes unknown clusters using known centers and addresses cluster sparsity by balancing distributions through a stable matching algorithm.
\item[$\circ$] We propose an unsupervised self-expertise approach that emphasizes hard negative samples with identical labels at each hierarchical level. 
\item [$\circ$]We present supervised self-expertise, which utilizes abstract pseudo-labels to generate weaker positive and stronger negative instances, facilitating rapid initial category clustering and enhancing generalization to novel categories.
\item [$\circ$]Empirically and theoretically, we demonstrate that our approach facilitates effective generalized category discovery with fine-grained abilities.
\end{itemize}

\vspace{-0.3em}
\section{Related Works}
\vspace{-0.8em}


\textbf{Generalized Category Discovery}
was introduced concurrently by Vaze \etal~\cite{vaze2022generalized} and Cao \etal \cite{cao2021open}. It provides models with unlabeled data from both novel and known categories, placing it within the realm of semi-supervised learning \cite{ouali2020overview, yang2022survey, rebuffi2020semi, oliver2018realistic, chapelle2009semi}. The unique challenge in generalized category discovery is handling categories without any labeled instances alongside already seen categories. There are primarily two approaches to address this challenge. One employs a series of prototypes as reference points, \eg, \cite{hao2023cipr, chiaroni2023parametric, wen2022simple, an2023generalized,choi2024contrastive,Xiao_2024_CVPR,wang2024beyond,zhang2022automatically,tan2024revisiting,yang2023bootstrap,kim2023proxy}. 
The second leverages local similarity as weak pseudo-labels per sample by utilizing sample similarities to form local clusters
\cite{pu2023dynamic,zhang2022promptcal,rastegar2023learn, gao2023opengcd,zhao2023learning,du2023fly} or by utilizing mean-teacher networks to address the challenges posed by noisy pseudo-labels~\cite{vaze2023clevr4,zhang2022promptcal, wen2022simple,wang2024sptnet}. Nonetheless, the foundation of these approaches is contrastive learning, which has previously been shown to falter in fine-grained classification \cite{cole2022does} due to strong augmentations in positives in comparison to nuanced visual differences between samples of the same category. 
To alleviate this, we introduce `self-expertise' aimed at hierarchical learning of known and unknown categories. 
Our method is particularly effective in overcoming the limited availability of positive samples per category and enhancing the identification of subtle differences among negative samples, which we deem the biggest challenge in fine-grained classification. 

\noindent\textbf{Hierarchical Representation Learning.}
Different approaches benefit from hierarchical categories. Zhang \etal~\cite{zhang2022use} use multiple label levels to enhance their representation through hierarchical contrastive learning. Guo \etal~\cite{guo2022hcsc} extract pseudo labels for hierarchical contrastive learning, where signals are positive within the same cluster. We also use hierarchical pseudo-labels, but instead employ negative samples from the same cluster for generalized category discovery. 

Otholt \etal~\cite{otholt2024guided} and Banerjee \etal~\cite{banerjee2024amend} proposed hierarchical approaches to address generalized category discovery. These works leverage neighborhood structures to delineate refined categories. Rastegar \etal~\cite{rastegar2023learn} learn an implicit category tree, facilitating hierarchical self-coding of categories, which maintains category similarity across all hierarchy levels. Differing from these works, our method leverages weak supervision from samples within each level of the hierarchy, which reduces misclassification impact on lower levels. Additionally, our focus on hard negatives for unsupervised self-expertise enhances the model's ability to discern nuanced distinctions, leading to better fine-grained classification. 
\vspace{-0.5em}
\section{Theoretical Framework for Self-Expertise}
\vspace{-0.8em}
\label{sec:theory}

\textbf{Notations.}
We denote the number of total categories with $K$ and the number of samples by $N$. For each random variable $c$, we indicate the number of associated samples by $|c|$. We use `$\ln$' for the natural logarithm and `$\lg$' for $\log_2$.
\noindent\textbf{Problem Definition.}
The challenge of generalized category discovery lies in classifying samples during inference as belonging to categories encountered during training or as entirely novel categories. To describe this formally, throughout the training phase, we have access to the input {\small$\mathcal{X}_\mathcal{S}$} for labeled and {\small$\mathcal{X}_\mathcal{U}$} for unlabelled data. However, our access to the labels is limited to {\small$\mathcal{Y}_\mathcal{S}$}, which represents the known categories. Our objective is to categorize unlabeled samples. 

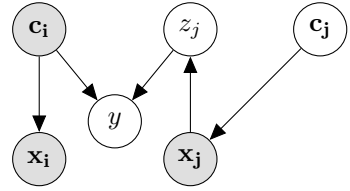
\begin{wrapfigure}{r}{0.4\textwidth}
  \centering
  \vspace{-1em}
\begin{tikzpicture}
  \node[latent, xshift=0cm](y) {$y$};
    \node[obs, above=of y, xshift=-1cm, yshift=-0.5cm]  (ci) {$\mathbf{c_i}$};
\node[latent, above=of y, xshift=1cm, yshift=-0.5cm]  (zj) {$z_j$};
  \node[obs, below=of zj, xshift=0cm,yshift=0cm] (xj) {$\mathbf{x_j}$};
  \node[obs, below=of ci, xshift=0cm,yshift=0cm]  (xi) {$\mathbf{x_i}$};
   \node[latent, right=of zj, xshift=-0cm]                               (cj) {$\mathbf{c_j}$};

  \edge {ci,zj} {y} ; %

        \edge {ci} {xi} ; %
            \edge {xj} {zj} ; %
            \edge {cj} {xj} ; %

\end{tikzpicture}
\vspace{-0.8em}
\caption{\textbf{Bayesian Network for the Generalized Category Discovery.} Shaded nodes are observed variables $\mathbf{x_i}$, $\mathbf{x_j}$ corresponds to images $i$ and $j$, and $\mathbf{c_i}$ and $\mathbf{c_j}$ which are the ground-truth category variable. $z_j$ is the latent category variable extracted from the model.}
\label{fig:baysmain}
\vspace{-2em}
\end{wrapfigure}
The possible labels for unlabeled input consist of both known and novel categories denoted as {\small$\mathcal{Y}_\mathcal{U}$}, where {\small$\mathcal{Y}_\mathcal{S}{\subset} \mathcal{Y}_\mathcal{U}$}.
We formulate the generalized category discovery problem as the Bayesian network in \cref{fig:baysmain}. In this network, $\mathbf{x_i}$ and $\mathbf{x_j}$ represent distinct samples or alternative perspectives of the same sample. The associated ground truth category variables, $\mathbf{c_i}$ and $\mathbf{c_j}$, are the focus of our information extraction process. 
In this Bayesian Network, we assume these two variables determine the distribution of $\mathbf{x_i}$ and $\mathbf{x_j}$. Here $z_j$ indicates the latent representation of the model for category variables, which is derived from $\mathbf{x_j}$. Contrastive training aims to estimate the equal distribution of the ground-truth category random variables accurately. Thus, for any pair of samples $i$ and $j$, our target is to approximate the following distribution closely:
\vspace{-0.5em}
\begin{equation}
    \label{eq:identymain}
    p(y{=}1|\mathbf{c_i},\mathbf{c_j}){=}\mathds{1}({\mathbf{{c_i}}{=}\mathbf{c_j}}),
\end{equation}
where $\mathds{1}$ signifies the identity operator that yields one when its internal condition is satisfied and zero otherwise.
Our approach involves minimizing the Kullback-Leibler (KL) divergence between the actual distribution $p$ and the estimated model distribution $\hat{p}$:
\vspace{-0.7em}
\begin{equation}
    \label{eq:dlmain}
    D_\mathrm{KL}[p(y|\mathbf{c_i},\mathbf{c_j})\parallel \hat{p}(y|\mathbf{x_i}, \mathbf{x_j})].
\end{equation}
When addressing Generalized Category Discovery in the context of labeled samples, $\mathbf{c_i}$ is treated as observed, while for unlabeled samples, both category variables are considered unobserved.

\noindent\textbf{Supervised Self-Expertise.}
In supervised contrastive learning, it is assumed that one of the context variables is observable. This assumption facilitates the parameter training process by directing it through the estimation of the conditional probability $\hat{p}(y|\mathbf{c_i},\mathbf{x_j})$. When dealing with a balanced dataset, the probabilities are uniformly distributed, such that $p(\mathbf{c_i}){=}\frac{1}{K}$ and $\hat{p}(z_j{=}k){=}\frac{1}{K}$, where $K$ is the total number of classes. Using the Bayesian Network from \cref{fig:baysmain}, we can derive an upper bound for the estimation discrepancy in supervised contrastive learning:
\vspace{-1.5em}
\begin{equation}
\label{eq:uppermain}
    D_\mathrm{KL}[p(y|\mathbf{c_i},\mathbf{c_j})\parallel \hat{p}(y|\mathbf{c_i},\mathbf{x_j})]{\leq} 
    \ln\frac {N}{K}.
\end{equation}
The derivation details are provided in the Appendix.
Here, we assume $\mathbf{c_i}$ is fixed, thus considering all the labels for supervised contrastive learning results in:
\vspace{-0.2em}
\begin{equation}
\label{eq:upperall}
    D_\mathrm{KL}[p(y|{c_i},c_j)\parallel \hat{p}(y|{c_i},{x_j})]{\leq}K\ln \frac{N}{K}.
    \vspace{-0.2em}
\end{equation}
In \cref{eq:upperall}, it is evident that reducing the value of $K$ diminishes the upper bound as long as $K{>}\ln\frac{N}{K}$, thereby aligning the model's distribution more closely with the true distribution. Using this property facilitates the modulation of abstraction levels across diverse categories. Denoting the upper limit for $K$ categories as $\mathcal{S}_K$, we observe that employing $\frac{K}{2}$ categories, instead of $K$, results in:
\begin{equation}
\label{eq:k2}
    \mathcal{S}_{K}{=}2\mathcal{S}_\frac{K}{2}{-}K\ln 2.
\end{equation}
This suggests that a reduction in $K$ correlates with a decrease in the upper bound $\mathcal{S}_\frac{K}{2}{>}K\ln2$. Let's consider implementing a hierarchical structure wherein, at each stage, the distribution is approximated by bifurcating into two categories. Subsequently, within each bifurcation, we estimate $\frac{K}{2}$ independently, thus effectively dealing with $K$ categories in a hierarchically extracted manner. We denote the upper bound for this hierarchical scheme as $\hat{\mathcal{S}}_K$, we show in the Appendix:
\begin{equation}
\label{eq:upperdivide}\hat{\mathcal{S}}_K{\leq}\hat{\mathcal{S}}_\frac{K}{2}.
\end{equation}
Thus, by leveraging a hierarchical approach, we observe a reduction in the upper bound of model error as the granularity of categories increases. This phenomenon underpins our introduction of supervised self-expertise, whereby the model refines its detection granularity. By hierarchically augmenting the resolution of detection (denoted as $K$), our approach aligns the model's granularity with the ground truth labels, optimizing model performance in categorization.

\noindent\textbf{Unsupervised Self-Expertise.}
In unsupervised contrastive learning, both $c_i$ and $c_j$ are unobserved, necessitating the approximation of this distribution solely based on inputs. This means we can rewrite \cref{eq:upperall} as: 
\begin{equation}
\label{eq:upperallunsup}
    D_\mathrm{KL}[p(y|c_i,c_j)\parallel \hat{p}(y|x_i,{x_j})]{\leq}\frac{K(K{+}1)}{2}\ln \frac{N}{K}.
\end{equation}
Notably, the upper bound decreases with the reduction of $K$. However, in contrast to its supervised counterpart, the presence of $K^2$ in the upper bound precludes the adoption of a hierarchical strategy to mitigate the upper bound by merely increasing $K$. To address this limitation, we propose an alternative method that refines the consideration of the KL divergence. Specifically, rather than evaluating the KL divergence across all pairs $(c_i, c_j)$, our approach focuses on pairs where $c_i{=}c_j$, which we call hard negatives for the hierarchical approach while considering the rest of pairs in a traditional unsupervised contrastive learning. This modification restricts the analysis to negative samples within the same category, thereby offering a pragmatic approximation of the overall KL divergence while maintaining a tractable upper bound:
\begin{equation}
\label{eq:upperallunsupfix}
    D_\mathrm{KL}[p(y|c_i,c_j,c_i{=}c_j)\parallel \hat{p}(y|x_i,{x_j})]{\leq}K\ln \frac{N}{K}.
\end{equation}
Given the inaccessibility of $c_i$ and $c_j$, our approach relies on the premise that $z_i$ and $z_j$ are equivalent for the selection of negative samples. Let's denote the upper bound in \cref{eq:upperallunsupfix} as $\mathcal{U}_K$. In a manner analogous to the procedure employed in the supervised variant, the application of a hierarchical categorization strategy necessitates the introduction of an adjusted upper bound, represented by $\hat{\mathcal{U}}_K$:
\begin{equation}
\label{eq:upperdivideunsup}\hat{\mathcal{U}}_K{\leq}\hat{\mathcal{U}}_\frac{K}{2}.
\end{equation}
This observation serves as the foundational motivation for our approach, which employs hierarchical hard negatives to develop unsupervised self-expertise. At every hierarchical level, our approach emphasizes the selection of negative samples from identical categories. This strategy mirrors our earlier derivation on supervised self-expertise, wherein we employed hierarchical labels to systematically reduce the upper bound delineated in \cref{eq:upperallunsupfix}. The reduction process continues incrementally until the resolution of the labels matches that of the ground truth. This methodology underpins our effort to enhance the precision of model predictions in an unsupervised learning context, bridging the gap towards achieving granular accuracy that parallels the fidelity of ground truth annotations.
Note that while the upper bound discussed here is derived based on $\ln$, transitioning to $\lg$ alters the upper bound only by a constant factor. Since halving the category count reduces $\lg K$ by one, we opt for $\lg$ over $\ln$ in the rest of the paper.

\vspace{-0.5em}
\section{Self-Expertise for Generalized Category Discovery}
\vspace{-0.8em}

\begin{figure*}[t]
\vspace{-1.5em}
\begin{center}
\includegraphics[trim={3.75cm 8cm 9cm 0cm},width=0.87\linewidth]{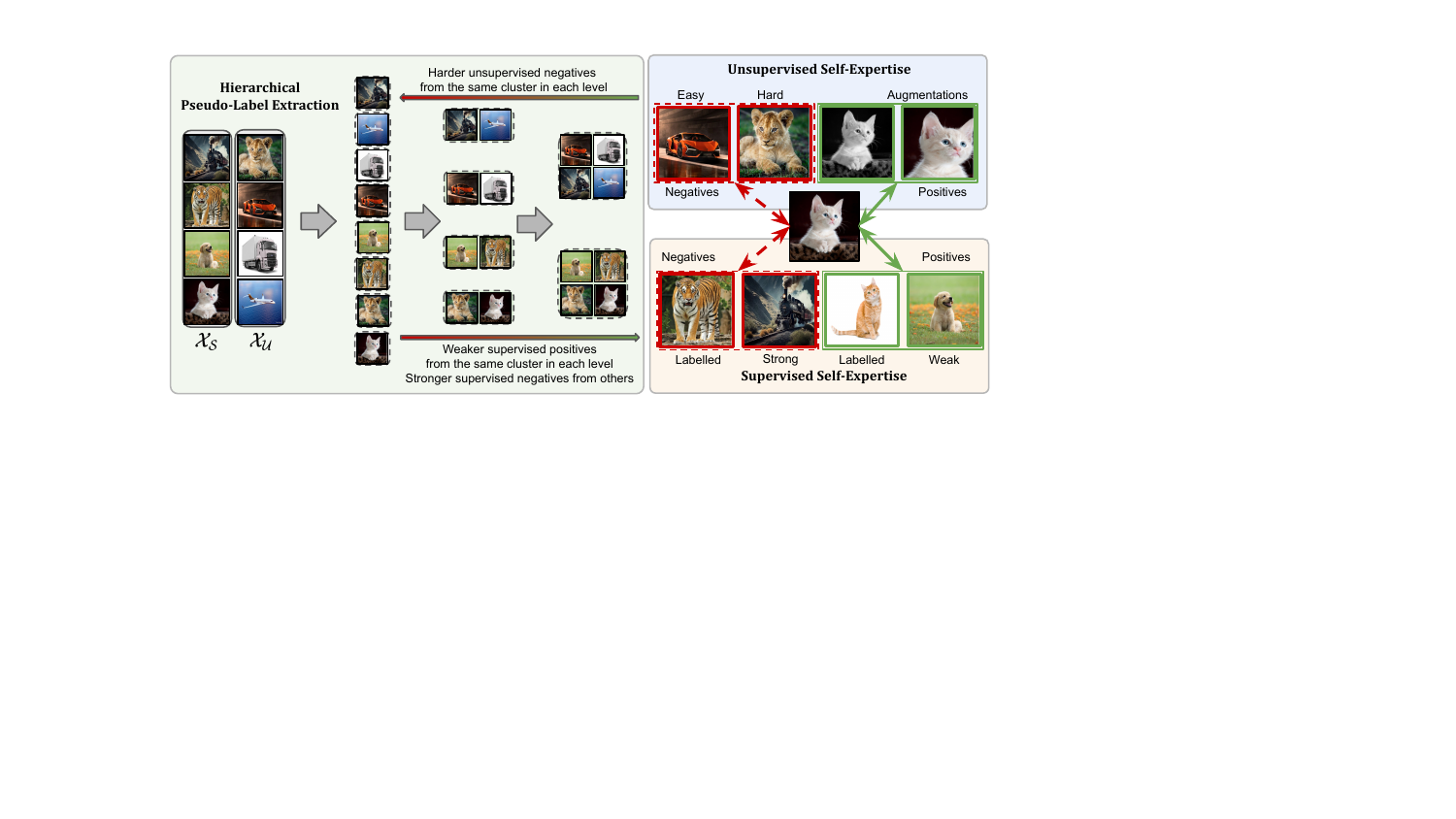}
\end{center}
\vspace{-0.5em}
\caption{\textbf{Self-expertise for Generalized Category Discovery.} Our method integrates three key components. The initial component is the Hierarchical Semi-Supervised K-means, which extracts pseudo-labels across multiple levels of expertise. Utilizing these pseudo-labels, the second component adopts unsupervised self-expertise by identifying hard negatives within each pseudo-label for enhanced differentiation across expertise tiers. The final component applies supervised self-expertise, recognizing samples sharing the same pseudo-label as weak positives to boost positive sample frequency while employing external pseudo-labels as strong negatives. This strategy accelerates cluster formation by capturing abstract-level similarities.} 
\label{fig:modelframe}
\vspace{-1.5em}
\end{figure*}

Our proposed method for fine-grained generalized category discovery has three components: hierarchical pseudo-label extraction, unsupervised self-expertise, and supervised self-expertise. As illustrated in \cref{fig:modelframe}, each phase synergistically contributes to achieving discriminative clustering, which is pivotal for the task.

\vspace{1em}
\noindent\textbf{Hierarchical Pseudo-Label Extraction.} This component addresses the challenge of optimizing supervisory signals while avoiding the erroneous allocation of unknown category samples to known categories. To achieve this, we implement a multi-tiered approach to pseudo-labeling for unlabeled samples, forming the foundation of our pseudo-label hierarchy.
\vspace{-0.5em}
\paragraph{Pseudo-label Initialization via Balanced Semi-Supervised K-means.} We propose the Balanced Semi-Supervised K-means (BSSK) algorithm. This algorithm generates pseudo-labels for the initial level of the subsequently established pseudo-label hierarchy. BSSK starts by establishing K-means centers for known categories by determining cluster centers for already labeled data. For novel categories, we select an equivalent number of random samples as cluster centers, ensuring that each cluster maintains a uniform size. This process yields the base level of our hierarchy, aligning pseudo-labels with the granularity of ground truth categories.
\vspace{-0.5em}
\paragraph{Hierarchical Expansion and Abstraction.} Based on BSSK, we introduce Hierarchical Semi-Supervised K-means (HSSK). For each subsequent $k$th level of abstraction, HSSK clusters the $k{-}1$th level's seen prototypes into half, effectively creating higher-level abstractions. All seen labels are projected onto these new hyperlabels. This is followed by BSSK, now with doubled cluster size compared to the previous level. This hierarchical structuring allows us to generate progressively abstracted and reliable pseudo-labels across various levels of category granularity. 
Pseudo-code for BSSK and HSSK is provided in the Appendix.

\begin{figure*}[t!]
\vspace{-0.5em}
\begin{subfigure}{.5\textwidth}
  \centering
  \includegraphics[width=0.7\linewidth]{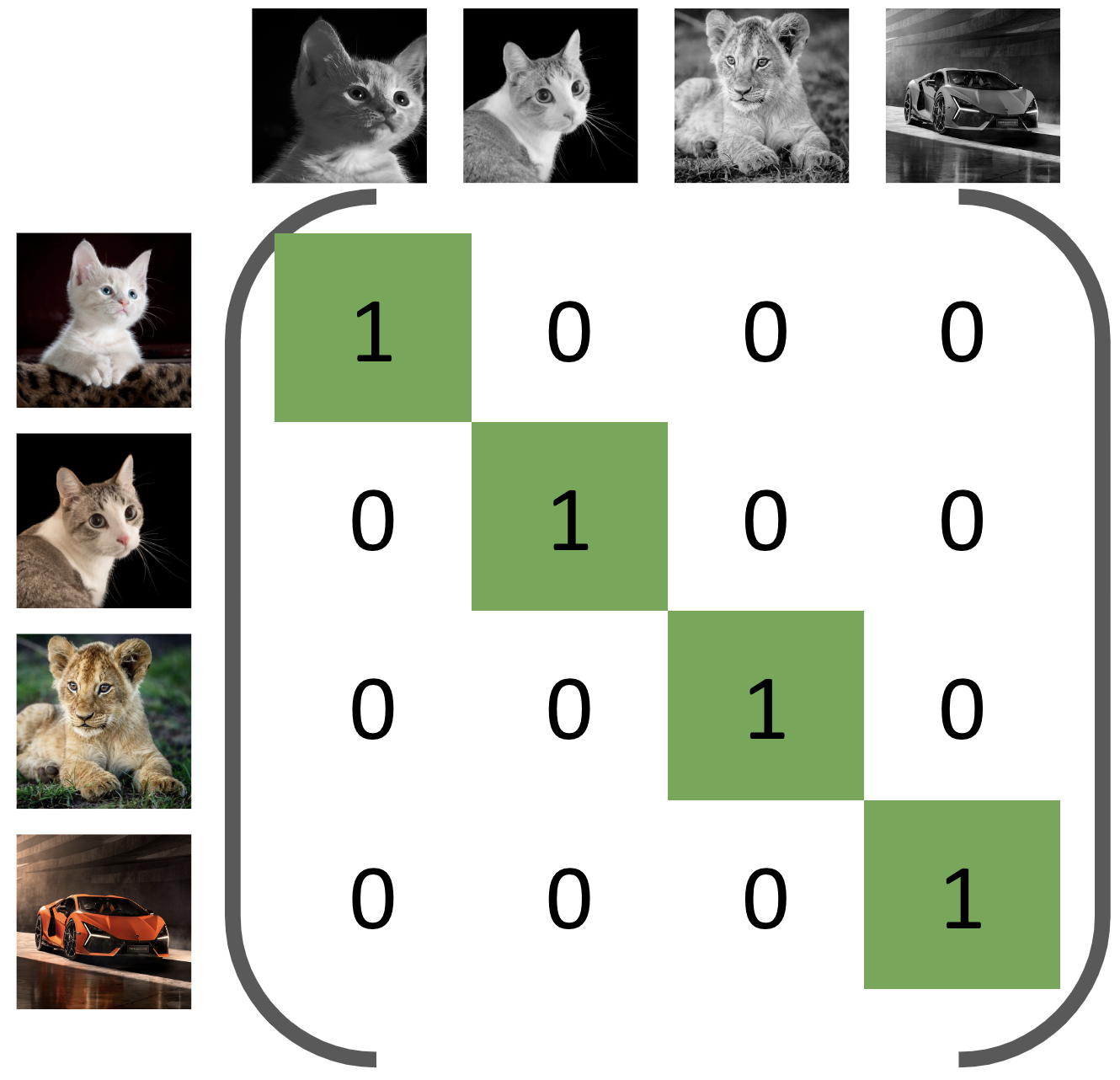}
  \caption{Unsupervised Contrastive Learning}
  \label{fig:figacons}
\end{subfigure}
\begin{subfigure}{.5\textwidth}
  \centering
  \includegraphics[width=0.7\linewidth]{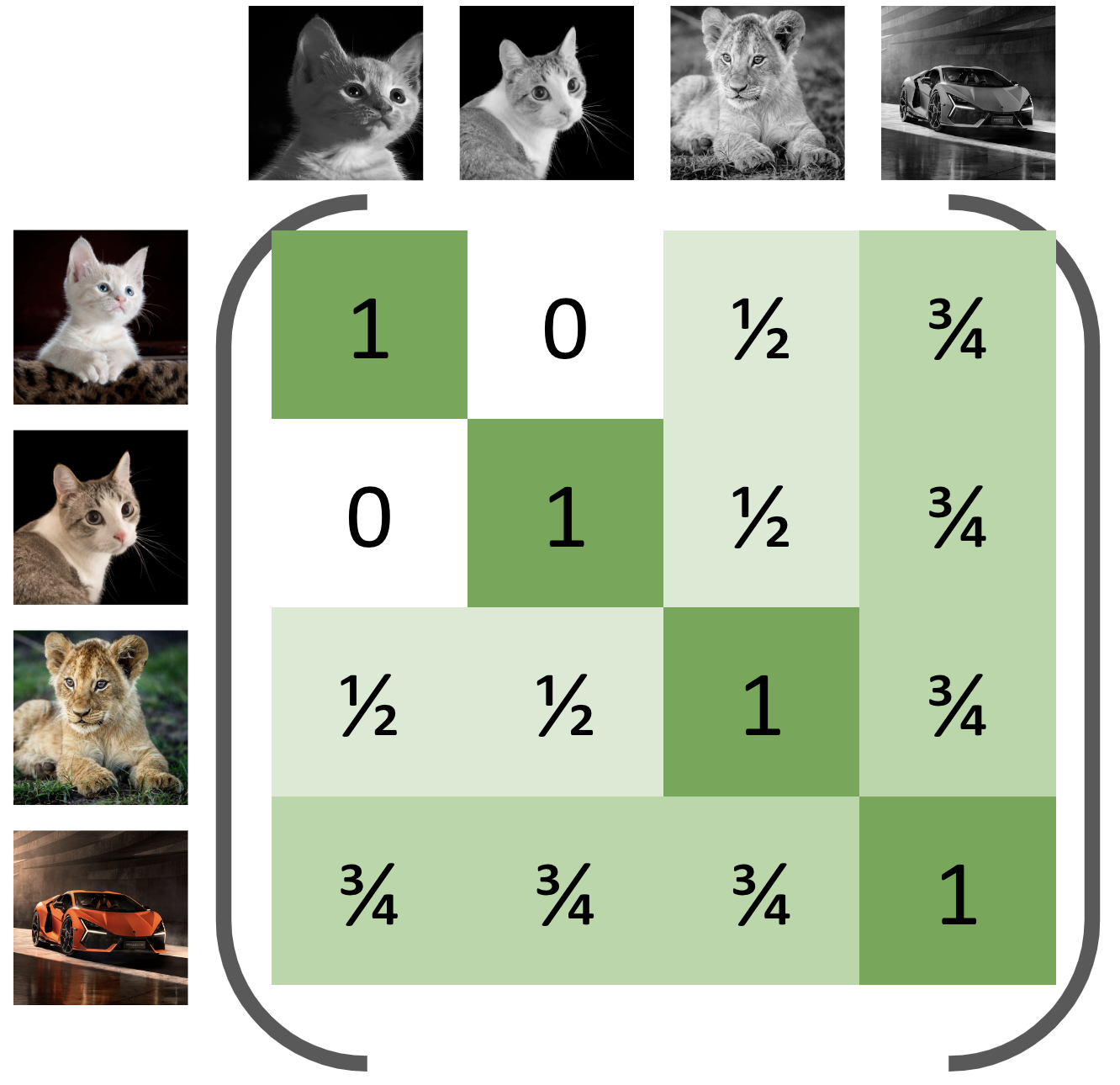}
  \caption{Unsupervised Self-Expertise}
  \label{fig:figbcons}
\end{subfigure}
\vspace{-2em}
\caption{\textbf{Illustrating the distinction in target matrix formulation for unsupervised self-expertise} \textit{(a) Unsupervised Contrastive Learning.} Each sample's augmented version is deemed positive, with all other samples marked as negative. \textit{(b)~Unsupervised Self-Expertise.} 
On the contrary, our method dynamically adjusts the negativity weight of each sample according to semantic similarity, treating categories with higher similarity (e.g., within the `cat' category) as more strict negatives. Conversely, semantically different categories (e.g., `lion' vs. `cat') incorporate a degree of uncertainty in their negativity, quantified as $\frac{1}{2}$ to reflect the semantic differences between negatives. Since the target matrix represents probabilities, normalization is required to ensure validity.}
\vspace{-1.75em}
\label{fig:unsuptarget}
\end{figure*}

\vspace{1em}
\noindent\textbf{Unsupervised Self-Expertise.}
In our approach, we confront the challenges posed by pseudo-labeling in early training stages, where model proficiency with known and unknown labels is limited. Pseudo-labels generated during this phase are often noisy, leading to sub-optimal model training. To mitigate this, we integrate unsupervised contrastive learning. 
However, this technique focuses on distinguishing augmented versions of a sample from others, including those within the same semantic context, potentially aggravating the initial issue.

To address these concerns, we adopt a strategy where the model is instructed to exclusively distance samples within the same clusters (pseudo-labels). This tactic might seem counterintuitive at first glance. However, it is fundamentally based on the notion that distancing a visually similar sample within the same cluster can significantly enhance the purity of that cluster. In contrast, samples that are semantically similar but belong to different clusters are not considered negative instances. This allows the model to either assimilate these samples during the training phase via supervised contrastive learning or to segregate them from other clusters. Our approach also systematically shifts focus towards more abstract category levels while simultaneously diminishing the importance of negative samples from these broader clusters. For instance, in traditional unsupervised contrastive learning, samples $i$ and $j$ are associated with an identity target matrix $I$, where $I_{ij} {=} \mathds{1}(i{=}j)$. In contrast, our unsupervised self-expertise necessitates the recalibration of these targets to reflect the semantic similarity between the two samples. To illustrate, we define the pseudo-label for samples $i$ and $j$ at the hierarchical level $k$ as $\mathbf{c_i}^k$ and $\mathbf{c_j}^k$, respectively. Consequently, we introduce an adjusted target matrix $Y$, comprising elements $y_{ij}$, calculated as:
\begin{equation}
    \label{eq:unsupcl}
    y_{ij} =\sum_{k=1}^{\lg K}\frac{\mathds{1}(\mathbf{c_i}^k{\neq}\mathbf{c_j}^k)}{2^k}.
\end{equation}
A comparison between the proposed adjusted target matrix and the conventional target matrix is illustrated in \cref{fig:unsuptarget}. Since the $y_{ij}$s will be interpreted as probabilities, the final $Y$ target matrix should be normalized. As depicted in \cref{fig:figacons}, standard unsupervised contrastive learning treats all negative instances uniformly, thereby ignoring their semantic dissimilarities. Conversely, our unsupervised self-expertise employs a refined target matrix, where cat instances are classified as strict negatives in \cref{fig:figbcons}, while the negativity of other instances is modulated according to their semantic distance from the positive instance. It is important to note that a linear combination of these target matrices can be employed, allowing for adjustment based on the specific granularity required by the task as:
\begin{equation}
    \label{eq:adjustedtarget}
    \hat{Y}=\alpha Y{+}(1{-}\alpha)I,
\end{equation}
where $\alpha$ represents the hyperparameter associated with label smoothing. Through empirical analysis, we demonstrate that an increased value of $\alpha$ encourages the model to pay greater attention to more nuanced details. Conversely, a reduced $\alpha$ value renders the model more adept at handling tasks that require a broader, more general approach. As a result, for the contrastive logits $P$ derived from our model, we use the binary cross entropy loss $\mathcal{L}_{\text{BCE}}$ to formulate the unsupervised self-expertise loss, $\mathcal{L}_{\text{USE}}$, as follows:
\begin{equation}
    \label{eq:unsuploss}
    \mathcal{L}_{\text{USE}} = \mathcal{L}_{\text{BCE}}(P,Y).
\end{equation}

\vspace{1em}
\noindent\textbf{Supervised Self-Expertise. }
\begin{figure*}[t]
\vspace{-1.75em}
\begin{center}
\includegraphics[width=\linewidth]{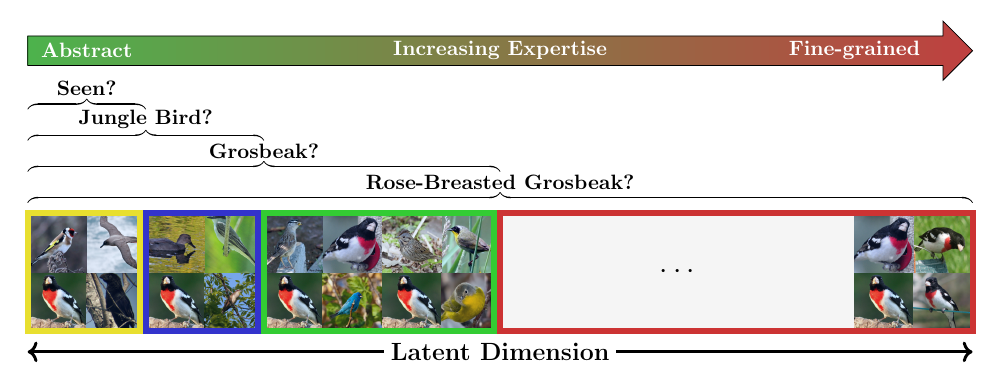}
\end{center}
\vspace{-2.5em}
\caption{\textbf{Supervised self-expertise.} Similar to playing a game of twenty questions, our method employs supervised self-expertise to discern sample attributes. As the process evolves, the attributes it discerns between are increasingly specific. Focusing on the Grosbeak classification, the model initially utilizes the leftmost segment of its latent representation (indicated by the yellow square) to ascertain whether a sample is seen. The model then allocates its representation's yellow and blue square parts to identify whether the subject is a jungle bird. Subsequently, it dedicates the latter part (represented by the green rectangle) to determine if this jungle bird is a Grosbeak.} 
\label{fig:representation}
\vspace{-1.75em}
\end{figure*}
Generated pseudo-labels from our hierarchical pseudo-label extraction are utilized as supervisory signals for the next epoch. Consider $\mathcal{L}s^k$ as the supervised contrastive learning specific to the pseudo-label level $k$. The aggregate supervised contrastive learning loss is represented by:
\vspace{-0.5em}
\begin{equation}
    \label{eq:sup}
\mathcal{L}_{\text{SSE}}={1\over2}(\sum_{k=0}^{\lg K}\frac{\mathcal{L}_s^k|{D\over 2^k}}{2^k}),
\end{equation}

where the term $\mathcal{L}_s^k|{D\over 2^k}$ reflects the supervised loss applied exclusively to the initial segment ${D\over 2^k}$ of the embedding vector $D$. This approach is grounded in the premise that higher hierarchy levels encounter an increased frequency of positive pairs. Yet, the ultimate objective is to learn pseudo-labels aligned with the ground truth labels. Consequently, the model is constrained to use only the first ${D\over 2^k}$ segment of the embedding for differentiating pseudo-labels at hierarchy level $k$. This implies that for distinguishing between various pseudo-labels at level $k{-}1$, which share a common higher-level pseudo-label at level $k$, the embedding dimensions from ${D\over 2^k}$ to ${D\over 2^{k-1}}$ are utilized. When $k{=}0$, we only use groundtruth labels for samples in known categories and utilize the full embedding vector for supervised contrastive learning. This ensures accurate label assignment for known categories upon training completion and facilitates the generation of informative pseudo-labels for novel categories. Utilizing different abstraction levels, we apply supervised contrastive learning to samples within the same cluster at different levels of hierarchy. The application of our supervised expertise to the representation is illustrated in \cref{fig:representation}.
Finally, for the tunable hyperparameter $\lambda$, our overall self-expertise loss function is expressed as:
\begin{equation}
    \label{eq:total}
\mathcal{L}_{\text{SE}}=(1-\lambda)\mathcal{L}_{\text{USE}}+\lambda\mathcal{L}_{\text{SSE}}.
\end{equation}

\vspace{-1em}
\section{Experiments}
\vspace{-0.5em}

\vspace{-0.3em}
\subsection{Experimental Setup}
\vspace{-0.2em}

\textbf{Datasets.}
We assess the efficacy of our approach on four fine-grained datasets: CUB-200~\cite{wah_branson_welinder_perona_belongie_2011}, FGVC-Aircraft~\cite{maji2013fine}, Stanford-Cars~\cite{krause20133d} and Oxford-IIIT Pet~\cite{parkhi2012cats}. Additionally, we demonstrate the adaptability of our method to more coarse-grained datasets CIFAR10~\cite{krizhevsky2009learning}, CIFAR100 ~\cite{krizhevsky2009learning} and ImageNet-100~\cite{deng2009imagenet}, highlighting its broader applicability beyond fine-grained classification tasks.  
{\colortext Finally, in the Appendix experiments section, we report on the challenging Herbarium-19 dataset~\cite{tan2019herbarium}, which is fine-grained and long-tailed, to show that our approach is effective even with non-uniform category distributions.} Detailed statistics of the datasets along with their train/test splits are also provided in the Appendix.

\noindent\textbf{Implementation Details.}
In our experiments, we adhered to the dataset division proposed by Vaze \etal \cite{vaze2022generalized}, where half of the categories in each dataset are designated as known, except for CIFAR100, where 80\% are used as known categories. The labeled set consists of 50\% of the samples from these known categories. The remainder of the known category data, along with all data from novel categories, comprise the unlabeled set. 
Following \cite{vaze2022generalized}, we use ViT-B/16 as our backbone, which is either pre-trained by DINOv1 \cite{caron2021emerging} on unlabelled ImageNet~1K \cite{krizhevsky2017imagenet}, or pretrained by DINOv2 \cite{oquab2024dinov} on unlabelled ImageNet~22K. We use the batch size of $128$ for training and set $\lambda{=}0.35$. For label smoothing, we use $\alpha{=}0.5$ for fine-grained datasets and $\alpha{=}0.1$ for coarse-grained datasets. 
Different from \cite{vaze2022generalized}, we froze the first 10 blocks of ViT-B/16 and fine-tuned the last two blocks instead of only the last one to have more parameters given that for each level, only a fraction of the latent dimension is considered.
\begin{table}[t!]
\vspace{-0.5em}
  \caption{\textbf{Comparison with state-of-the-art for fine-grained image classification.} Bold and underlined numbers indicate the best and second-best accuracies, respectively. Our method is well suited for fine-grained datasets, profits from stronger backbones, and has strong performance for all three experimental settings (\textit{All}, \textit{Known}, and \textit{Novel})}
  \label{tab:cub}
  \centering
  \begin{threeparttable}
  \resizebox{1\linewidth}{!}{
\begin{tabular}{c|l|ccc|ccc|ccc|ccc}
\toprule
&&\multicolumn{3}{c}{\textbf{CUB-200}}& \multicolumn{3}{c}{\textbf{FGVC-Aircraft}}&\multicolumn{3}{c}{\textbf{Stanford-Cars}}&\multicolumn{3}{c}{\textbf{Average}}\\ \cmidrule(lr){3-5} \cmidrule(lr){6-8} \cmidrule(lr){9-11} \cmidrule(lr){12-14}&\textbf{Method}&All&Known&Novel&All&Known&Novel&All&Known&Novel&All& Known&Novel\\
\midrule
\multirow{12}{*}{\rotatebox{90}{DINOv1}}&ORCA$^\dag$ \cite{cao2021open}&36.3&	43.8&32.6&31.6&32.0&31.4&31.9&42.2&26.9&33.3&39.3&30.3\\
&GCD \cite{vaze2022generalized}	&51.3&	56.6&	48.7&	45.0	&41.1&46.9&39.0&57.6&29.9&45.1&51.8&41.8\\
&GPC \cite{zhao2023learning}&52.0&55.5&47.5&43.3&40.7&44.8& 38.2& 58.9&27.4&44.5&51.7&39.9\\
&XCon \cite{fei2022xcon}&52.1&54.3&51.0&47.7&44.4&49.4&40.5&58.8&31.7&46.8 &52.5&44.0\\
&SimGCD \cite{wen2022simple}&60.3&65.6&57.7&	54.2&	59.1&	51.8&53.8&71.9&45.0 &56.1&65.5&51.5\\
&PIM \cite{chiaroni2023parametric}&62.7&75.7&	56.2&	-&	-&	-&43.1&66.9&31.6&-&-&-\\
&PromptCAL \cite{zhang2022promptcal}	&62.9&64.4&	62.1&	52.2&	52.2&	52.3&50.2&70.1&	40.6&55.1&62.2&51.7\\
&DCCL \cite{pu2023dynamic}	&63.5&	60.8&64.9	&	-&	-&	-&43.1&55.7&36.2&-&-&-\\
&AMEND \cite{banerjee2024amend}&64.9&75.6&59.6&52.8&61.8&48.3&56.4&73.3&48.2&58.0&70.2&52.0\\
&$\mu$GCD \cite{vaze2023clevr4}&65.7&68.0&64.6& 53.8&55.4&53.0&56.5&68.1 &\bf{50.9} &58.7&63.8&56.2\\
&SPTNet \cite{wang2024sptnet}&65.8&68.8&65.1&\bf{59.3}&61.8&\bf{58.1}&\bf{59.0}&\bf{79.2}&49.3&\underline{61.4}&69.9&\underline{57.5}\\
&CMS \cite{choi2024contrastive}&68.2&\underline{76.5}&64.0&56.0&63.4&52.3&56.9&\underline{76.1}&47.6&60.4&\underline{72.0}&54.6\\
&GCA~\cite{otholt2024guided}&68.8&73.4&\underline{66.6}&52.0&57.1&49.5&54.4&72.1&45.8&58.4&67.5&54.0\\
&InfoSieve \cite{rastegar2023learn}                 & \underline{69.4} &\bf{77.9} &65.2 
 &56.3 &\underline{63.7} &52.5&  55.7 &74.8 &46.4&60.5&\bf{72.1}&54.7\\
&TIDA \cite{wang2023discover} &-&-&-&54.6&61.3&52.1  &54.7&72.3&46.2&-&-&-\\

\cmidrule{2-14}
\rowcolor{gray!25}&\textbf{SelEx (Ours)}&\bf{73.6} &75.3 &\bf{72.8}&\underline{57.1} &\bf{64.7}&\underline{53.3}& \underline{58.5} &75.6&\underline{50.3}&\bf{63.0}&71.9&\bf{58.8}\\\midrule

\multirow{4}{*}{\rotatebox{90}{DINOv2}}
&GCD$^*$ \cite{vaze2022generalized}&71.9& 71.2& 72.3&55.4 &47.9& 59.2& 65.7&67.8&64.7&64.3&62.3&65.4\\
&SimGCD$^*$ \cite{wen2022simple}&71.5&\underline{78.1}&68.3& 63.9& \underline{69.9} &60.9&71.5&81.9&66.6&69.0&76.6&65.3\\
&$\mu$GCD$^*$ \cite{vaze2023clevr4}&\underline{74.0}&75.9&\underline{73.1}&\underline{66.3}&68.7&\underline{65.1}&\underline{76.1}& \underline{91.0}& \underline{68.9}&\underline{72.1}&\underline{78.5}&\underline{69.0}\\
\cmidrule{2-14}
\rowcolor{gray!25}&\textbf{SelEx (Ours)}            &\textbf{87.4}&\textbf{85.1}&\textbf{88.5}
 &  \textbf{79.8}& \textbf{82.3} &\textbf{78.6}
&  \bf{82.2} &\bf{93.7}&\bf{76.7}&\bf{83.1}&\bf{87.0}&\bf{81.3}\\
\bottomrule
\end{tabular}
}
\begin{tablenotes}
      \item[]$^*$ reported from \cite{vaze2023clevr4} and $^\dag$ reported from \cite{zhang2022promptcal}.
    \end{tablenotes}
    \end{threeparttable}
    \vspace{-1.5em}
\end{table}

\subsection{Comparison with State-of-the-Art}

 \begin{wraptable}{r}{0.45\textwidth}
\vspace{-3.5em}
  \caption{\textbf{Comparison with state-of-the-art for Oxford-IIIT Pet classification.} Since the Oxford Pet dataset is small, these results demonstrate our methods' robustness to overfitting.}
  \label{tab:pet}
  \centering
  \resizebox{1\linewidth}{!}{
\begin{tabular}{l|ccc}
\toprule
&\multicolumn{3}{c}{\textbf{Oxford-IIIT Pet}}\\ \cmidrule(lr){2-4} 
\textbf{Method} &All & Known  & Novel\\
\midrule
k-means \cite{arthur2007k}	&77.1	&70.1&80.7\\

GCD \cite{vaze2022generalized}	&	80.2	&85.1&77.6\\
XCon \cite{fei2022xcon}&	86.7  &	91.5&84.1\\


DCCL \cite{pu2023dynamic}	&88.1	&88.2&88.0\\
InfoSieve \cite{rastegar2023learn} &\underline{91.8} &\bf{92.6} &\underline{91.3}\\
\midrule
\rowcolor{gray!25}SelEx (DINOv1)           &\bf{92.5} & \underline{91.9}&\bf{92.8}\\ 
\rowcolor{gray!25}SelEx (DINOv2)           &\bf{95.6} & \bf{96.5}& \bf{95.1}\\
\bottomrule
\end{tabular}
}
\vspace{-3em}
\end{wraptable}
\textbf{Fine-grained image classification.}
 We evaluate our model's effectiveness across three fine-grained datasets in \cref{tab:cub}.  The results demonstrate our method's capability in handling fine-grained categories, as it consistently outperforms others in both all and novel category classification within these datasets. 
 The success can be attributed to the model's hierarchical approach to category analysis, which is pivotal in differentiating between closely related categories that demand acute attention to specific details. Additionally, as indicated in Table \ref{tab:pet}, our method also leads in performance for both all and novel categories in the Oxford Pet dataset. Despite its small size, which typically poses a risk of overfitting, our model's strong performance on this dataset further indicates its robustness. 
\begin{table}[ht!]
\vspace{-0.5em}
  \caption{\textbf{Comparison with state-of-the-art for coarse-grained image classification.} Bold and underlined numbers show the best and second-best accuracies. Our method has a consistent performance for the three experimental settings (\textit{All}, \textit{Known}, \textit{Novel}). Our method is especially suitable for known categories in all three datasets.}
  \label{tab:cifar10}
  \centering
  \begin{threeparttable}
  \resizebox{1\linewidth}{!}{
\begin{tabular}{c|l|ccc|ccc|ccc|ccc}
\toprule
&&\multicolumn{3}{c}{\textbf{CIFAR-10}}  & \multicolumn{3}{c}{\textbf{CIFAR-100}}&\multicolumn{3}{c}{\textbf{ImageNet-100}}&\multicolumn{3}{c}{\textbf{Average}}\\ \cmidrule(lr){3-5} \cmidrule(lr){6-8} \cmidrule(lr){9-11} \cmidrule(lr){12-14}
&\textbf{Method} &All & Known  & Novel &  All & Known  & Novel&All & Known  & Novel&All & Known  & Novel\\
\midrule
\multirow{12}{*}{\rotatebox{90}{DINOv1}}
&ORCA$^\dag$ \cite{cao2021open}         & 96.9 & 95.1 & 97.8 & 74.2 & 82.1  & 67.2&79.2	&93.2	&72.1&83.4&90.1&79.0\\
&GCD \cite{vaze2022generalized}    & 91.5 & \underline{97.9} & 88.2 & 73.0 & 76.2  & 66.5&74.1 & 89.8&	66.3&79.5 & 88.0  &73.7 \\
&GPC \cite{zhao2023learning} &90.6& 97.6& 87.0& 75.4 &84.6& 60.1&75.3& 93.4 & 66.7&80.4 &\underline{91.9}& 71.3\\
&XCon \cite{fei2022xcon} & 96.0&97.3&95.4&74.2&81.2&60.3&77.6& 93.5 &69.7&82.6&90.7&75.1\\
&SimGCD \cite{wen2022simple}       & 97.1 & 95.1 &98.1 &80.1 & 81.2 & 77.8& 83.0 &93.1&77.9&86.7 & 89.8& 84.6\\
&PIM \cite{chiaroni2023parametric}       & 94.7 & 97.4 & 93.3 & 78.3 & 84.2 & 66.5&83.1&\underline{95.3}&	77.0&85.4 & \bf{92.3} & 78.9\\
&PromptCAL \cite{zhang2022promptcal} &\underline{97.9} & 96.6 & \underline{98.5} & 81.2 & 84.2 & 75.3&83.1	&92.7	&78.3 &\underline{87.4} & 91.2 & 84.0\\


&DCCL \cite{pu2023dynamic}        & 96.3 & 96.5 & 96.9 & 75.3 & 76.8  & 70.2&80.5&90.5	&76.2&84.0 & 87.9  & 81.1\\



&AMEND \cite{banerjee2024amend}&96.8&94.6&97.8&81.0&79.9&\bf{83.3}&83.2&92.9&78.3&87.0&89.1&\bf{86.5}\\
&SPTNet \cite{wang2024sptnet} &97.3&95.0&\bf{98.6}& 81.3&84.3&75.6&\bf{85.4}&93.2&\bf{81.4}&\bf{88.0}&90.8&\underline{85.2}\\
&CMS \cite{choi2024contrastive}&-&-&-&\underline{82.3}&\bf{85.7}&75.5&\underline{84.7}&\bf{95.6}&\underline{79.2}&-&-&-\\
&GCA~\cite{otholt2024guided}&95.5&95.9&95.2&\bf{82.4}&\underline{85.6}&75.9&82.8&94.1&77.1&86.9&\underline{91.9}&82.7\\
&InfoSieve \cite{rastegar2023learn}                  &94.8&97.7 &93.4
& 78.3& 82.2 &70.5& 80.5 &93.8 &73.8&84.5& 91.2 &79.2\\
&TIDA \cite{wang2023discover} &\bf{98.2}&\underline{97.9}&\underline{98.5}  &\underline{82.3}&83.8&\underline{80.7} &-&-&-&-&-&-\\

\cmidrule{2-14}
\rowcolor{gray!25}&\textbf{SelEx (Ours)}         & 95.9&\bf{98.1} &94.8&\underline{82.3}&85.3&76.3
&83.1& 93.6 &77.8&87.1& \bf{92.3}&83.0\\
\bottomrule
\end{tabular}
}
\begin{tablenotes}
      \item[]$^\dag$ reported from \cite{zhang2022promptcal}.
    \end{tablenotes}
    \end{threeparttable}
    \vspace{-1em}
\end{table}

\begin{figure*}[t!]
\begin{subfigure}{.5\textwidth}
  \centering
  \resizebox{\textwidth}{!}{
   \centering
  \includegraphics[width=0.9\linewidth]{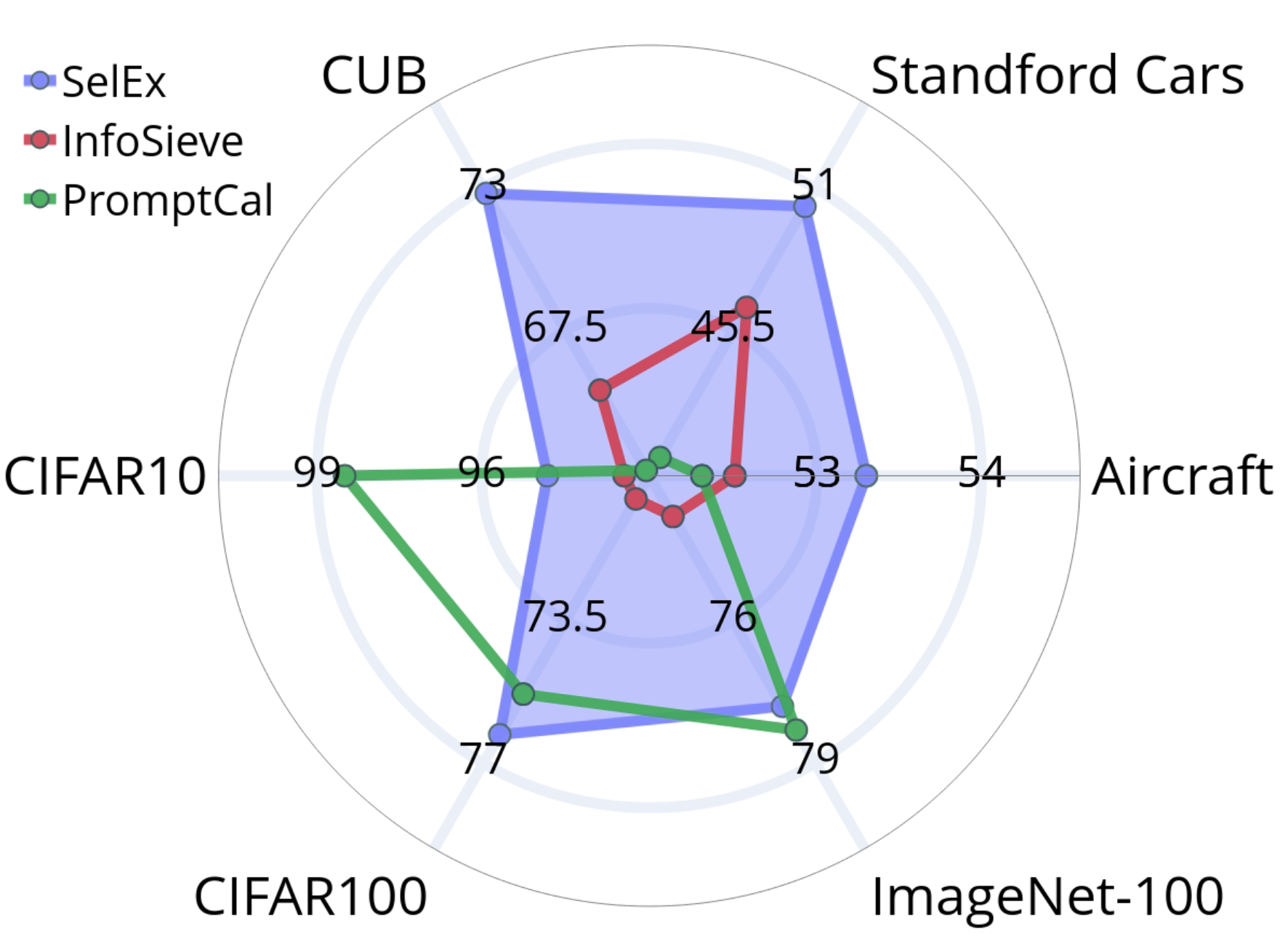}
  }
  \caption{Novel}
  \label{fig:figr2}
\end{subfigure}
\begin{subfigure}{.5\textwidth}
  \centering
\resizebox{\textwidth}{!}{
   \centering
  \includegraphics[width=0.9\linewidth]{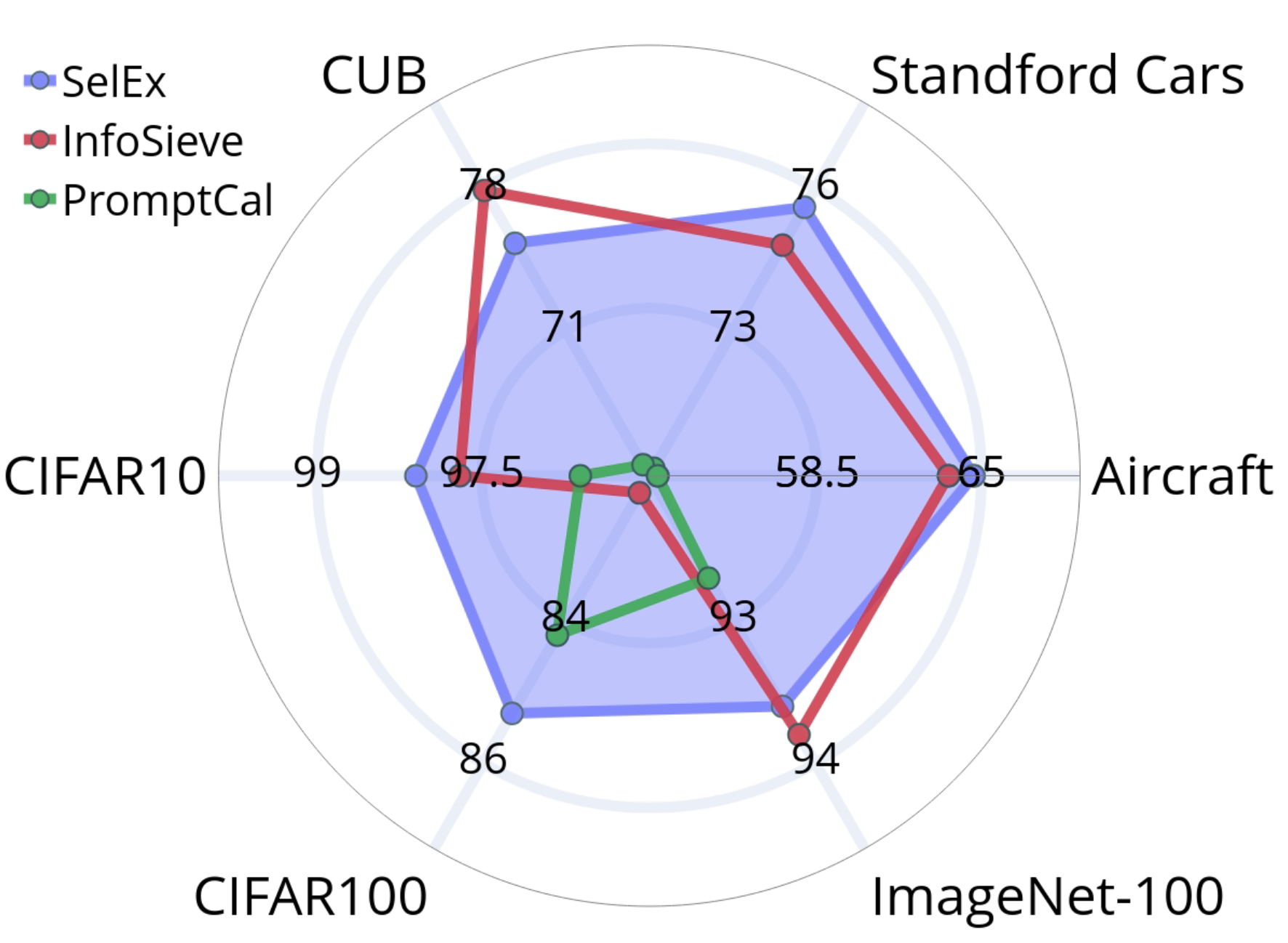}
}\caption{Known}
  \label{fig:figr3}
\end{subfigure}
\vspace{-1.5em}
\caption{\textbf{Model Performances Across Diverse Datasets:} PromptCAL \cite{zhang2022promptcal} excels in coarse-grained datasets, while InfoSieve \cite{rastegar2023learn} specializes on fine-grained datasets. SelEx is strong on both and is especially proficient in discovering novel fine-grained categories.}
\label{fig:radar}
\vspace{-2em}
\end{figure*}

\vspace{1em}
\noindent\textbf{Coarse-grained image classification.}
We also evaluate our model on three coarse-grained datasets: CIFAR10/100~\cite{krizhevsky2009learning} and ImageNet-100~\cite{deng2009imagenet}. \cref{tab:cifar10} presents a comparative analysis of our proposed method with existing state-of-the-art approaches in generalized category discovery. Our method, originally designed for fine-grained category discovery, demonstrates competitive performance on coarse-grained datasets across both known and novel categories. Despite the potential shallow or absent hierarchical structures in these datasets, our approach shows a notable enhancement in performance over the traditional non-hierarchical baseline method, GCD~\cite{vaze2022generalized}. \cref{fig:radar} presents a radar chart comparing the performance of our proposed method with that of the state-of-the-art methods across various datasets. Specifically, we contrast our approach against InfoSieve \cite{rastegar2023learn} for fine-grained datasets and PromptCAL \cite{zhang2022promptcal} for coarse-grained.
\subsection{Ablative studies}
We evaluate the individual effects of method components in this section. All ablative experiments are performed on CUB with the DINOv1 backbone. {\colortext We present additional ablations, time complexity, and failure cases in the Appendix.}
\begin{table}[t]

 \caption{\textbf{Ablation study on the effectiveness of each model component and hierarchy} using CUB-200. \textit{(a) Effect of Each Component} Each component contributes to our improved performance on known and novel categories. \textit{(b) Effect of different hierarchy levels} With more hierarchy, the model's performance increases. All pseudo-labels from previous levels are also used in each level of contrastive learning.
 }
\begin{tabular}{lr}

  \begin{minipage}{.42\linewidth}
  \resizebox{1\linewidth}{!}{
  \label{tab:compseff}
  \centering
\begin{tabular}{ccc|ccc}
\toprule
\multicolumn{6}{c}{\textbf{(a) Effect of Each Component}}\\ \cmidrule(lr){1-6}
 $\mathbf{\text{HSSK}}$ &
   $\mathbf{\mathcal{L}_{\text{USE}}}$ & $\mathbf{\mathcal{L}_{\text{SSE}}}$   &  All & Known  & Novel\\
\midrule
& & &  46.1 &49.2 &44.5\\

 \gr{\ding{51}}  &  &  &  62.6&71.9&58.2 \\
  & \gr{\ding{51}} &  &47.0& 52.2& 44.4\\
  &  & \gr{\ding{51}} &55.5&53.5&56.6 \\
 \gr{\ding{51}}  & \gr{\ding{51}} &  &54.7 &66.7&48.6\\
  \gr{\ding{51}}  &  & \gr{\ding{51}} &68.7 &72.3&66.9\\
   & \gr{\ding{51}} & \gr{\ding{51}} & 56.7 &51.8 &59.1\\

\midrule
  \gr{\ding{51}} & \gr{\ding{51}} &\gr{\ding{51}
  }&\bf{73.6} &\bf{75.3} &\bf{72.8}\\              
\bottomrule
\end{tabular}
}
  \end{minipage} &

\begin{minipage}{.6\linewidth}
\label{table:hierarchy}
\centering
\resizebox{0.9\linewidth}{!}{
\begin{tabular}{llccc}
\toprule
\multicolumn{5}{c}{\textbf{(b) Effect of Hierarchy Levels}}\\ \cmidrule(lr){1-5}
\textbf{Hierarchy}&Pseudo-labels  &  All & Known  & Novel\\
\midrule
Baseline (None)&+0  &62.6&71.9&58.2 \\
Level 1 &+200  &63.8 &74.9&58.3\\
Level 2 &+100 &69.2&\bf{76.8} &65.3\\
Level 3&+50 &70.0&74.5&67.9\\
Level 4 &+24 &71.0&74.4&69.4\\
Level 5 &+12&69.0&72.5& 67.2\\
Level 6 &+6 &72.5 &75.4&71.1\\
Level 7 &+2&\bf{73.6} &75.3 &\bf{72.8} \\
\bottomrule
\end{tabular}}
  \end{minipage}\\
\end{tabular}
\vspace{-2em}
\end{table}

\noindent\textbf{Effect of each component}.
\cref{tab:compseff} (a) examines the effect of our three key method components: Hierarchical Semi-Supervised K-means (HSSK), unsupervised self-expertise ($\mathcal{L}_{\text{USE}}$), and supervised self-expertise ($\mathcal{L}_{\text{SSE}}$).
The results demonstrate that the Hierarchical Semi-Supervised K-means approach yields the most significant improvements across both known and novel categories. Our unsupervised self-expertise loss, denoted as $\mathcal{L}_{\text{USE}}$, shows a particular affinity for enhancing known categories. This is in line with our initial hypothesis, considering that these categories benefit from supervision signals. Such signals facilitate the attraction of semantically similar samples, even if they are initially distant in the embedding space. Concurrently, this approach effectively disregards semantically similar yet distant negative samples, preventing any repulsion until they converge into the same cluster. When integrated with hierarchical semi-supervised k-means, the unsupervised self-expertise loss extends its benefits to novel categories, leveraging the presence of semantic labels. Our supervised self-expertise loss, $\mathcal{L}_{\text{SSE}}$, unsurprisingly excels in aiding novel categories while also contributing positively to known ones. We attribute this to the fact that, although hierarchical structures are advantageous for known categories with robust label-based supervision, novel categories lack such ground-truth labels. As a result, pseudo-labels at finer granularities may introduce noise. However, as we ascend the hierarchy, these pseudo-labels for novel categories gain reliability, offering more effective supervision. In conclusion, the combination of all three components – hierarchical semi-supervised k-means, unsupervised self-expertise, and supervised self-expertise – yields the most optimal results for both known and novel categories, as demonstrated in our experiments.

\noindent \textbf{The effect of hierarchy level.}
In \cref{table:hierarchy} (b), we compare model performance across varying hierarchy levels. These hierarchy levels are incorporated into the training phase for all three model components. Specifically, the Baseline component employs supervised contrastive learning using only the ground-truth labels, which are limited to samples that have been labeled. Level $1$ is identified as the base level of the hierarchy, utilizing pseudo-labels that offer semantic detail comparable to ground-truth labels, thereby enriching our dataset with an additional $200$ pseudo-labels for samples without labels. As we ascend through the hierarchy levels, the quantity of pseudo-labels decreases by half, as detailed in the accompanying table, until reaching the apex level. This topmost level introduces the most abstract categorization, distinguishing between `seen' and `unseen' samples.
The results depicted in \cref{table:hierarchy} (b) indicate a notable trend: integrating additional hierarchical levels appears to be particularly advantageous for unknown categories. This observation can be attributed to increased granularity between categories at finer hierarchy levels, resulting in heightened uncertainty and noise in pseudo-labels. This phenomenon underscores the efficacy of our model in handling complex, hierarchical category structures, especially in scenarios involving unknown category distinctions.

\begin{wrapfigure}{r}{0.44\textwidth}
\vspace{-1.95em}
\includegraphics[width=\linewidth]{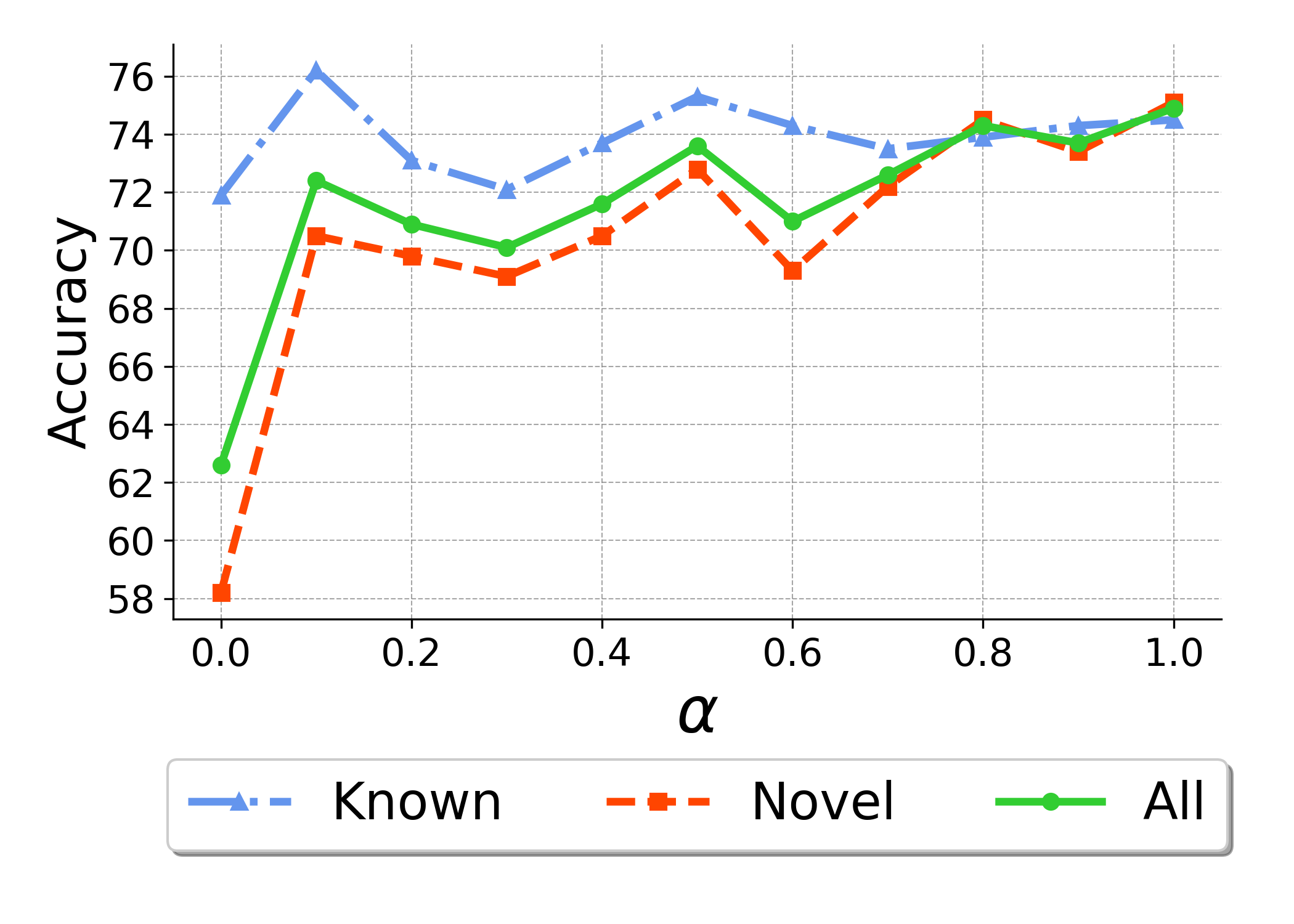}
\vspace{-2.5em}
\caption{\textbf{Effect of smoothing hyperparameter constant}. A higher smoothing hyperparameter strengthens the focus on negative samples within the cluster; this enhances performance on novel categories while decreasing it on known categories. 
}\label{table:smoothingconst}
\vspace{-3em}
\end{wrapfigure}

\noindent\textbf{Effects of smoothing hyperparameter.}
In our unsupervised self-expertise, we adopted a smoothing hyperparameter $\alpha$ to modulate the uncertainty threshold for negative samples outside a given cluster. Specifically, when $\alpha{=}1$, the model exclusively incorporates negative samples from its own cluster. Conversely, setting $\alpha{=}0$ equalizes the treatment of all negative samples, aligning with traditional unsupervised contrastive learning. We conducted experiments with varying $\alpha$ values, as detailed in \cref{table:smoothingconst}. Our findings indicate an enhancement in the model's performance on novel categories as $\alpha$ increases. This improvement is attributed to the fact that standard unsupervised contrastive learning indiscriminately distances all non-matching samples, including those with semantic similarities.

In contrast, our unsupervised self-expertise concentrates on cluster-specific samples. This allows semantically related samples outside the cluster to be less repelled, which is particularly beneficial for novel categories since they do not have ground-truth labels to counteract this repelling through supervised contrastive learning. Hence, a higher $\alpha$ enhances novel category identification performance. It is essential to highlight that dataset granularity can influence the choice of the hyperparameter 
 $\alpha$. Specifically, a more fine-grained dataset necessitates a larger value of $\alpha$ to discern the subtle differences between samples. We demonstrate the impact of $\alpha$ {\colortext to balance the probability and uncertainty} on various datasets in the Appendix.

\vspace{-0.5em}
\section{Conclusion}
\vspace{-0.3em}
This work presents self-expertise in identifying and categorizing known and previously unknown categories, focusing on fine-grained distinctions. We introduce a method that utilizes hierarchical structures to effectively bridge the gap between labeled data for known categories and unlabeled data for novel categories. This is achieved by generating hierarchical pseudo-labels, which guide both supervised and unsupervised learning phases of our self-expertise framework. The supervised phase is designed to incrementally increase the complexity of differentiation tasks, thereby accelerating the training process and enhancing the formation of distinct clusters for unknown categories. This strategy improves the model's ability to generalize to novel categories. In the unsupervised phase, we integrate a label-smoothing hyperparameter, compelling the model to concentrate on negative samples within a localized context and to make finer distinctions. This approach enhances the model's fine-grained categorization capabilities. Overall, our work demonstrates the effectiveness of self-expertise in handling unknown and fine-grained categorization tasks. {\colortext 
In the Appendix section titled `Discussions,' we outline the limitations of our work and propose directions for future research.}
\vspace{-0.5em}

\section*{Acknowledgments}
\vspace{-0.8em}
{\colortext This work is part of the project Real-Time Video Surveillance Search with project number 18038, which is (partly) financed by the Dutch Research Council (NWO) domain Applied and Engineering/ Sciences (TTW).}

\bibliographystyle{splncs04}
\bibliography{main}

\newpage
\appendix
\vspace{-0.3em}
\section{Related Works}
\vspace{-0.8em}
To provide a more in-depth analysis of the relevant literature, this appendix section delves into additional related works. 
\vspace{-0.3em}
\subsection{Self-Supervised Learning}
\vspace{-0.2em}

Self-supervised learning has transformed the analysis of large-scale unlabeled data, learning rich representations. He~\etal~\cite{he2020momentum} introduced the concept of momentum contrast with MoCo, enhancing the quality of learned representations by utilizing two encoders and a contrastive loss to learn image features that distinguish between different views~\cite{he2020momentum}. Following this, Chen~\etal~ \cite{chen2020simple} simplified the self-supervised learning pipeline with SimCLR, by maximizing agreement between different augmented versions of the same image, using a contrastive loss function~\cite{chen2020simple}. Caron~\etal~\cite{caron2020unsupervised} introduced SwAV, which employs a clustering mechanism to enhance consistency between cluster assignments, thereby improving learning efficacy~\cite{caron2020unsupervised}. By employing a self-distillation with information noise injection, DINO~\cite{caron2021emerging} efficiently captures complex visual features without labeled data. DINOv2~\cite{oquab2024dinov} further refines this approach with key improvements, enhancing both feature quality and model adaptability.
\vspace{-0.3em}
\subsection{Open-Set Recognition}
\vspace{-0.2em}
The advent of open set recognition marked a significant milestone in the evolution of computational models, addressing the complexities of processing real-world data. This domain was first conceptualized by Scheirer~\etal~\cite{scheirer2012toward}, who laid the foundational theory for models capable of discerning between known and unknown data categories, a principle further developed by subsequent research~\cite{bendale2015towards, scheirer2014probability}. The pioneering application of deep learning techniques to tackle open-set recognition challenges was introduced through the development of OpenMax~\cite{bendale2016towards}. The core objective within open-set recognition is the accurate identification of known categories while effectively filtering out novel category samples. Various methodologies have been proposed to emulate the concept of ``otherness'' essential for distinguishing unknown categories. These approaches range from identifying significant reconstruction errors~\cite{oza2019c2ae, yoshihashi2019classification}, measuring the deviation from a set of predefined prototypes~\cite{chen2021adversarial, chen2020learning, shu2020p}, to differentiating samples generated through adversarial processes~\cite{ge2017generative, kong2021opengan, neal2018open, yue2021counterfactual}. Despite its advances, a notable limitation of open-set recognition is its inclination to disregard all instances of novel classes, potentially omitting valuable information. This challenge underscores the ongoing quest for more sophisticated models capable of not only identifying the unknown but also accommodating the continuous expansion of the knowledge domain.
\vspace{-0.3em}
\subsection{Novel Category Discovery}
\vspace{-0.2em}
The foundation for this line of inquiry was laid by Han~\etal.~\cite{han2019learning}, who adapted classification models to recognize novel categories based on knowledge from known categories. Initial strategies, as documented in several studies~\cite{han2019deep, hsu2017learning, hsu2019multi}, typically employed a bifurcated approach. Initially, these methods focused exclusively on learning representations from annotated data, subsequently applying the acquired categorical structures to identify unknown categories in a distinct phase. However, the research trajectory has recently veered towards unified methodologies. For instance, studies like~\cite{fini2021unified,han2020automatically,zhao2021novel,zhong2021openmix,roy2022class,rizve2022towards,yu2024continual,liu2023large,liu2024novel} exemplify this trend by integrating the learning process, simultaneously leveraging both labeled and unlabeled datasets to enhance representation learning. 
Despite the significant progress in this area, a critical issue remains unaddressed: the assumption that known and novel categories are mutually exclusive. This assumption undermines the performance of models in real-world settings, where instances may belong to either known or novel categories. This issue is what \textit{Generalized Category Discovery} seeks to address.
\vspace{-0.3em}
\subsection{Generalized Category Discovery}
\vspace{-0.2em}
Generalized category discovery was introduced by Vaze~\etal~\cite{vaze2022generalized} and Cao~\etal~\cite{cao2021open}. It provides models with unlabeled data from both novel and known categories, placing it within the realm of semi-supervised learning, a domain thoroughly investigated in the machine learning literature~\cite{ouali2020overview, yang2022survey, rebuffi2020semi, oliver2018realistic, chapelle2009semi}. The unique challenge in generalized category discovery is handling categories without any labeled instances, which introduces additional complexity. There are primarily two approaches to address this challenge. The first employs a series of prototypes as reference points, \eg,~\cite{hao2023cipr, chiaroni2023parametric, wen2022simple,wang2024beyond,zhang2022automatically,tan2024revisiting,choi2024contrastive,Xiao_2024_CVPR,yang2023bootstrap,kim2023proxy,an2023generalized}. 
The second approach leverages local similarity as weak pseudo-labels for each sample and utilizes sample similarities to form local clusters~\cite{pu2023dynamic,zhang2022promptcal, hao2023cipr, chiaroni2023parametric, rastegar2023learn, banerjee2024amend, otholt2024guided,gao2023opengcd,zhao2023learning,du2023fly}. Studies such as~\cite{vaze2023clevr4,zhang2022promptcal, wen2022simple,wang2024sptnet} employ mean-teacher networks to tackle the issues arising from noisy pseudo-labels. Additionally, other research efforts leverage cues from alternative modalities to identify categories in a multimodal fashion~\cite{zheng2024textual,ouldnoughi2023clip,an2023generalized2}. Our approach introduces `self-expertise', a novel concept aimed at hierarchical learning of known and unknown categories. This technique emphasizes the focus on the samples from identical clusters at each level. This method is particularly effective in overcoming the limited availability of positive samples per category, while also enhancing the identification of subtle differences among negative samples.  

\vspace{-0.3em}
\subsection{Hierarchical Representation Learning}
\vspace{-0.2em}
In the realm of leveraging hierarchical categories for enhanced representation learning, several approaches have been introduced. Zhang~\etal~\cite{zhang2022use} employ multiple label levels to augment their models' representational capacity through hierarchical contrastive learning. Similarly, Guo~\etal~\cite{guo2022hcsc} extract pseudo-labels to facilitate hierarchical contrastive learning, with a unique emphasis on ensuring signals remain positive within identical clusters. The hierarchical structure of categories has been explored in previous works such as~\cite{Lang_2024_CVPR,Wu_2024_CVPR}, aiming to enhance the identification of novel categories within the context of open-set recognition. Additionally, Rastegar~\etal~\cite{rastegar2016mdl} utilize a hierarchical structure for multimodal data to infer missing modalities, which can be categories. An alternative perspective on extracting hierarchical structures of categories through self-supervision can be viewed through the framework of deep metric learning~\cite{kim2020proxy,yang2022hierarchical}. 
Our research parallels these efforts by also utilizing hierarchical pseudo-labels. However, our methodology diverges by incorporating negative samples from the same cluster to foster generalized category discovery, setting our approach apart.

Moreover, Otholt~\etal~\cite{otholt2024guided} and Banerjee~\etal\cite{banerjee2024amend} have proposed hierarchical strategies aimed at generalized category discovery, primarily leveraging neighborhood structures to define refined categories with precision. Rastegar~\etal~\cite{rastegar2023learn} introduced a technique for learning an implicit category tree, which aids in the hierarchical self-coding of categories, ensuring category similarity is maintained throughout all levels of the hierarchy. Similarly, Wang~\etal~\cite{wang2023discover} leverage taxonomic context consistency to maintain coherence in the targeted labels, ensuring alignment across both coarse and fine-grained pseudo-labels. Our work distinguishes itself from these methodologies by employing weak supervision from samples within each hierarchy level to minimize the impact of misclassifications on lower levels. Furthermore, our approach emphasizes hard negatives within the same cluster for unsupervised self-expertise, thereby improving the model's capability to identify subtle distinctions within fine-grained classifications. 

\clearpage
\section{Theory}

\begin{wrapfigure}{r}{0.45\textwidth}
  \centering
\vspace{-3.5em}
\begin{tikzpicture}
  \node[latent, xshift=0cm](y) {$y$};
    \node[obs, above=of y, xshift=-1cm, yshift=-0.5cm]  (ci) {$\mathbf{c_i}$};
\node[latent, above=of y, xshift=1cm, yshift=-0.5cm]  (zj) {$z_j$};
  \node[obs, below=of zj, xshift=0cm,yshift=0cm] (xj) {$\mathbf{x_j}$};
  \node[obs, below=of ci, xshift=0cm,yshift=0cm]  (xi) {$\mathbf{x_i}$};
   \node[latent, right=of zj, xshift=-0cm]                               (cj) {$\mathbf{c_j}$};

  \edge {ci,zj} {y} ; %

        \edge {ci} {xi} ; %
            \edge {xj} {zj} ; %
            \edge {cj} {xj} ; %



\end{tikzpicture}
\vspace{-0.8em}
\caption{\textbf{Bayesian Network for the Generalized Category Discovery.} Shaded nodes are observed variables $\mathbf{x_i}$, $\mathbf{x_j}$ corresponds to images $i$ and $j$, and $\mathbf{c_i}$ and $\mathbf{c_j}$ which are the ground-truth category variable. $z_j$ is the latent category variable extracted from the model.}
\label{fig:bays}
\vspace{-4em}
\end{wrapfigure}
In this section, we motivate our approach of hierarchical contrastive learning and why it is particularly well-suited for category discovery in unseen data. 

\subsection{Notations and Definitions}
Let's consider the simple Bayesian networks depicted in Fig. \ref{fig:bays}. Here, $\mathbf{x_i}$ and $\mathbf{x_j}$ are different samples or different views of the same sample which are observed. Their corresponding ground truth context variables $\mathbf{c_i}$ and $\mathbf{c_j}$s are variables we aim to extract information about. These context variables can be partly observed in the form of labels, as is the case for supervised contrastive learning. However, these context variables are unobserved for unsupervised contrastive learning, which we show with $z_i$ and $z_j$, respectively. Random variable $y$ will indicate if its two parents have the same value, or in other terms; it will provide the contrastive labels. 

Let's assume that we have $K$ total categories and the number of total samples as $N$. For each context variable $c$, we show the number of samples assigned to it by $|c|$. Hence, $|z_i|$ indicates the number of samples that the model has assigned to modulo set ${\{z_i\}}_{K}$ and $|\mathbf{c_i}|$ indicates how many samples the true distribution has assigned to labels ${\{\mathbf{c_i}\}}_{K}$. Finally, we show the true distribution of samples with $p$ and the approximated one with $\hat{p}$. In the next sections, we use labels instead of ground truth context variables, but note that these labels can have different hierarchy levels from the ground truth labels. To prevent ambiguity, we emphasize ground truth labels whenever we use them.  

\subsection{Problem Definition}
For contrastive training, the goal is to estimate the true distribution of equality of the ground truth context variables. Hence, for samples $i$ and $j$, the goal is to approximate the following true distribution,
\begin{equation}
    \label{eq:identy}
    p(y{=}1|\mathbf{c_i},\mathbf{c_j})=\mathds{1}({\mathbf{{c_i}}{=}\mathbf{c_j}}),
\end{equation}
in which $\mathds{1}$ is the identity operator, which is one only when its inner condition holds and zero otherwise.
Depending on the problem we aim to solve, in training, we have no access to the ground truth context variables as in unsupervised contrastive learning, or we have partial access as in supervised contrastive learning. 
In both contrastive learning formulations, we aim to minimize the KL divergence between the true distribution $p$ and our model distribution $\hat{p}$ which in case of unsupervised contrastive learning will be,
\begin{equation}
    \label{eq:dl}
    D_\mathrm{KL}[p(y|\mathbf{c_i},\mathbf{c_j})\parallel \hat{p}(y|\mathbf{x_i}, \mathbf{x_j})],
\end{equation}
and for supervised contrastive learning will be,
\begin{equation}
    \label{eq:dl}
    D_\mathrm{KL}[p(y|\mathbf{c_i},\mathbf{c_j})\parallel \hat{p}(y|\mathbf{c_i}, \mathbf{x_j})].
\end{equation}
In the generalized category discovery problem, for labeled samples, we can consider the context variable as $\mathbf{c_i}$ since it is observed, while for unlabeled samples, both context variables are unobserved. 
\subsection{Supervised Contrastive Learning}
\setcounter{theorem}{0}
\begin{theorem}
\label{theo:kolmo}
Consider two samples $i$ and $j$ from the Bayesian network \cref{fig:bays}. If we have a total of $N$ samples and $K$ categories and the dataset is balanced, we will have the following upper bound if only $i$'s label $\mathbf{c_i}$ is known:
\begin{align}  
\label{eq:uppersup}
    D_\mathrm{KL}[p(y|\mathbf{c_i},\mathbf{c_j})\parallel \hat{p}(y|\mathbf{c_i},\mathbf{x_j})]\leq 
    \ln\frac{N}{K}. 
\end{align}

\end{theorem}
\textit{Proof.} According to the Bayesian network shown in \cref{fig:bays}, we have:
\begin{equation}
\label{eq:eq1}
   \hat{p}(y|\mathbf{c_i},\mathbf{x_j})=\sum_{z_j=1}^K\hat{p}(z_j|\mathbf{x_j})\hat{p}(y|\mathbf{c_i},z_j).
\end{equation}
Note that since $\sum_{z_j=1}^K\hat{p}(z_j|\mathbf{x_j}){=}1$ and $p(y|\mathbf{c_i},\mathbf{c_j})$ is independent of $z_j$\footnote{Note that $p$ is the true distribution and $\hat{p}$ is the distribution that is derived from the bayesian network.}, we can consider that
\begin{equation}
\label{eq:eq2}
    p(y|\mathbf{c_i},\mathbf{c_j}){=}\sum_{z_j=1}^K\hat{p}(z_j|\mathbf{x_j})p(y|\mathbf{c_i},\mathbf{c_j}).
\end{equation} 
Also, since KL divergence is convex, we can have the following inequality:
\begin{multline}
\label{eq:eq3}
        D_\mathrm{KL}[\sum_{z_j=1}^K\hat{p}(z_j|\mathbf{x_j})p(y|\mathbf{c_i},\mathbf{c_j})
    {\parallel}\sum_{z_j=1}^K\hat{p}(z_j|\mathbf{x_j})\hat{p}(y|\mathbf{c_i},z_j)]\\\leq\sum_{z_j=1}^K\hat{p}(z_j|\mathbf{x_j})D_\mathrm{KL}[p(y|\mathbf{c_i},\mathbf{c_j}){\parallel} \hat{p}(y|\mathbf{c_i},z_j)].
\end{multline}
If we use equations \cref{eq:eq1,eq:eq2,eq:eq3} on the left-hand side of the inequality \cref{eq:uppersup}, we will have the following: 
\begin{align*}
    D_\mathrm{KL}[p(y|\mathbf{c_i},\mathbf{c_j}){\parallel}\hat{p}(y|\mathbf{c_i},\mathbf{x_j})]=&D_\mathrm{KL}[p(y|\mathbf{c_i},\mathbf{c_j})
    {\parallel}\sum_{z_j=1}^K\hat{p}(z_j|\mathbf{x_j})\hat{p}(y|\mathbf{c_i},z_j)]\\
    =&D_\mathrm{KL}[\sum_{z_j=1}^K\hat{p}(z_j|\mathbf{x_j})p(y|\mathbf{c_i},\mathbf{c_j})
    {\parallel}\sum_{z_j=1}^K\hat{p}(z_j|\mathbf{x_j})\hat{p}(y|\mathbf{c_i},z_j)]\\ 
    \leq&\sum_{z_j=1}^K\hat{p}(z_j|\mathbf{x_j})D_\mathrm{KL}[p(y|\mathbf{c_i},\mathbf{c_j}){\parallel} \hat{p}(y|\mathbf{c_i},z_j)],
\end{align*}
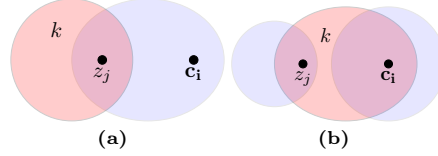
\begin{wrapfigure}{r}{0.5\textwidth}
\vspace{-2em}
  \centering
\begin{subfigure}{.24\textwidth}
  \centering
\centering
\resizebox{1\linewidth}{!}{
\begin{tikzpicture}
  \draw[fill=blue, opacity=0.1] (1.25,0) ellipse (1.25cm and 1cm);

\draw[fill=red, opacity=0.2] (0,0) circle (1cm);
  \filldraw [black] (0.5,0) circle (2pt) node[anchor=north] {$z_j$};

  \filldraw [black] (2,0) circle (2pt) node[anchor=north] {$\mathbf{c_i}$};
  \node at (-0.25,0.5) {$k$};
\end{tikzpicture}}
\caption{}
  \label{fig:sfig1}
\end{subfigure}%
\begin{subfigure}{.24\textwidth}
  \centering
  \resizebox{1\linewidth}{!}{
\begin{tikzpicture}
  \draw[fill=red, opacity=0.2] (1.25,0) ellipse (1.25cm and 1cm);

\draw[fill=blue, opacity=0.1] (0,0) circle (0.75cm);
  \filldraw [black] (0.5,0) circle (2pt) node[anchor=north] {$z_j$};
\draw[fill=blue, opacity=0.1] (2,0) circle (1cm);
  \filldraw [black] (2,0) circle (2pt) node[anchor=north] {$\mathbf{c_i}$};
  \node at (0.9,0.5) {$k$};
\end{tikzpicture}}
  \caption{}
  \label{fig:sfig2}
\end{subfigure}
\vspace{-0.5em}
\caption{\textbf{Different scenarios for two sample category variables in supervised contrastive learning.} The latent variable $z_j$ is extracted from the model. Model does the red assignments, while blue assignments indicate the ground truth category variable. Since only $\mathbf{c_i}$ is known, the model makes a mistake if: (a) Both $i$ and $j$ have the same labels and the model has assigned $j$ to another label $k$. (b) $i$ and $j$ have different labels but $j$ is assigned to the label $k{=}\mathbf{c_i}$.} 
\label{fig:appclusterssup}
\vspace{-5em}
\end{wrapfigure}
For $D_\mathrm{KL}[p(y|\mathbf{c_i},\mathbf{c_j}){\parallel} \hat{p}(y|\mathbf{c_i},z_j)]$, two possible scenarios can happen, i.e., $\mathbf{c_i}{=}\mathbf{c_j}$ and $\mathbf{c_i}{\neq}\mathbf{c_j}$. \cref{fig:appclusterssup} depicts these different scenarios in which the model makes a mistake in assigning the label. We call these two scenarios intra-class divergence and inter-class divergence, respectively. 
Note that since the dataset is balanced, we can consider that $|z_j|{=}|\mathbf{c_j}|{=}\frac{N}{K}$. We also assume $\hat{p}(\mathbf{c_j}){\approx}p(\mathbf{c_j}){=}\hat{p}(z_j){=}\frac{1}{K}$.
\subsubsection{Intra-class Divergence.}
If $\mathbf{c_i}{=}\mathbf{c_j}$, we will have $p(y{=}0|\mathbf{c_i},\mathbf{c_j}){=}0$, according to \cref{eq:identy}. So we will have:
\begin{multline}
\label{eq:eq5}
 D_\mathrm{KL}[p(y|\mathbf{c_i},\mathbf{c_j}){\parallel} \hat{p}(y|\mathbf{c_i},\mathbf{c_j}, z_j{=}k)]{=}\\
 -\ln \hat{p}(y{=}1|\mathbf{c_i},\mathbf{c_j},z_j{=}k)\\\mathds{1}( \hat{p}(y{=}1|\mathbf{c_i},\mathbf{c_j},z_j{=}k){\neq}0).
\end{multline} 
In this equation $\mathds{1}( \hat{p}(y{=}1|\mathbf{c_i},z_j{=}k){\neq}0)$ ensures that only valid $k$s has been considered. We don't include this function in the next equations for simplicity but implicitly consider the condition it imposes on $k$. This scenario is depicted in \cref{fig:sfig1}.
Note that unlike \cite{oord2018representation}, we consider $\hat{p}(y)$ to be a Bernoulli distribution; this is because we only consider one sample at a time to see if this sample belongs to the same class or not. This is because, in their approach, categorical cross entropy forces the model to use one positive sample while considering all the remainder as negatives. But in the contrastive version, we can have multiple positives per each batch. Similar to \cite{oord2018representation}, we can write this probability as follows:
\begin{equation}
\label{eq:eqconp}
 \hat{p}(y{=}1|\mathbf{c_i}, \mathbf{c_j}, z_j{=}k){=}\frac{\frac{\hat{p}(z_j=k|\mathbf{c_i},\mathbf{c_j})}{\hat{p}(z_j=k)}}{\sum_{l=1}^K\frac{\hat{p}(z_j=l|\mathbf{c_i},\mathbf{c_j})}{\hat{p}(z_j=l)}} . 
\end{equation}
Using Bayes theorem, we will have the following:
\begin{equation}
\label{eq:eqzxj}
\hat{p}(z_j{=}k|\mathbf{c_j}){=}\frac{|z_j{=}k|}{|c_j|}\hat{p}(\mathbf{c_j}|z_j{=}k){=}\hat{p}(\mathbf{c_j}|z_j{=}k).
\end{equation}
We can rewrite \cref{eq:eq5} as:
\begin{equation}
\label{eq:eq6}
 D_\mathrm{KL}[p(y|\mathbf{c_i},\mathbf{c_j}){\parallel} \hat{p}(y|\mathbf{c_i},\mathbf{c_j},z_j)]{=}\overbrace{-\ln\frac{\hat{p}(z_j{=}k|\mathbf{c_j})}{\hat{p}(z_j{=}k)}}^A+\underbrace{\ln\sum_{l=1}^K\frac{\hat{p}(z_j{=}l|\mathbf{c_j})}{\hat{p}(z_j{=}l)}}_B .
\end{equation}
To calculate $A$, since in \cref{eq:eq5}, the condition is that $\hat{p}(y{=}1|\mathbf{c_i},\mathbf{c_j},z_j{=}k){\neq}0$, we can conclude that $\hat{p}(z_j{=}k|\mathbf{c_j}){>}0$. At least one sample of $\mathbf{c_j}$ or $\mathbf{c_i}$ should have $z_j{=}k$, or else this probability will be zero. So we have: $\hat{p}(z_j{=}k|\mathbf{c_j}){\geq}\frac{1}{|\mathbf{c_j}|}{=}\frac{K}{N}$. Therefore, we can find an upper bound for $A$:
\begin{equation}
\label{eq:eqA1}
\mathrm{A}{\leq}-\ln(\frac{K^2}{N}).
\end{equation}
To calculate $B$, 
\begin{equation}
\label{eq:eqB1}
\mathrm{B}{=}\ln({K}\sum_{l=1}^K\hat{p}(z_j{=}l|\mathbf{c_j})){=}\ln K.
\end{equation}
If we use equation \cref{eq:eqB1,eq:eqA1,eq:eq3} we have:
\begin{equation}
\label{eq:AplusB}
D_\mathrm{KL}[p(y|\mathbf{c_i},\mathbf{c_j}){\parallel}\hat{p}(y|\mathbf{c_i},\mathbf{x_j})]{\leq}\ln\frac{N}{K}.
\end{equation}



\subsubsection{Inter-class Divergence}
Here we have $\mathbf{c_j}{\neq}\mathbf{c_i}$, for this scenario which is depicted in \cref{fig:sfig2}, we have a few changes from the previous part, particularly \cref{eq:eq5} will be as:
\begin{multline}
\label{eq:2eq5}
 D_\mathrm{KL}[p(y|\mathbf{c_i},\mathbf{c_j}){\parallel} \hat{p}(y|\mathbf{c_i},\mathbf{c_j}, z_j{=}k)]{=}\\
 -\ln \hat{p}(y{=}0|\mathbf{c_i},\mathbf{c_j},z_j{=}k)\mathds{1}( \hat{p}(y{=}0|\mathbf{c_i},\mathbf{c_j},z_j{=}k){\neq}0).
\end{multline}
Since we can consider $\hat{p}(y{=}0|\mathbf{c_i},\mathbf{c_j},z_j{=}k){=}1{-}\hat{p}(y{=}1|\mathbf{c_i},\mathbf{c_j},z_j{=}k)$, \cref{eq:eq6} becomes
\begin{equation}
\label{eq:2eq6}
 D_\mathrm{KL}[p(y|\mathbf{c_i},\mathbf{c_j}){\parallel} \hat{p}(y|\mathbf{c_i},\mathbf{c_j},z_j)]{=}\overbrace{-\ln\sum_{l\neq k}\frac{\hat{p}(z_j{=}l|\mathbf{c_j})}{\hat{p}(z_j{=}l)}}^A+\underbrace{\ln\sum_{l=1}^K\frac{\hat{p}(z_j{=}l|\mathbf{c_j})}{\hat{p}(z_j{=}l)}}_B .
\end{equation}
To calculate $A$
\begin{equation}
\label{eq:2eqA1}
\mathrm{A}{=}-\ln(K\sum_{l{\neq}k}\hat{p}(z_j{=}l|\mathbf{c_j})){=}{-}\ln K{-}\ln(1-\hat{p}(z_j{=}l|\mathbf{c_j})).
\end{equation}
Again, since in \cref{eq:eq5}, the condition is that $\hat{p}(y{=}0|\mathbf{c_i},\mathbf{c_j},z_j{=}k){\neq}0$, we can conclude that $\hat{p}(z_j{=}k|\mathbf{c_j}){<}1$. At least one sample of $\mathbf{c_j}$ should have $z_j{=}k$, or else this probability will be zero. So we have: $\hat{p}(z_j{=}k|\mathbf{c_j}){\leq}(1{-}\frac{1}{|\mathbf{c_j}|}){=}1{-}\frac{K}{N}$. Therefore, we can find a similar upper bound for $A$:
\begin{equation}
\label{eq:2eqA2}
\mathrm{A}{\leq}{-}\ln(\frac{K^2}{N}).
\end{equation}
We have already calculated $B$ in \cref{eq:eqB1}. If we use equation \cref{eq:eqB1,eq:2eqA1,eq:eq3} we again have:
\begin{equation}
\label{eq:2AplusB}
D_\mathrm{KL}[p(y|\mathbf{c_i},\mathbf{c_j}){\parallel}\hat{p}(y|\mathbf{c_i},\mathbf{x_j})]{\leq}\ln\frac{N}{K}.
\end{equation}

Since in both cases, the upper bound holds the theorem is proven. $\square$
\vspace{3em}
\subsection{Unsupervised Contrastive Learning}
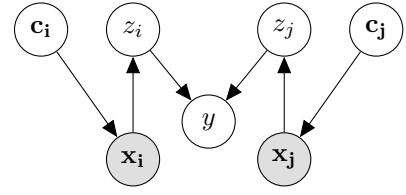
\begin{wrapfigure}{r}{0.45\textwidth}
  \centering
\vspace{-2em}
\begin{tikzpicture}
  \node[latent, xshift=0cm](y) {$y$};
    \node[latent, above=of y, xshift=-1cm, yshift=-0.5cm]  (zi) {$z_i$};
\node[latent, above=of y, xshift=1cm, yshift=-0.5cm]  (zj) {$z_j$};
  \node[obs, below=of zj, xshift=0cm,yshift=0cm] (xj) {$\mathbf{x_j}$};
  \node[obs, below=of zi, xshift=0cm,yshift=0cm]  (xi) {$\mathbf{x_i}$};
   \node[latent, right=of zj, xshift=-0.5cm]                               (cj) {$\mathbf{c_j}$};
      \node[latent, left=of zi, xshift=0.5cm]                               (ci) {$\mathbf{c_i}$};

  \edge {zi,zj} {y} ; %
\edge {ci} {xi} ; %
        \edge {xi} {zi} ; %
            \edge {xj} {zj} ; %
            \edge {cj} {xj} ; %

\end{tikzpicture}
\vspace{-0.8em}
\caption{\textbf{Bayesian Network for the Generalized Category Discovery.} Shaded nodes are observed variables $\mathbf{x_i}$, $\mathbf{x_j}$ corresponds to images $i$ and $j$, and $\mathbf{c_i}$ and $\mathbf{c_j}$ which are the ground-truth category variable. $z_i$ and $z_j$ are the latent category variables extracted from the model.}
\label{fig:baysunsup}
\vspace{-4em}
\end{wrapfigure}
For the unsupervised scenario, both labels are unknown, this means that we only have access to the inputs $\mathbf{x_i}$ and $\mathbf{x_j}$. This means that for unlabelled samples they will follow the Bayesian network shown in \cref{fig:baysunsup}. We can state a similar theorem for the unsupervised case as follows: 
\begin{theorem}
\label{theo:unsup}
Consider two samples $i$ and $j$ from the Bayesian network \cref{fig:bays}. If we have a total of $N$ samples and $K$ categories and the dataset is balanced, we will have the following upper bound if neither of $i$ and $j$ labels is known:
\begin{align}  
\label{eq:uppersup2}
    D_\mathrm{KL}[p(y|\mathbf{c_i},\mathbf{c_j})\parallel \hat{p}(y|\mathbf{x_i},\mathbf{x_j})]\leq 
    \ln\frac{N}{K}. 
\end{align}

\end{theorem}
\textit{Proof.} According to the Bayesian network shown in \cref{fig:baysunsup}, we have:
\begin{equation}
\label{eq:ueq1}
   \hat{p}(y|\mathbf{x_i},\mathbf{x_j})=\sum_{z_i=1}^K\hat{p}(z_i|\mathbf{x_i})\sum_{z_j=1}^K\hat{p}(z_j|\mathbf{x_j})\hat{p}(y|z_i,z_j).
\end{equation}
Note that since $\sum_{z_j=1}^K\hat{p}(z_j|\mathbf{x_j}){=}1$ and $\sum_{z_i=1}^K\hat{p}(z_i|\mathbf{x_i}){=}1$ and $p(y|\mathbf{c_i},\mathbf{c_j})$ is independent of $z_i$ and $z_j$, we can consider that
\begin{equation}
\label{eq:ueq2}
    p(y|\mathbf{c_i},\mathbf{c_j}){=}\sum_{z_i=1}^K \hat{p}(z_i|\mathbf{x_i})\sum_{z_j=1}^K\hat{p}(z_j|\mathbf{x_j})p(y|\mathbf{c_i},\mathbf{c_j}).
\end{equation} 
Also, since the KL divergence is convex, we can have the following inequality:
\begin{multline}
\label{eq:ueq3}
        D_\mathrm{KL}[\sum_{z_i=1}^K\hat{p}(z_i|\mathbf{x_i})\sum_{z_j=1}^K\hat{p}(z_j|\mathbf{x_j})p(y|\mathbf{c_i},\mathbf{c_j})
    {\parallel}\sum_{z_i=1}^K\hat{p}(z_i|\mathbf{x_i})\sum_{z_j=1}^K\hat{p}(z_j|\mathbf{x_j})\hat{p}(y|z_i,z_j)]\\\leq\sum_{z_i=1}^K\hat{p}(z_i|\mathbf{x_i})\sum_{z_j=1}^K\hat{p}(z_j|\mathbf{x_j})D_\mathrm{KL}[p(y|\mathbf{c_i},\mathbf{c_j}){\parallel} \hat{p}(y|z_i,z_j)].
\end{multline}
If we use equations \cref{eq:ueq1,eq:ueq2,eq:ueq3} on the left-hand side of the inequality \cref{eq:uppersup2}, we will have the following: 
\begin{align*}
    D_\mathrm{KL}[p(y|\mathbf{c_i},\mathbf{c_j}){\parallel}\hat{p}(y|\mathbf{x_i},\mathbf{x_j})]=&D_\mathrm{KL}[p(y|\mathbf{c_i},\mathbf{c_j})
    {\parallel}\sum_{z_i=1}^K\hat{p}(z_i|\mathbf{x_i})\sum_{z_j=1}^K\hat{p}(z_j|\mathbf{x_j})\hat{p}(y|z_i,z_j)]\\
    \leq&\sum_{z_i=1}^K\hat{p}(z_i|\mathbf{x_i})\sum_{z_j=1}^K\hat{p}(z_j|\mathbf{x_j})D_\mathrm{KL}[p(y|\mathbf{c_i},\mathbf{c_j}){\parallel} \hat{p}(y|z_i,z_j)].
\end{align*}
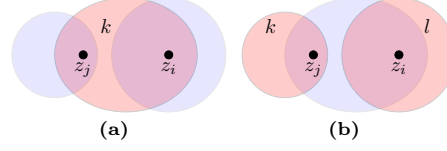
\begin{wrapfigure}{r}{0.5\textwidth}
\vspace{-0em}
  \centering
\begin{subfigure}{.24\textwidth}
  \centering
  \resizebox{1\linewidth}{!}{
\begin{tikzpicture}
  \draw[fill=red, opacity=0.2] (1.25,0) ellipse (1.25cm and 1cm);

\draw[fill=blue, opacity=0.1] (0,0) circle (0.75cm);
  \filldraw [black] (0.5,0) circle (2pt) node[anchor=north] {$z_j$};

\draw[fill=blue, opacity=0.1] (2,0) circle (1cm);
  \filldraw [black] (2,0) circle (2pt) node[anchor=north] {$z_i$};
  \node at (0.9,0.5) {$k$};
\end{tikzpicture}}
  \caption{}
  \label{fig:sfig3}
\end{subfigure}
\begin{subfigure}{.24\textwidth}
  \centering
  \resizebox{1\linewidth}{!}{
\begin{tikzpicture}
  \draw[fill=blue, opacity=0.1] (1.25,0) ellipse (1.25cm and 1cm);

\draw[fill=red, opacity=0.2] (0,0) circle (0.75cm);
  \filldraw [black] (0.5,0) circle (2pt) node[anchor=north] {$z_j$};

\draw[fill=red, opacity=0.2] (2,0) circle (1cm);
  \filldraw [black] (2,0) circle (2pt) node[anchor=north] {${z_i}$};
  \node at (-0.25,0.5) {$k$};
  \node at (2.5,0.5) {$l$};

\end{tikzpicture}}
  \caption{}
  \label{fig:sfig4}
\end{subfigure}
\vspace{-0.5em}
\caption{\textbf{Different scenarios for two sample category variables in unsupervised contrastive learning.} The latent variables $z_i$ and $z_j$ are extracted from the model. Model does the red assignments, while blue assignments indicate the ground truth category variable. For unsupervised contrastive learning, neither of the labels is known, so the model makes a mistake if (c) $i$ and $j$ have different labels and are assigned to the same label, (d) $i$ and $j$ have the same label, but model assign them to different labels.} 
\label{fig:appclustersunsup}
\vspace{-2em}
\end{wrapfigure}
Similar to supervised version for  $D_\mathrm{KL}[p(y|\mathbf{c_i},\mathbf{c_j}){\parallel} \hat{p}(y|z_i,z_j)]$, two possible scenarios can happen, i.e., $\mathbf{c_i}{=}\mathbf{c_j}$ and $\mathbf{c_i}{\neq}\mathbf{c_j}$. \cref{fig:appclustersunsup} depicts these different scenarios in which the model makes a mistake in assigning the label. We call these two scenarios intra-class divergence and inter-class divergence, respectively. 
Note that since the dataset is balanced, again we can consider that $|z_i|{=}|z_j|{=}|\mathbf{c_i}|{=}|\mathbf{c_j}|{=}\frac{N}{K}$. We also assume $\hat{p}(\mathbf{c_i}){\approx}p(\mathbf{c_i}){=}\hat{p}(z_i){=}\frac{1}{K}$.
In the transition from supervised to unsupervised scenario, the adaptation of the supervised derivations necessitates simply substituting $\mathbf{c_i}$ with $z_i$. For brevity, we omit the derivation of these adapted equations.

\subsection{Self-Expertise.} In the paper, we introduced an upper bound for a dataset comprising $N$ samples across $K$ categories, denoted as $\mathcal{S}_K$. To further refine our understanding, we propose an alternative upper bound, $\hat{\mathcal{S}}_K$, which leverages the hierarchical structure inherent among the categories. This hierarchical framework can be conceptualized as a Markov chain, where each category's relevance is determined independently of the samples and hyper-labels, except for its immediate predecessor and its children in the hierarchy. Our objective is to delineate the conditions under which $\hat{\mathcal{S}}_K$ serves as an effective upper bound, taking into account the hierarchical relationships among categories. This approach aims to capture the nuanced dependencies within the category structure, thereby providing a more granular and accurate upper limit for the distribution of samples among the categories.
\begin{equation}
\label{eq:sk}
\sum_{l{=}1}^{\lg K}D_\mathrm{KL}[p(y|\mathbf{c_i}^l,\mathbf{c_j}^l){\parallel}\hat{p}(y|\mathbf{c_i}^l,\mathbf{x_j},\mathbf{c_j}^{l{-}1})]{\leq}\hat{\mathcal{S}}_K,
\end{equation}
let $l$ denote the hierarchy level. Consider that at level $l$, given $K$ distinct categories, each cluster $\mathbf{c_j}^{l-1}$ contains $\frac{N}{K}$ samples, as opposed to the full $N$ samples without considering the hierarchy. Upon substituting these adjusted values for $K$ and $\frac{K}{2}$ into the established upper bound in \cref{eq:uppersup}, we obtain the following:
\begin{equation}
\label{eq:skequal}
\hat{\mathcal{S}}_K{=}\lg K \ln \frac{N}{K^2}, \quad \hat{\mathcal{S}}_{\frac{K}{2}}{=}(\lg K-1) \ln \frac{4N}{K^2}\Rightarrow \hat{\mathcal{S}}_{K}{\leq}\hat{\mathcal{S}}_{\frac{K}{2}}.\quad \qed
\end{equation}
Employing analogous reasoning for the unsupervised self-expertise component, we derive:
\begin{equation}
\label{eq:skequal}
 \hat{\mathcal{U}}_{K}{\leq}\hat{\mathcal{U}}_{\frac{K}{2}}.
\end{equation}




\clearpage
\vspace{-0.3em}
\section{Method}
\vspace{-0.8em}
For a deeper understanding of the proposed methods, this appendix section provides a detailed explanation and corresponding pseudocode for the algorithms briefly introduced in the main text.
\subsection{Balanced Semi-Superivsed K-means}
Our algorithm consists of three key stages: 
\begin{enumerate}
\item Semi-Supervised K-means Centers Initialization,
\item Updating novel categories while balancing clusters,
\item Semi-supervised K-means for final assignment based on the refined centers.
\end{enumerate}

\begin{algorithm}
\caption{Enhanced Semi-Supervised K-Means Initialization}
\label{alg:initial}
\begin{algorithmic}[1]
\STATE {\bfseries Input:} Labeled data points $x_i^l$ with labels $y_i^l$, unlabeled data points $x_j^u$, cluster size threshold $C$, counts of labeled categories $Y_c^l$ and unlabeled categories $Y_c^u$
\STATE {\bfseries Output:} Refined initial positions for cluster centroids
\STATE {\bfseries Procedure Initialization:} Initialize centroids $c_1,\cdots,c_k$ as the centroids of labeled data $x_i^l$ for each category $y_i^l=c_1,\cdots,c_k$. Exclude the $C$ nearest data points to each of these initial centroids to refine the cluster quality.
\REPEAT
\STATE Select a random data point $x_j$ from the pool of unlabeled data.
\STATE Exclude the $C$ nearest neighbors of $x_j$ to ensure diverse centroid selection.
\STATE Incorporate $x_j$ into the collection of centroids.
\UNTIL{Achievement of $K$ centroids}
\end{algorithmic}
\end{algorithm}

\begin{figure*}[h]
\begin{center}
\includegraphics[width=0.8\linewidth]{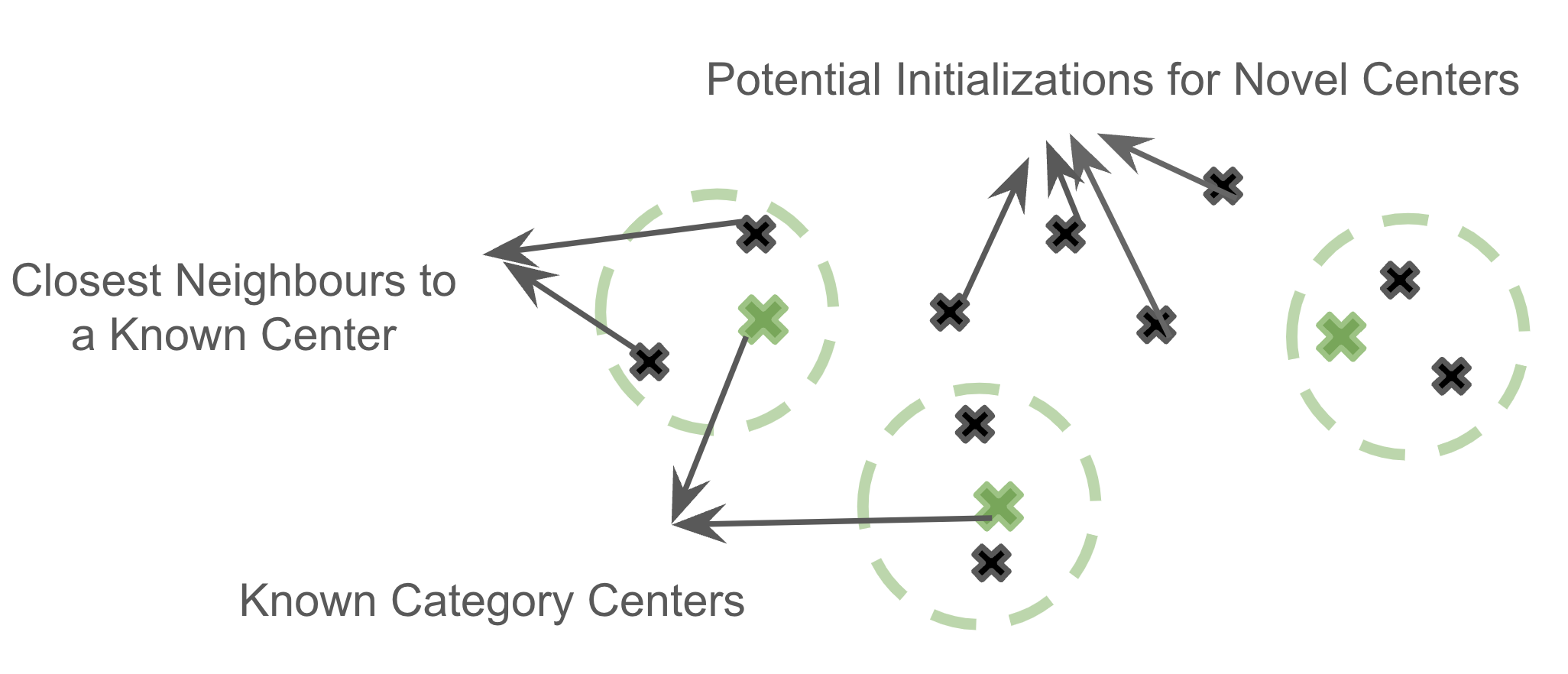}
\end{center}
\vspace{-1.5em}
\caption{\textbf{Cluster Center Initialization for Balanced Semi-Supervised K-Means.} Leveraging the advantage of having access to samples from known categories, we initiate the cluster centers for these categories directly using the available partial data from corresponding clusters (shown with green crosses). Given the predefined cluster size, denoted as $C$, it is prudent to avoid selecting initial centers for new categories from within the immediate vicinity of $C$ of these established centers. Furthermore, upon determining a center for each novel category, we deliberately exclude its $C$ closest samples in the dataset. This approach ensures that the initial centers for K-means clustering are adequately spaced apart, promoting a more balanced and effective clustering process. } 
\label{fig:initial}
\vspace{-1.5em}
\end{figure*}
\noindent\textbf{Semi-Supervised K-means Centers Initialization.} 
In our proposed method, Balanced Self-Supervised K-means (BSSK), the initial step involves the identification of K-means cluster centers for pre-labeled categories. This process is executed by calculating the cluster centers from the labeled data corresponding to the known categories. In scenarios involving balanced datasets characterized by a cluster size $C$, we enhance dataset purity by excluding the $C$ nearest samples surrounding each data point to mitigate cluster overlap and ensure distinct cluster formation. Subsequently, for the incorporation of novel categories into the learning framework, BSSK adopts a strategy of randomly selecting samples from the pool of remaining data to serve as new cluster centers. Following the selection of each novel cluster center, we apply a similar purification step by removing the $C$ closest samples in its vicinity. This systematic approach not only facilitates the exploration of new categories within the dataset but also maintains the integrity of cluster separation, ensuring a robust framework for the effective integration of both known and novel categories in self-supervised learning environments.

The pseudocode for this initialization is detailed in \cref{alg:initial}, with a practical illustration provided in \cref{fig:initial}. It is important to highlight that in scenarios involving unbalanced datasets, this phase of the process is bypassed. Instead, we adopt an initial cluster center selection mechanism akin to that used in Semi-Supervised K-means \cite{vaze2022generalized}. This adjustment ensures our method remains robust and adaptable to diverse dataset characteristics.

\begin{algorithm}[ht]
   \caption{Refinement of Novel Category Centers}
   \label{alg:MSSK}
   \begin{algorithmic}[1]
       \REQUIRE Defined cluster size $C$, Initial cluster centers $\mu^1, \ldots, \mu^K$
       \ENSURE Generation of pseudo-labels for unlabeled data
       \FOR{each iteration $iter = 1$ to $n_{iter}$}
           \STATE Assign each data point $x_j$ to its nearest cluster center.
           \IF{specified cluster size $C$ is not null}
               \STATE Ensure each cluster adheres to the defined size $C$ by applying balancing strategy from Algorithm~\ref{alg:Balance}.
           \ENDIF
           \STATE Recalculate the centers for the updated clusters.
       \ENDFOR
   \end{algorithmic}
\end{algorithm}
\begin{figure*}[t]
\begin{center}
\includegraphics[width=0.7\linewidth]{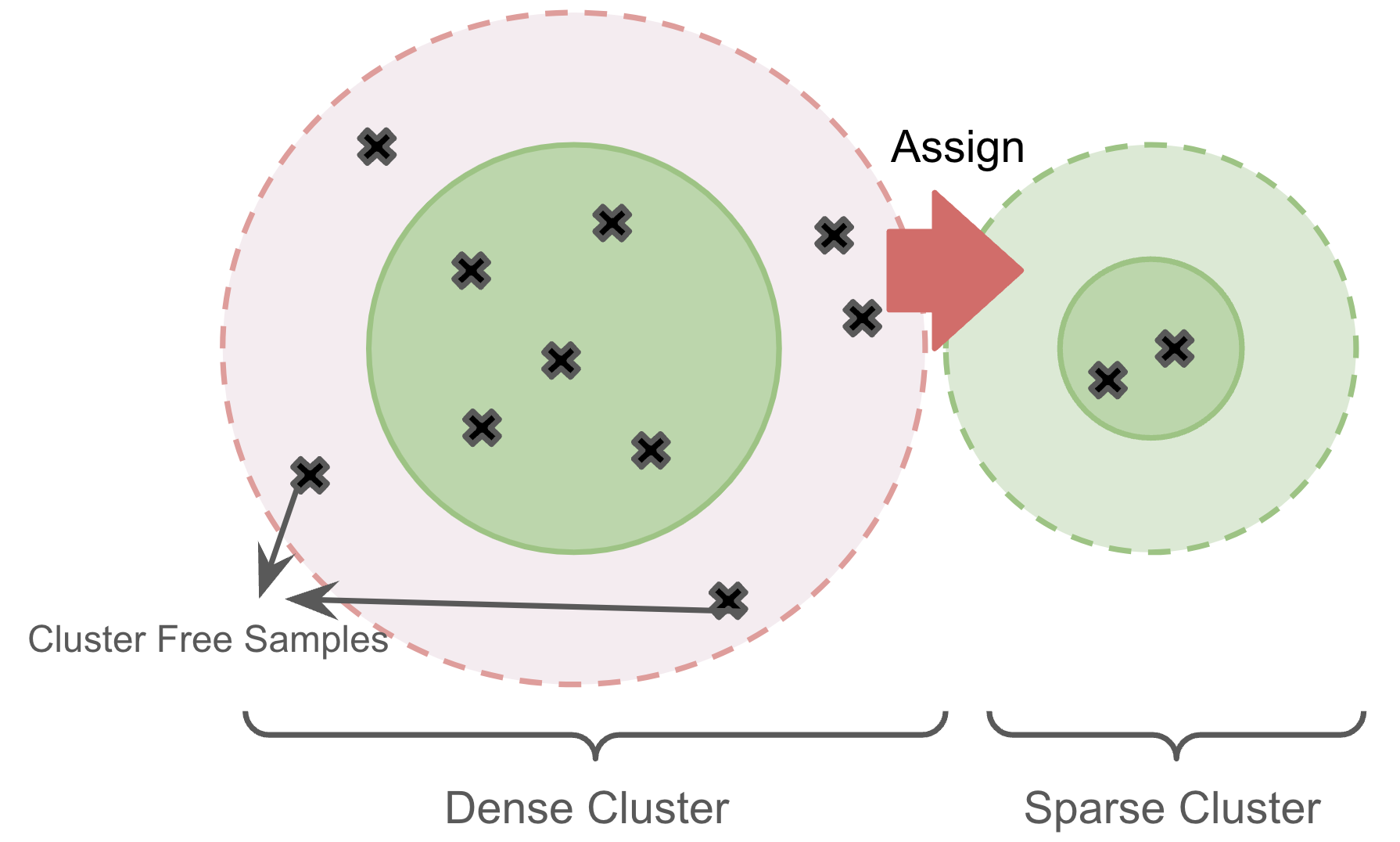}
\end{center}
\vspace{-1.5em}
\caption{\textbf{Balancing Cluster Allocation Through Stable Matching.} Following the assignment of clusters, it is anticipated that certain clusters will emerge as \textit{Dense}, possessing a number of samples exceeding the designated cluster size $C$, while others will manifest as \textit{sparse}, characterized by a sample count below $C$. A proposed strategy involves permitting the centers of Dense clusters to select their $C$ nearest samples, thereafter releasing any additional samples previously assigned to them as \textit{Cluster-free}. These \textit{Cluster-free} samples will then gravitate towards and select the nearest sparse cluster available to them, thereby balancing the allocation.} 
\label{fig:balance}
\vspace{-1.5em}
\end{figure*}

\begin{algorithm}[ht]
   \caption{Balanced Distribution of Category Clusters}
   \label{alg:Balance}
   \begin{algorithmic}[1]
       \REQUIRE Designated cluster size \(C\), Initial set of clusters \(c^1, \ldots, c^K\), Preliminary cluster assignments.
       \ENSURE Enhanced pseudo-labels for untagged data
        \FOR{\(k = 1\) \TO \(K\)}
           \STATE Arrange samples within cluster \(c^k\) in ascending order based on proximity to the central point \(\mu^k\).
           \STATE Determine the total sample count within cluster \(|c^k|\).
           \IF{\(|c^k| > C\)}
               \STATE Retain the \(C\) samples closest to \(\mu^k\) within \(c^k\), categorizing the remainder as \textit{cluster-free}.
           \ELSIF{\(|c^k| < C\)}
               \STATE Designate cluster \(c^k\) as \textit{sparse}.
           \ENDIF
        \ENDFOR
       \STATE Redirect all \textit{cluster-free} samples \(x_j\) towards the nearest \textit{sparse} cluster.
   \end{algorithmic}
\end{algorithm}
\noindent\textbf{Updating novel categories while balancing clusters.} 
In our proposed approach, we diverge from the conventional Semi-Supervised K-means \cite{vaze2022generalized} methodology, which updates all cluster centers simultaneously. Our premise is that the cluster centers corresponding to known data categories are already well-established and reliable. However, traditional K-means does not ensure the optimization of cluster centers for these known categories. To address this, our method selectively updates the cluster centers for novel categories while maintaining the integrity of the known category centers. This strategy ensures that, by the conclusion of this process, the centers for known categories are optimally positioned, whereas the centers for novel categories have been effectively adjusted. This technique is detailed in \cref{alg:MSSK}.

For balanced datasets, our method enforces uniformity in cluster size, setting $C{=}\frac{N}{K}$, through a stable matching algorithm. Initially, each sample is allocated to its nearest cluster. Subsequently, if a cluster receives assignments exceeding 
$C$, it retains only the 
$C$ closest samples, releasing the others. These unassigned samples are then considered "cluster-free." Concurrently, clusters with fewer than 
$C$ assignments are identified as sparse. Each cluster-free sample is then allocated to the nearest sparse cluster. This procedure ensures stability within the algorithm, guaranteeing that no sample-cluster pair would prefer an alternative arrangement to their current one. The pseudocode for achieving balance is outlined in \cref{alg:Balance}, with the corresponding matching process illustrated in Figure \cref{fig:balance}. The robustness and stability of our matching algorithm are further validated by a subsequent theorem, reinforcing the effectiveness of our approach within the framework of this study.
\begin{theorem}
\label{theo:bssk}
The balancing stage of BSSK is a stable matching.
\end{theorem}
\begin{proof}
Suppose, for the sake of contradiction, that our assumption is false. Therefore, we have two pairs of sample clusters, $(x_i,c_k)$ and $(x_j,c_l)$, which would prefer to be paired with each other but are currently assigned differently using the BSSK algorithm. This implies that $x_j$ is closer to $c_k$ than $x_i$ is, and since $c_k$ retains its closest assigned samples, it follows that either: 

\begin{enumerate}
    \item $x_j$ was never assigned to $c_k$ using K-means. 
    \item $x_j$ was assigned to $c_k$ by K-means but was later re-assigned to another cluster.
\end{enumerate}

In the first scenario, for the pairing to be considered unstable, $x_j$ would need to prefer $c_k$ over $c_l$. However, if this were the case, $x_j$ should have been assigned to $c_k$ instead of $c_l$, leading to a contradiction.

In the second scenario, if $x_j$ was re-assigned by BSSK, it suggests that $c_k$ was identified as a dense cluster. Consequently, $x_i$ cannot be re-assigned to it later as it is not part of a sparse cluster, and $x_j$ was not among the $C$ nearest neighbours of $c_k$ while $x_i$ was. This scenario implies that $c_k$ prefers $x_i$ over $x_j$, contradicting the initial claim of instability in the pairing.

\noindent Therefore, by contradiction, the theorem is proven. $\square$
\end{proof}

\begin{algorithm}[ht]
   \caption{Balanced Semi-Supervised Kmeans (BSSK)}
   \label{alg:BSSK}
   \begin{algorithmic}[1]
       \REQUIRE Cluster size $C$, Initial centers $\mu^1, \ldots, \mu^K$
       \ENSURE Pseudo-labels for unlabeled data
       \STATE Use algorithm \cref{alg:MSSK} to find good initial centers for novel categories. 
       \FOR{$iter = 1$ \TO $n_{iter}$}
           \STATE Assign each data point $x_j$ to the closest cluster center.
           \STATE Update centers of all clusters.
       \ENDFOR
   \end{algorithmic}
\end{algorithm}

\noindent\textbf{Final assignment using semi-supervised Kmeans.} 
In the culminating phase of our Balanced Semi-Supervised K-means (BSSK) methodology, we implement a traditional Semi-Supervised K-means algorithm, leveraging both known and novel clusters identified in earlier steps. This approach is motivated by two primary considerations. Firstly, the initial identification of moderately optimal novel and known cluster centers reduces the likelihood of K-means converging to suboptimal cluster centers, thereby enhancing the optimization process for both known and novel clusters. Secondly, it introduces a degree of flexibility regarding cluster size constraints. This adaptability ensures that even if the predefined cluster sizes are not precisely accurate or strictly enforced across the dataset, the model retains its ability to accurately assign samples. \cref{alg:BSSK} outlines the comprehensive framework of our balanced semi-supervised strategy, which is instrumental in generating pseudo-labels for the subsequent levels of category hierarchy.

\begin{algorithm}[h]
   \caption{Hierarchical Semi-Supervised Kmeans (HSSK)}
   \label{alg:HSSK}
\begin{algorithmic}[1]
   \REQUIRE Input data $X$, labeled samples $\mathcal{Y}_{\mathcal{S}}$, initial cluster size $C$
   \ENSURE Hierarchical Pseudo-labels for unlabeled data

   \STATE \textbf{Initialization:} Apply BSSK to obtain initial pseudo-labels $\hat{y}_j^1$ and cluster centers $\mu_1^1, \cdots, \mu_K^1$ for unlabeled data.
   \STATE Set hierarchy level $h \gets 1$.

   \WHILE{number of labeled prototypes $n_{l}/2 > 1$}
       \STATE Perform K-means clustering of labeled prototypes into $n_{l}/2$ groups.
       \STATE Assign hyperlabels $y^h$ to corresponding labeled data $y$ using K-means results.
       \STATE Update number of labeled prototypes: $n_{l} \gets n_{l}/2$.
       \STATE Update cluster size: $C \gets 2C$.
       \STATE Apply BSSK with updated $C$ and hyperlabels $y^h$ to obtain pseudo-labels and prototypes for the current level, resulting in $\mu^h$ and $\hat{y}^h$.
   \ENDWHILE
\end{algorithmic}
\end{algorithm}
\begin{figure*}[h]
\vspace{-2.5em}
\begin{center}
\includegraphics[width=0.7\linewidth]{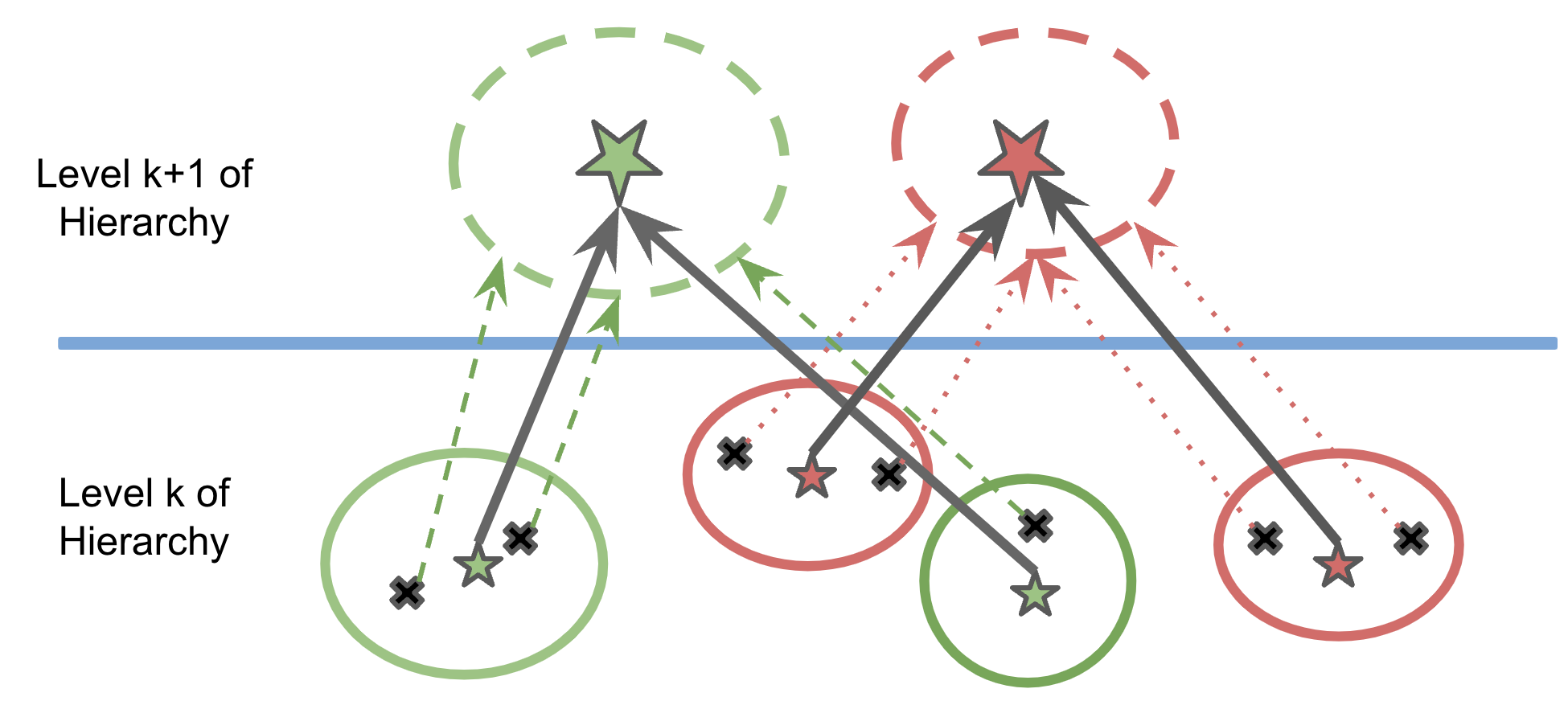}
\end{center}
\vspace{-1.5em}
\caption{\textbf{A single step of the Hierarchical Semi-Supervised K-Means (HSSK) algorithm.} Red circles represent novel category prototypes, and green circles denote known category prototypes, with the cluster centers for each indicated by a star symbol. Transitioning from level 
$k$ to $k{+}1$, HSSK initially divides the prototypes of known categories in half, and applies the same partitioning process to the prototypes of novel categories. Subsequently, each sample within these clusters is assigned to the corresponding hyper label associated with the cluster centers. This methodology is anticipated to compel the model to segregate the representations of novel and known categories, thereby enhancing its discriminative capability.} 
\label{fig:HSSK}
\vspace{-1.5em}
\end{figure*}
\subsection{Hierarchical Semi-Supervised K-means}
In our methodology, the initial step involves the generation of pseudo-labels through the application of the Balanced Semi-Supervised K-means (BSSK) algorithm. This foundational stage establishes our hierarchy's base level, ensuring that the generated pseudo-labels correspond in granularity to the existing ground truth categories. We achieve this by determining $K{=}|\mathcal{Y}_\mathcal{U}|$ clusters and assigning them centers $\mu_1^1, \cdots, \mu_K^1$, with the initial subset of these centers $\mu_1^1\cdots\mu_{|\mathcal{Y}_{\mathcal{S}}|}^1$ designated for the clusters associated with labeled data. The notation’s superindex~$1$ signifies this is the first level of abstraction.

Subsequently, for each subsequent level $k$, we conduct a twofold operation: halving the cluster centers of known prototypes from the previous level and applying the same halving procedure to cluster centers representing novel categories. We then reassign all samples linked to clusters at the $(k{-}1)^{th}$ level to the newly formed clusters at level $k$, based on their hyperlabel associations. This iterative process is carried out until we delineate the most abstract dichotomy between `known' and `novel' categories. In scenarios involving balanced datasets, it is also pertinent to note that clusters at level $k$ are designed to be double the size of those at level $k-1$, thus maintaining a consistent growth pattern across levels. For each iteration of the BSSK algorithm corresponding to a specific level $k$, we tailor the algorithm to accommodate the clustering characteristics of that level.

The procedural details of this Hierarchical Semi-Supervised K-means (HSSK) approach are succinctly described in our pseudocode, referenced in \cref{alg:HSSK}. Additionally, we provide a visual representation of the HSSK mechanism, illustrated in \cref{fig:HSSK}, to show a single step of hierarchical structuring.
\begin{figure*}[h]
\begin{center}
\includegraphics[width=\linewidth]{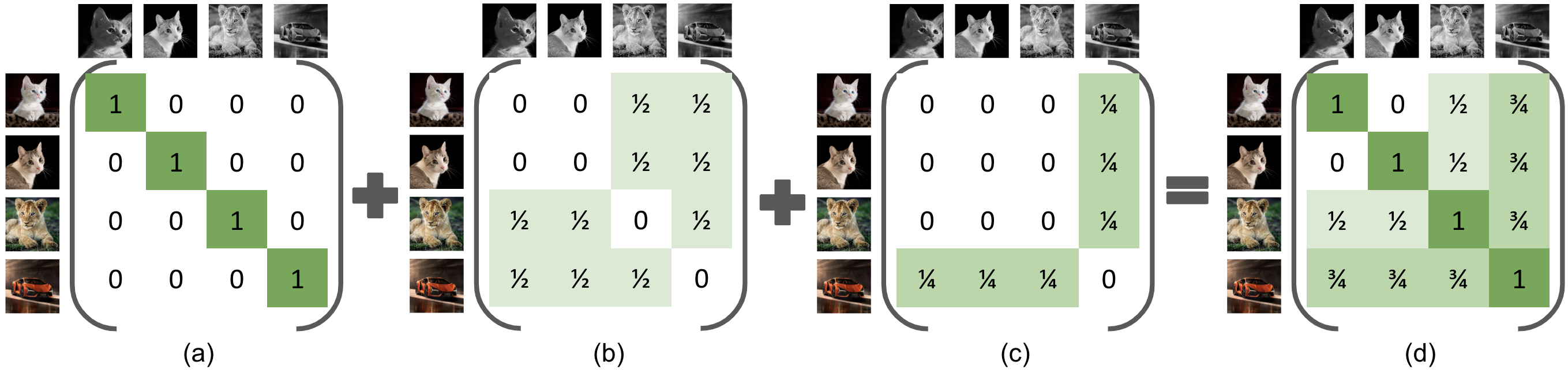}
\end{center}
\vspace{-1.5em}
\caption{\textbf{Construction Process of the Unsupervised Self-Expertise Target Matrix.} \textit{(a)} Each sample is uniquely identified, ensuring distinctiveness. \textit{(b)} Non-feline samples are associated with an uncertainty level of $0.5$, allowing for moderate differentiation ambiguity. \textit{(c)} Non animal samples are differentiated with an additional uncertainty of $0.25$, ensuring more focus on distinguishing among animal categories. \textit{(d)} The final aggregated matrix, subject to normalization, emphasizes high confidence in distinguishing between individual cats, while permitting greater uncertainty for non-animal entities, such as cars.} 
\label{fig:exp}
\vspace{-0.5em}
\end{figure*}

\begin{algorithm}[ht]
\caption{Hierarchical Label Smoothing}
\label{alg:HLS}
\begin{algorithmic}[1]
\REQUIRE Unlabeled data samples $x_j^u$, desired cluster size $C$, and cluster centers $\mu^1, \ldots, \mu^K$
\ENSURE Refined smoothed target matrix $Y$.
\STATE \textbf{Initialization:}
\FOR{each cluster $k = 1$ to $K$}
\STATE Determine cluster radius $r^k$ for cluster $k$
\ENDFOR
\STATE Compute pairwise distances $d_{ij}$ between data samples
\STATE Calculate label smoothing weights $y_{ij}$ using: $y_{ij}{=} \sum_k \mathds{1}(d_{ij} > r^k) / 2^k$ and $y_{ii}{=}1$
\STATE Normalize the $Y$ matrix to ensure validity.
\end{algorithmic}
\end{algorithm}

\subsection{Unsupervised Self-Expertise}
In the main text, we outlined the development of an unsupervised self-expertise target matrix designed to prioritize the model's attention on the most challenging negative samples. The pseudocode for extracting these hard negatives via a Label Smoothing target matrix is detailed in \cref{alg:HLS}. Additionally, the incremental formation of this matrix is visually depicted in \cref{fig:exp}.

\begin{figure*}[ht]
\begin{center}
\includegraphics[width=\linewidth]{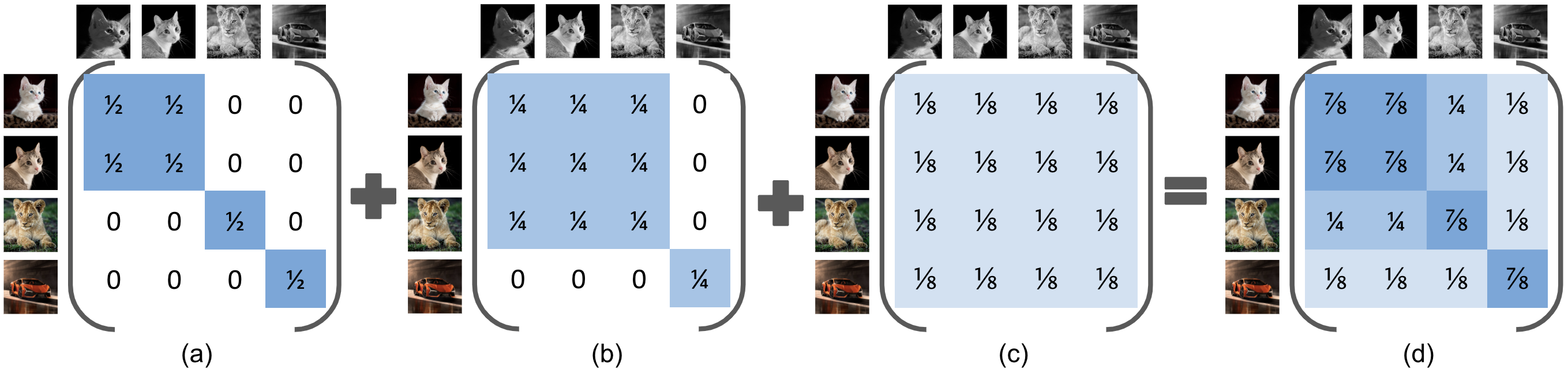}
\end{center}
\vspace{-1.5em}
\caption{\textbf{Construction Process of the Supervised Self-Expertise Target Matrix.} \textit{(a)} In the matrix, same-category pseudo-labels are treated as partially positive $(0.5)$ with all others marked negative, targeting the first half of the latent representation. \textit{(b)} Feline samples are assigned a positivity level of 
$0.25$, aimed at enhancing their similarity detection, affecting only the first quarter of latent dimensions. \textit{(c)} A distinction is made for all `known' samples with an added uncertainty level of 
$0.125$, concentrating on differentiating between known and unknown samples, impacting one-eighth of the latent space. \textit{(d)} The comprehensive matrix, applied to the initial one-eighth of latent dimensions and normalized, focuses on enhancing category generalization.} 
\label{fig:supex}
\vspace{-1.5em}
\end{figure*}

\noindent  \textbf{Intuition of Unsupervised Self-Expertise.}
Considering visually similar samples within the same cluster as negatives can seem counter-intuitive at first. However, consider two visually similar yet semantically distinct samples, such as `sparrow' and `finch,' misclassified under the same category of `sparrow.' Our method actively differentiates `finch' from `sparrow' despite their visual resemblance. Note that for fine-grained datasets with many categories, the majority of visually similar samples do not share the same semantic categories. By treating the visually similar samples as harder negatives, we make our unsupervised expertise focus on these samples while supervised self-expertise does the more coarse distinctions at the category level. 
\subsection{Supervised Self-Expertise}
In alignment with the paradigm of unsupervised self-expertise, our approach necessitated the construction of a target matrix specific to supervised self-expertise, designed to encapsulate both weakly positive instances and strongly negative examples. Figure \ref{fig:supex} elucidates the stepwise development of this matrix across various strata of the category hierarchy.

\noindent \textbf{Justification of  Latent Utilization.}
The rationale behind allocating the first half of the latent dimension to more abstract categories is twofold. Firstly, this allocation allows the model to make a rough categorization \eg `feline' or `canine'. Subsequently, the differentiation between precise categories \eg `cat' and `lion' is handled by the second half. 
Utilizing the full dimension to simultaneously repel `cat' and `lion' samples, while attracting them at a higher abstraction level, impedes the model's capacity to distinguish between these categories. This is primarily because the dominance of `feline' samples obscures the finer distinctions between `cat' and `lion'. By segregating `cat' and `feline' expertise, we ensure that a segment of the latent dimension is exclusively dedicated to discerning between finer-grained categories.

\clearpage
\section{Experiments}

\subsection{Experimental Setup}

{
\begin{table}[t]
\caption{\textbf{Statistics of datasets and their data splits for the generalized category discovery task.} The first four datasets are fine-grained image classification datasets, while the next three are coarse-grained datasets. The Herbarium19 dataset is both fine-grained and long-tailed.}
\vspace{-2em}
\label{tab:data}
\begin{center}
\begin{sc}
\resizebox{0.9\linewidth}{!}{
\begin{tabular}{lrrrr}
\toprule
& \multicolumn{2}{c}{\textbf{Labelled}} & \multicolumn{2}{c}{\textbf{Unlabelled}}\\ \cmidrule(lr){2-3} \cmidrule(lr){4-5}
\textbf{Dataset}  & \#Images & \#Categories   &  \#Images & \#Categories  \\
\midrule
CUB-200 \cite{wah_branson_welinder_perona_belongie_2011} & 1.5K & 100 & 4.5K &200 \\
FGVC-Aircraft \cite{maji2013fine}& 3.4K& 50& 6.6K& 100\\
Stanford-Cars \cite{krause20133d}& 2.0K & 98 & 6.1K &196 \\ 
Oxford-Pet \cite{parkhi2012cats}& 0.9K& 19& 2.7K& 37\\
\midrule
CIFAR-10 \cite{krizhevsky2009learning}       & 12.5K & 5 & 37.5K & 10   \\  
CIFAR-100 \cite{krizhevsky2009learning}       & 20.0K & 80 & 30.0K & 100 \\ 
ImageNet-100 \cite{deng2009imagenet}& 31.9K& 50& 95.3K& 100\\
\midrule
Herbarium19 \cite{tan2019herbarium}& 8.9K& 341& 25.4K& 683\\

\bottomrule
\end{tabular}}
\end{sc}
\end{center}
\vspace{-2em}
\end{table}
}

\textbf{Datasets.}
In this work, we evaluate the performance of our proposed method across several datasets to demonstrate its effectiveness and versatility. We conduct experiments on four detailed datasets, namely Caltech-UCSD Birds-200-2011 (CUB-200)\cite{wah_branson_welinder_perona_belongie_2011}, Fine-Grained Visual Classification of Aircraft (FGVC-Aircraft)\cite{maji2013fine}, Stanford Cars~\cite{krause20133d}, and Oxford-IIIT Pet~\cite{parkhi2012cats}, to assess its efficacy in fine-grained image recognition tasks. These tasks require the identification of subtle distinctions among highly similar categories, such as different bird species, aircraft models, car brands, and pet breeds, thereby posing significant challenges for image recognition systems. Furthermore, we extend our evaluation to more general coarse-grained datasets, including CIFAR10, CIFAR100~\cite{krizhevsky2009learning}, and a subset of ImageNet comprising 100 categories, referred to as ImageNet-100~\cite{deng2009imagenet}. This extension showcases the adaptability of our method to broader classification challenges beyond fine-grained tasks. CIFAR10 and CIFAR100 feature coarse-grained categories encompassing a wide range of objects, such as vehicles and animals, while ImageNet-100 offers a diverse set of general categories from the larger ImageNet dataset. To provide a comprehensive overview of our experiments, we include a detailed summary of the datasets' characteristics, including their statistical distribution and train/test splits, in \cref{tab:data}.

\noindent\textbf{CUB-200}~\cite{wah_branson_welinder_perona_belongie_2011}, or Caltech-UCSD Birds-200-2011, emphasizes the importance of discerning minute details across different bird species, underlining the complexity of fine-grained image recognition tasks.

\noindent\textbf{FGVC-Aircraft}~\cite{maji2013fine} challenges image recognition models with its focus on aircraft, where variations in design significantly affect structural appearances, highlighting the dataset's fine-grained nature.

\noindent\textbf{Stanford Cars}~\cite{krause20133d} introduces the complexity of recognizing car brands from various angles and in different colors, showcasing the nuanced challenges within fine-grained datasets.

\noindent\textbf{Oxford-IIIT Pet}~\cite{parkhi2012cats} centers on the classification of cat and dog breeds, where limited data availability increases the risk of overfitting, stressing the dataset's fine-grained and challenging aspects.

\noindent\textbf{CIFAR10/100}~\cite{krizhevsky2009learning} are broad-spectrum datasets that encompass a variety of general object categories, including vehicles and animals, illustrating the diversity of challenges in coarse-grained image classification.

\noindent\textbf{ImageNet-100}~\cite{deng2009imagenet} serves as a representative subset of the extensive ImageNet database, focusing on 100 varied categories to evaluate the generalizability of our approach in a broader context.

\noindent\textbf{Herbarium-19}~\cite{tan2019herbarium}, although not mentioned in the main text, could be considered in further discussions as it represents a unique dataset focused on plant species and is long-tailed, illustrating the application of fine-grained classification in when categories are unbalanced. Its inclusion could enrich the diversity of datasets evaluated in the paper, emphasizing the method's applicability across different domains.

\noindent\textbf{Implementation Details.}
In our experiments, we adhered to the dataset division proposed by Vaze \etal \cite{vaze2022generalized}, where half of the categories in each dataset are designated as known, except for CIFAR100, where 80\% are reused as known categories. The labeled set consists of 50\% of the samples from these known categories. The remainder of the known category data, along with all data from novel categories, comprise the unlabeled set. 
Following \cite{vaze2022generalized}; we use ViT-B/16 as our backbone, which is either pre-trained by DINOv1 \cite{caron2021emerging} on unlabelled ImageNet~1K \cite{krizhevsky2017imagenet}, or pretrained by DINOv2 \cite{oquab2024dinov} on unlabelled ImageNet~22K. We use the batch size of $128$ for training and set $\lambda{=}0.35$. For label smoothing, we use $\alpha{=}0.5$ for fine-grained datasets and $\alpha{=}0.1$ for coarse-grained datasets. 
Different from \cite{vaze2022generalized}, we froze the first 10 blocks of ViT-B/16 and fine-tuned the last two blocks instead of only the last one to have more parameters given that for each level, only a fraction of the latent dimension is considered.\footnote{Our code is available at: \url{https://github.com/SarahRastegar/SelEx}} 

\noindent\textbf{Evaluation Metrics.}
We use our proposed Balanced Semi-Supervised K-means (BSSK) to cluster the extracted embeddings. Then, we use the Hungarian algorithm \cite{wright1990speeding} to solve the optimal assignment of emerged clusters to their ground truth labels. We report the accuracy of the model's predictions on \textit{All}, \textit{Known}, and \textit{Novel} categories. Accuracy on \textit{All} is calculated using the whole unlabelled test set, consisting of known and unknown categories. For \textit{Known}, we consider the samples with labels known during training. Finally, for \textit{Novel}, we consider samples from the unlabelled categories at train time.

\subsection{Additional Empirical Results.}

\noindent\textbf{Aggregated Performance on Different Kinds of Datasets.} 
In this section, we present a comparative analysis of our method against other approaches, as summarized in Table \ref{tab:aggr}. The table consolidates the performance metrics across different datasets. It is evident that our method not only demonstrates respectable performance across both kinds of datasets but also shows superior performance in fine-grained datasets. Furthermore, when considering the aggregated performance metrics, it becomes clear that our method surpasses the competing methods, underscoring its effectiveness and robustness in a variety of settings.
\begin{table}[ht!]
\vspace{-2em}
  \caption{\textbf{Comparison with state-of-the-art for Aggregated datasets} Bold numbers show the best accuracies. Our method has a consistent performance for the three experimental settings (\textit{All}, \textit{Known}, \textit{Novel}) on fine and coarse-grained datasets.}
  \vspace{-0.5em}
  \label{tab:aggr}
  \centering
  \resizebox{0.9\linewidth}{!}{
\begin{tabular}{c|l|ccc|ccc|ccc}
\toprule
&& \multicolumn{3}{c}{\textbf{Fine-grained}} & \multicolumn{3}{c}{\textbf{Coarse-grained}}& \multicolumn{3}{c}{\textbf{Average}}\\ \cmidrule(lr){3-5} \cmidrule(lr){6-8}\cmidrule(lr){9-11}
&\textbf{Method}  & All & Known  & Novel &  All & Known  & Novel&All & Known  & Novel\\
\midrule
\multirow{8}{*}{\rotatebox{90}{DINOv1}}&GPC \cite{zhao2023learning} &44.5&51.7&39.9& 80.4 &91.9& 71.3&62.5&71.8&55.6\\
&GCD \cite{vaze2022generalized} &  45.1&51.8&41.8& 79.5 & 88.0  &73.7&62.3&69.9&57.8\\

&XCon\cite{fei2022xcon} &46.8 &52.5&44.0&82.6&90.7&75.1&64.7&71.6&59.6\\
&PromptCAL \cite{zhang2022promptcal}&55.1&62.2&51.7 & \bf{87.4} & 91.2 & 84.0&71.3&76.7&67.9 \\
&SimGCD \cite{wen2022simple}       & 56.1&65.5&51.5 &86.7 & 89.8& \underline{84.6}&71.4&77.7&68.1\\

&GCA~\cite{otholt2024guided}&58.4&67.5&54.0&86.9&\underline{91.9}&82.7&\underline{72.7}&79.7&68.4\\
&AMEND \cite{banerjee2024amend}&58.0&70.2&52.0&87.0&89.1&\bf{86.5}&72.5&79.7&\underline{69.3}\\
&InfoSieve \cite{rastegar2023learn}                 &\underline{60.5}&\bf{72.1}&\underline{54.7}& 84.5& 91.2 &79.2&72.5&\underline{81.7}&67.0\\
\midrule
\rowcolor{gray!25}&\textbf{SelEx (Ours)}         &\bf{63.0}&\underline{71.9}&\bf{58.8}&\underline{87.1}& \bf{92.3}&83.0
&\bf{75.1}&\bf{82.1}&\bf{70.9}\\
\bottomrule
\end{tabular}
}
\vspace{-1.5em}
\end{table}

\begin{wraptable}{r}{0.45\textwidth}
\vspace{-3.5em}
{
    \centering

  \caption{\textbf{Comparison with state-of-the-art for long-tailed image classification on \textbf{Herbarium-19} dataset.} Bold and underlined numbers, respectively, show the best and second-best accuracies. Our method improves the baseline GCD \cite{vaze2022generalized} for the three experimental settings (\textit{All}, \textit{Known}, and \textit{Novel}). This table shows that our method is also suited for long-tailed settings with unbalanced categories.}
  \label{tab:herb}
  \resizebox{\linewidth}{!}{

\begin{tabular}{c|l|ccc}
\toprule
&\textbf{Method}  &All & Known  & Novel \\
\midrule
\multirow{6}{*}{\rotatebox{90}{DINOv1}}&GCD~\cite{vaze2022generalized}	&35.4&51.0&27.0\\
&CMS~\cite{choi2024contrastive}&36.4&54.9&26.4\\
&InfoSieve~\cite{rastegar2023learn} &41.0 &55.4 &33.2\\
&PIM~\cite{chiaroni2023parametric}&42.3&56.1&34.8\\


&SPTNet~\cite{wang2024sptnet}&43.4&58.7&35.2\\
&SimGCD~\cite{wen2022simple}&44.0&58.0&\underline{36.4}\\

&AMEND~\cite{banerjee2024amend}&\underline{44.2}&\underline{60.5}&35.4\\
&$\mu$GCD~\cite{vaze2023clevr4}&\bf{45.8}&\bf{61.9}&\bf{37.2}\\
\midrule
\rowcolor{gray!25}&\textbf{SelEx (Ours)} &39.6& 54.9 &31.3
\\        \bottomrule
\end{tabular}
}
\vspace{-3em}
}
\end{wraptable}

\noindent\textbf{Long-tailed Datasets.} 
In this section, we present \cref{tab:herb} which outlines the efficacy of our proposed method when applied to the Herbarium dataset, known for its long-tailed distribution and fine-grained categorization. Numerous real-world datasets demonstrate a power-law distribution. By simply omitting our algorithm's cluster size balancing component, we can apply it to long-tailed scenarios. Despite our model's initial assumption of category uniformity, using this trick, the results depicted in the table reveal that it achieves respectable performance on this challenging dataset. This table shows enhancements over the baseline GCD method, with gains of~4.2\% overall,~3.9\% in known, and~4.3\% in novel categories, surpassing the baseline method GCD. This observation suggests that self-expertise is advantageous not only for uniform category distributions but also for non-uniform ones.

\subsection{Ablative Studies}

\begin{figure*}[t!]
  \centering
  \resizebox{0.6\textwidth}{!}{
   \centering
  \includegraphics[width=0.9\linewidth]{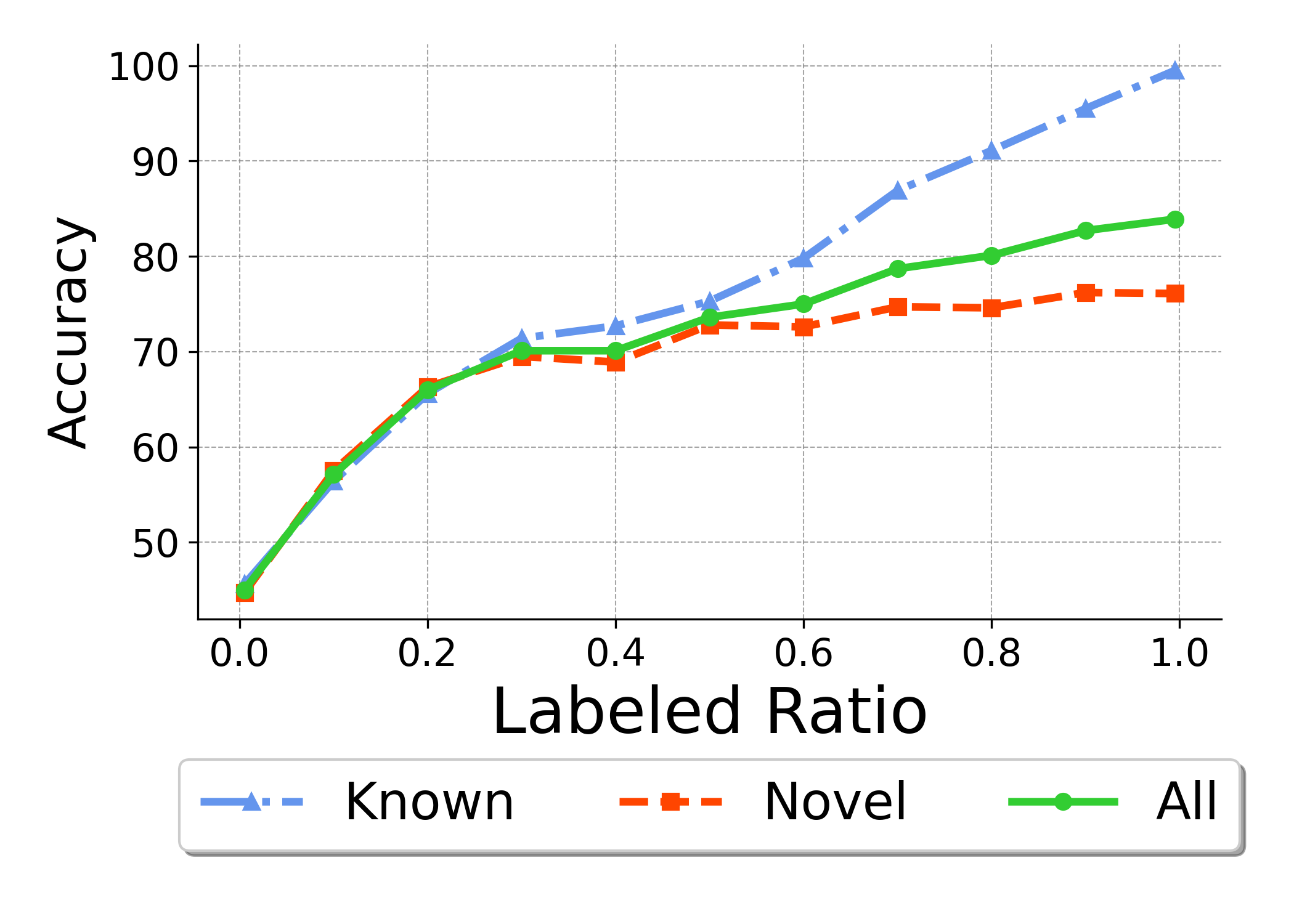}
  }
\vspace{-1.5em}
\caption{\textbf{Impact of varying labeled sample ratios for known categories.} Illustration of the model's proficiency in identifying novel categories across a spectrum of labeled data ratios for known categories. Showcases the model's consistent performance in discovering novel categories, even at a 0\% labeling ratio—akin to unsupervised clustering—versus the optimal 100\% labeling scenario, which transforms the task into one of novel class discovery. }
\label{fig:ratio}
\end{figure*}

\noindent\textbf{Impact of varying labeled sample ratios for known categories.} 
In this section, we explore the impact of varying the proportion of labeled samples from known categories during training on model performance. In the primary experiments detailed in the main body of the text, labels were available for 50\% of the samples from known categories. As illustrated in \cref{fig:ratio}, we investigate the consequences of altering the ratio of labeled samples, ranging from nearly 0\%—which essentially positions the model as an unsupervised clustering approach—to 100\%, where the problem can be posed as novel class discovery. Notably, we observe an anticipated enhancement in performance for known categories as the volume of labeled samples increases. Intriguingly, the performance on novel categories remains robust, achieving a score of 68 with only a 20\% labeling ratio, which is on par with the 50\% ratio utilized in our main experiments. This finding indicates that our model is capable of identifying novel categories effectively, even with a significantly reduced number of labeled samples for known categories.

\begin{figure*}[t!]
  \centering
  \resizebox{0.6\textwidth}{!}{
   \centering
  \includegraphics[width=0.9\linewidth]{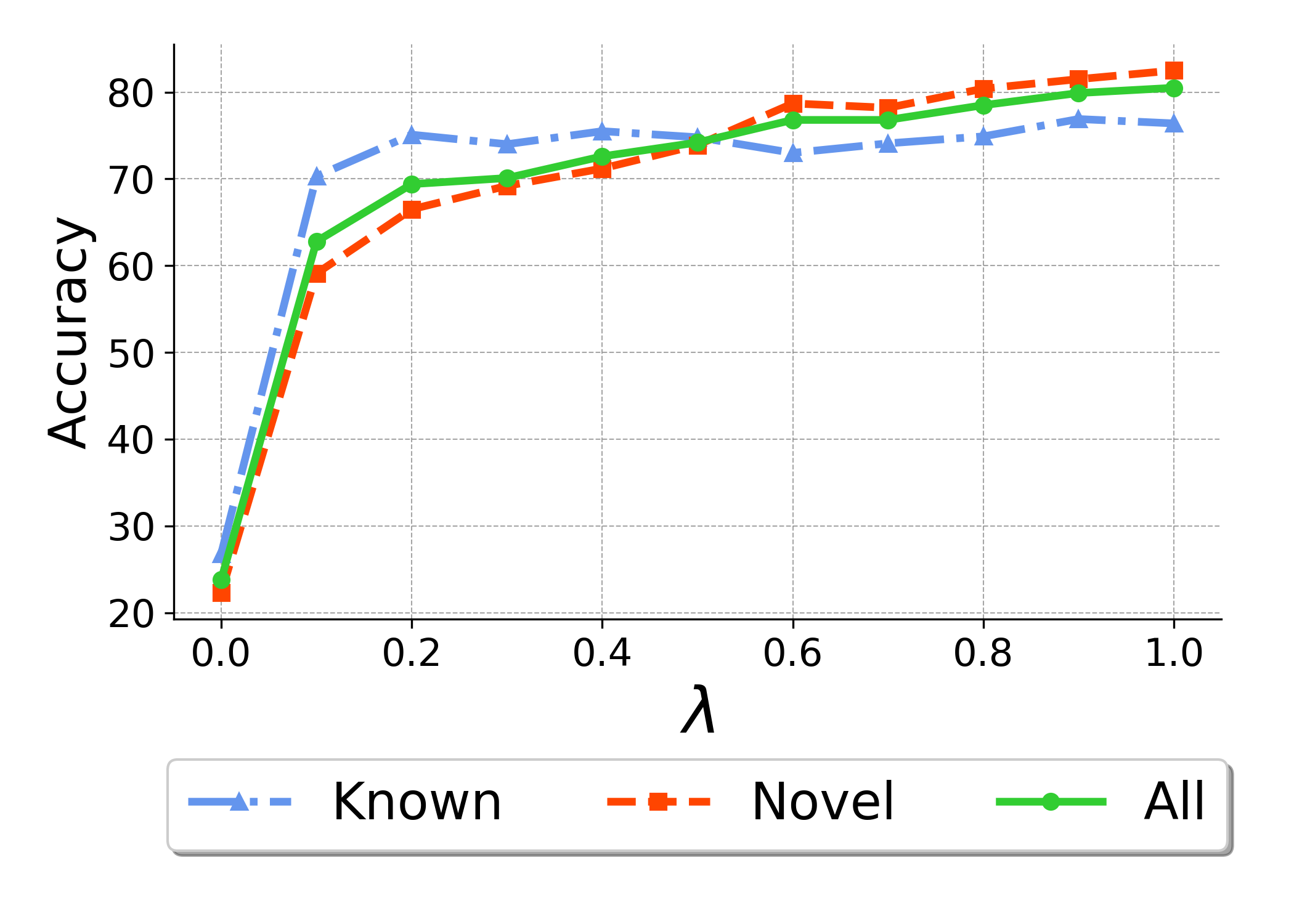}
  }
\vspace{-1.5em}
\caption{\textbf{Impact of varying $\lambda$ values on performance.} highlighting the balance between supervised and unsupervised learning for novel category recognition. Demonstrates increased performance in novel categories with higher $\lambda$, and discusses the plateau in known category performance beyond a $\lambda$ of $0.2$, due to pseudo-label noise and diminished unsupervised learning effectiveness. This emphasizes the importance of optimizing $\lambda$ to enhance novel category detection while managing the limitations posed by noise in pseudo-labels and unsupervised learning approaches.} 
\label{fig:lambda}
\end{figure*}

\noindent\textbf{Impact of the Supervised and Unsupervised Mixing Hyperparameter $\lambda$.} 
In this section, we modulate the mix between supervised learning and unsupervised self-expertise by varying the parameter $\lambda$, assessing its impact on model performance across different levels. The findings are summarized in \cref{fig:lambda}. An overarching observation is that an increase in $\lambda$ generally enhances performance, particularly in the context of novel category recognition. This improvement is attributed to our utilization of pseudo-labels for the supervision of novel categories, contrasting with the Generalized Category Discovery (GCD), which prioritizes optimization for known category classification, often at the expense of overall performance. The Self-Expertise (SelEx) approach, in contrast, leverages pseudo-labels for novel category supervision, leading to superior performance in these categories as $\lambda$ increases. However, beyond a $\lambda$ value of $0.2$, there is a negligible impact on the performance of known categories, with a slight decline observed. This can likely be attributed to the inherent noise in pseudo-labels, which adversely affects known category accuracy. Furthermore, with a higher $\lambda$, the effectiveness of unsupervised self-expertise diminishes, indicating a potential compromise in the model's ability to differentiate between semantically distinct yet visually similar samples.

\begin{figure*}[t!]
\begin{subfigure}{.5\textwidth}
  \centering
  \resizebox{\textwidth}{!}{
   \centering
  \includegraphics[width=0.9\linewidth]{Images/Accuracy_clustering.png}
  }
  \caption{CUB}
  \label{fig:figalpha0}
\end{subfigure}
\begin{subfigure}{.5\textwidth}
  \centering
  \resizebox{\textwidth}{!}{
   \centering
  \includegraphics[width=0.9\linewidth]{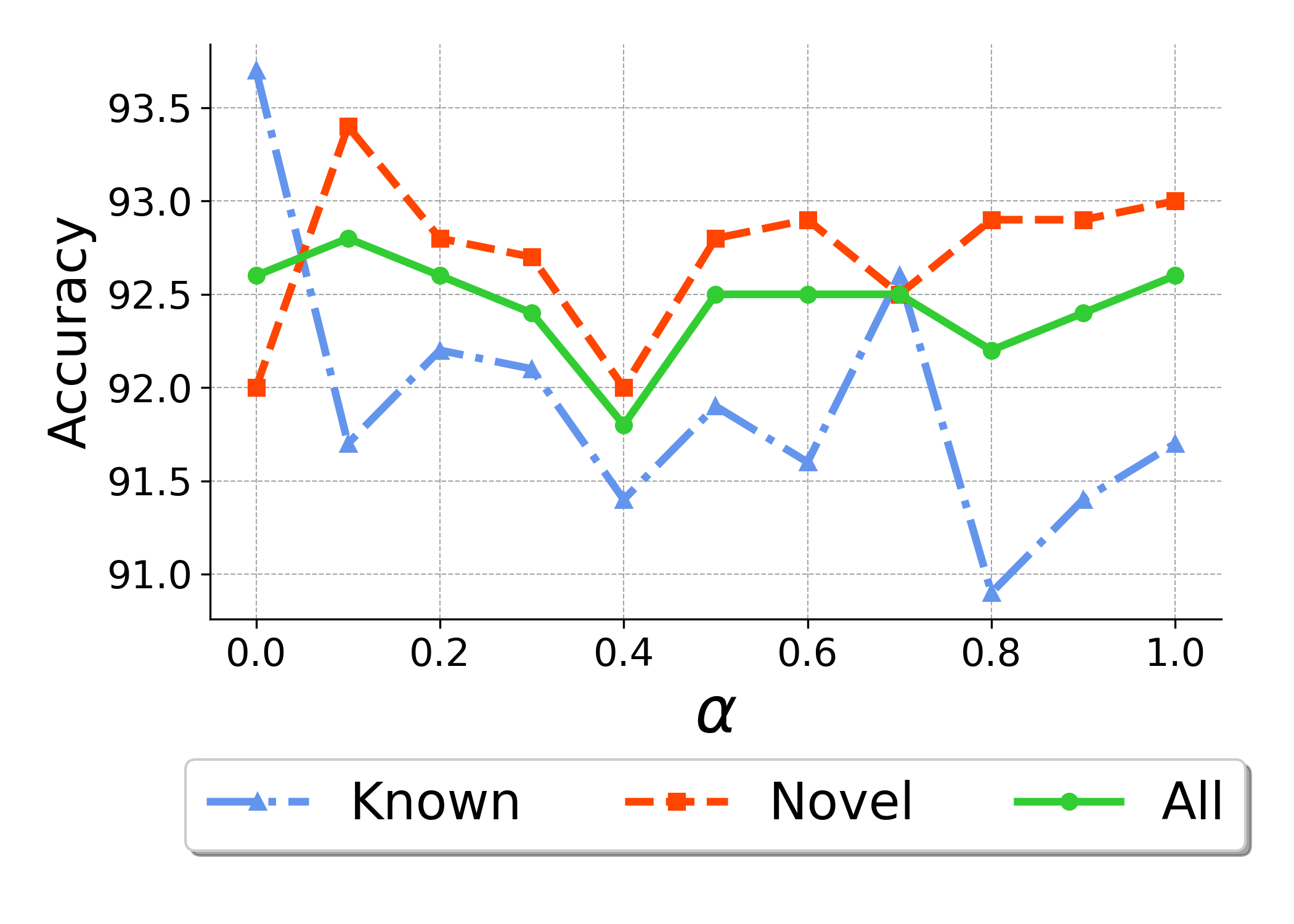}
  }
  \caption{Oxford-IIIT Pet}
  \label{fig:figalpha1}
\end{subfigure}\\
\begin{subfigure}{.5\textwidth}
  \centering
  \resizebox{\textwidth}{!}{
   \centering
  \includegraphics[width=0.9\linewidth]{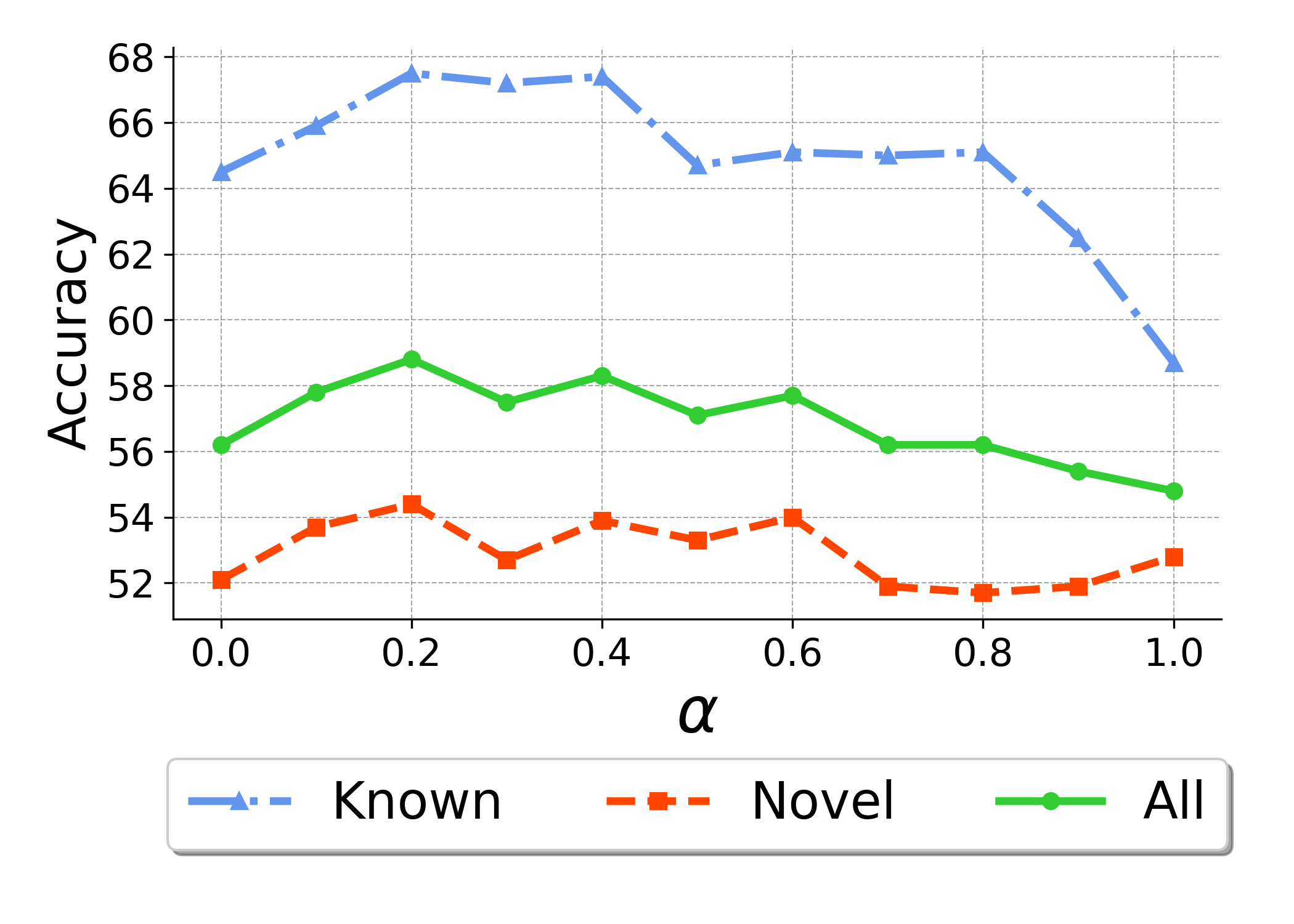}
  }
  \caption{Aircraft}
  \label{fig:figalpha2}
\end{subfigure}
\begin{subfigure}{.5\textwidth}
  \centering
\resizebox{\textwidth}{!}{
   \centering
  \includegraphics[width=0.9\linewidth]{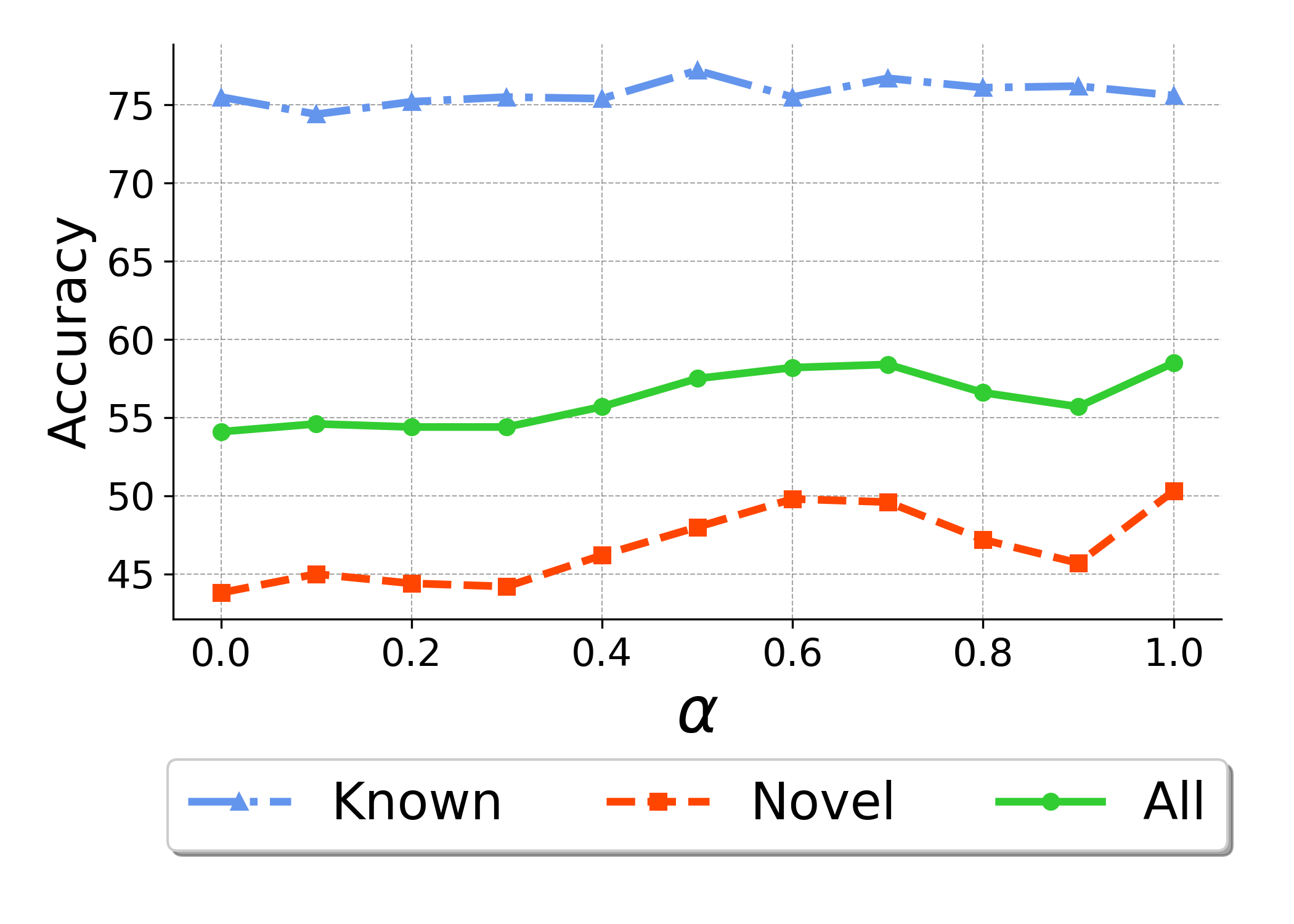}
}\caption{Standford Cars}
  \label{fig:figalpha3}
\end{subfigure}
\vspace{-1.5em}
\caption{\textbf{Impact of Label Smoothing Hyperparameter $\alpha$ Across Various Datasets.} This figure illustrates the differential effects of label smoothing, with increased smoothing typically enhancing performance on novel categories but impeding it on known ones. Furthermore, the benefits of more extensive smoothing are more pronounced for fine-grained datasets such as CUB, in contrast to Stanford Cars.}
\label{fig:alphaall}
\end{figure*}

\noindent\textbf{Effects of smoothing hyperparameter} 
In this section, paralleling our main discussion, we delve into the impacts of the hyperparameter $\alpha$ across various datasets, building upon our initial introduction of $\alpha$ within the context of unsupervised self-expertise. To recap, $\alpha$ serves as a smoothing hyperparameter that adjusts the uncertainty threshold for excluding negative samples from clusters other than the one under consideration. To elucidate, with $\alpha{=}1$, our model limits its incorporation of negative samples to those originating within its own cluster. On the flip side, an $\alpha{=}0$ setting treats all negative samples uniformly, resonating with the principles of conventional unsupervised contrastive learning approaches. Through a series of experiments, whose $\alpha$ values are meticulously outlined in \cref{fig:alphaall}, we observed a notable uplift in model efficacy on unseen categories as $\alpha$ escalated. This boost is chiefly ascribed to the inherent limitation of traditional unsupervised contrastive learning methods, which indiscriminately push apart all non-identical samples, neglecting their potential semantic proximities. Our graphical representations underscore a general trend: as $\alpha$ ascends, the model's performance on novel categories improves, albeit at the expense of its effectiveness on familiar categories. Noteworthy is the pattern observed across datasets, except for Pets, where an initial performance increment for known categories is visible until reaching a saturation point. Beyond this plateau, the model's performance either stabilizes—as observed in the CUB and Stanford Cars datasets—or begins to wane. This trend underscores the heightened efficacy of $\alpha$ in amplifying performance for datasets with finer granularity, such as CUB, thereby suggesting the aptness of unsupervised self-expertise for more fine-grained datasets.

\begin{figure*}[t!]
  \centering
  \resizebox{0.6\textwidth}{!}{
   \centering
  \includegraphics[width=\linewidth]{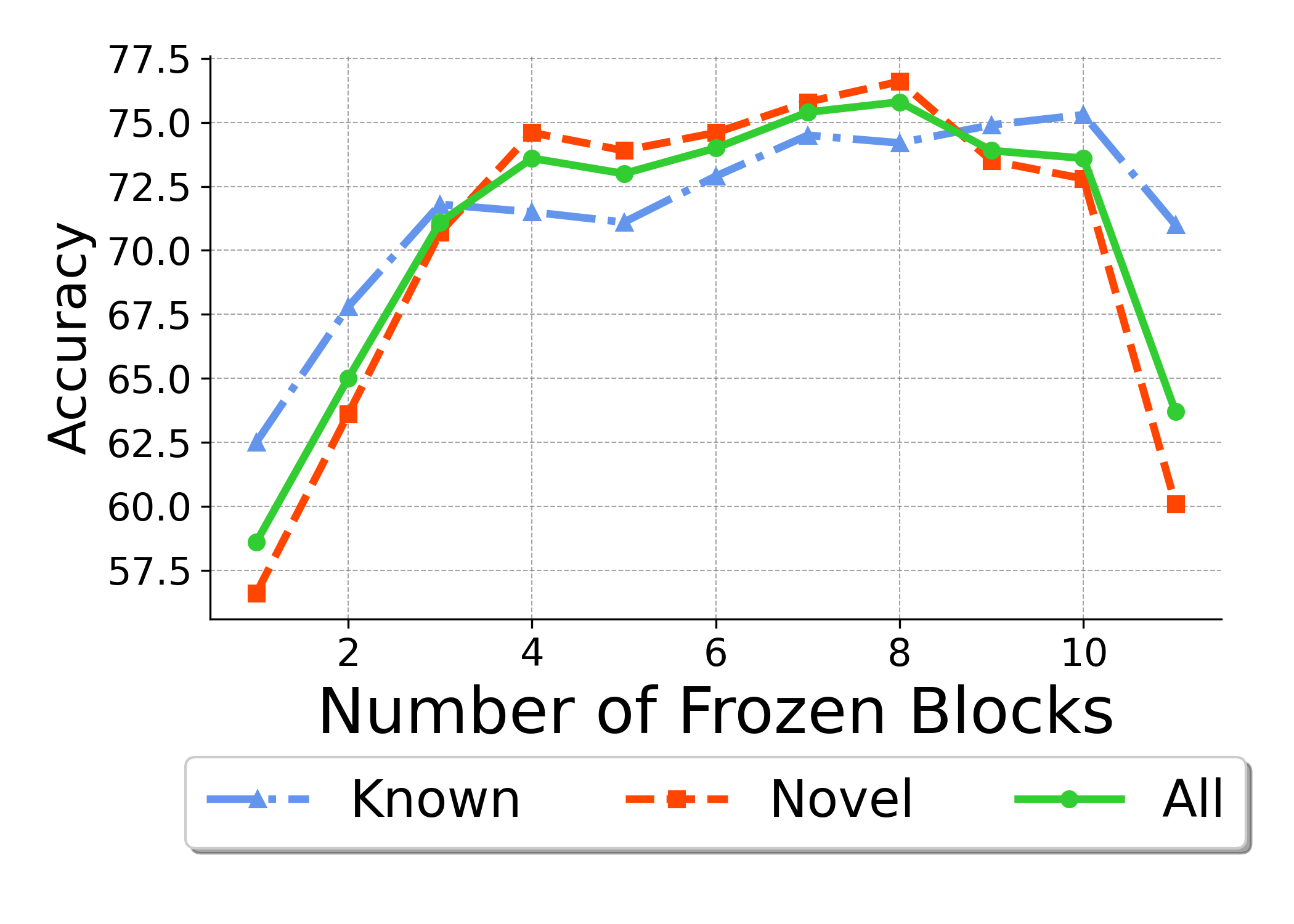}
  }
\vspace{-1.5em}
\caption{\textbf{Performance impact of varying frozen block counts on model adaptability.} This figure demonstrates the model's enhanced ability to learn novel categories with fewer frozen blocks (optimal at 8 frozen blocks) and improved recognition of known categories with increased frozen blocks (optimal at 10 frozen blocks), highlighting the balance between adaptability and stability derived from the pretrained DINOv1.}
\label{fig:frozen}
\end{figure*}

\noindent\textbf{Effects of number of frozen blocks} 
In this section, we delve into the impact of varying the quantity of frozen blocks on our model's efficacy. It's important to underline that an increase in the number of frozen blocks translates to a lesser deviation in the model's weights from those of the pre-trained architecture. Specifically, a scenario with zero frozen blocks indicates complete parameter adaptability to the target dataset, whereas a setup with twelve frozen blocks signifies no alteration from the initial DINOv1 pre-trained weights. The outcomes, as illustrated in Figure \cref{fig:frozen}, demonstrate a superior model performance with enhanced adaptability, particularly for recognizing novel categories. In contrast, for known categories, our findings reveal a roughly ascending trend, suggesting that minimal fine-tuning yields better results for these categories. Based on these observations, freezing eight blocks offers optimal results for novel categories, while, as we have used in the main paper, freezing ten blocks is more advantageous for known categories.





\begin{table}[ht!]
  \caption{\textbf{Comparison with state-of-the-art with Estimated Number of Categories} Bold numbers show the best accuracies. We leverage the estimation technique from Vaze \etal \cite{vaze2022generalized} to determine category counts—231 for CUB and 230 for Stanford Cars. It's noted that the category estimations by DCCL \cite{pu2023dynamic} and GCA \cite{otholt2024guided} diverge from other methods. Our approach outperforms other methods in all test scenarios (\textit{All}, \textit{Known}, \textit{Novel}), demonstrating robustness even with indeterminate category numbers.}
  \label{tab:est}
  \centering
  \resizebox{0.85\linewidth}{!}{
\begin{tabular}{c|l|c|ccc|ccc|ccc}
\toprule
&&& \multicolumn{3}{c}{\textbf{CUB-200}} & \multicolumn{3}{c}{\textbf{Stanford-Cars}}& \multicolumn{3}{c}{\textbf{Average}}\\ \cmidrule(lr){4-6} \cmidrule(lr){7-9}\cmidrule(lr){10-12}
&\textbf{Method} &\#Classes & All & Known  & Novel &  All & Known  & Novel&All & Known  & Novel\\
\midrule
\multirow{6}{*}{\rotatebox{90}{DINOv1}}&GCD \cite{vaze2022generalized} &Known& 51.3&	56.6&	48.7&39.0&57.6&29.9&45.2&57.1&39.3\\

&DCCL \cite{pu2023dynamic}        &Known&63.5&60.8&64.9&43.1&55.7&36.2&53.3&58.3&50.6\\

&SimGCD \cite{wen2022simple}       &Known&60.3&65.6&57.7	&53.8&71.9&45.0 &57.1&68.8&51.4\\

&GCA~\cite{otholt2024guided}&Known&\underline{68.8}&\underline{73.4}&\underline{66.6}&54.4&\underline{72.1}&45.8&\underline{61.6}&\underline{72.8}&56.2\\

&$\mu$GCD~\cite{vaze2023clevr4}&Known&65.7&68.0&64.6&\underline{56.5}&68.1 &\bf{50.9} &61.1&68.1&\underline{57.8}\\
\midrule
\rowcolor{gray!25}&\textbf{SelEx (Ours)} &Known&\bf{73.6} &\bf{75.3} &\bf{72.8}& \bf{58.5} &\bf{75.6}&\underline{50.3}&\bf{66.1}&\bf{75.5}&\bf{61.6}\\
\midrule
\multirow{6}{*}{\rotatebox{90}{DINOv1}}&GCD \cite{vaze2022generalized} &Estimated& 47.1&55.1&44.8& 35.0 & 56.0  &24.8&41.1&55.6&34.8\\

&DCCL \cite{pu2023dynamic}        &Estimated&\underline{63.5}&60.8&\underline{64.9}&43.1&55.7&36.2&53.3&58.3&50.6\\

&SimGCD \cite{wen2022simple}       & Estimated&61.5&\underline{66.4}&59.1 &49.1 & 65.1& 41.3&55.3&65.8&50.2\\

&GCA~\cite{otholt2024guided}&Estimated&62.0&65.2&60.4&54.5&\underline{71.2}&46.5&58.3&\underline{68.2}&53.5\\

&$\mu$GCD~\cite{vaze2023clevr4}&Estimated&62.0&60.3&62.8&\underline{56.3}&66.8&\bf{51.1}&\underline{59.2}&63.6&\underline{57.0}\\
\midrule
\rowcolor{gray!25}&\textbf{SelEx (Ours)} &        Estimated&\bf{72.0}& \bf{72.3} &\bf{71.9}&\bf{58.7}&\bf{75.3} &\underline{50.8}
&\bf{65.4}&\bf{73.8}&\bf{61.4}\\
\bottomrule
\end{tabular}
}
\vspace{0em}
\end{table}
\noindent\textbf{Estimate the number of categories} 
In the main text, we operated under the premise that the number of categories was predetermined. This assumption may not hold in practical scenarios. Therefore, in this segment, we examine the efficacy of our proposed SelEx model when the category count is deduced using a readily available technique, and we compare our results with several existing methods. For the estimation of category counts, we employ the method proposed by Vaze et al. \cite{vaze2022generalized}, yielding 231 categories for the CUB dataset and 230 for the Stanford Cars dataset. It is important to highlight that the category count estimations provided by DCCL \cite{pu2023dynamic} and GCA \cite{otholt2024guided} deviate from those obtained by other methodologies. As depicted in Table \ref{tab:est}, our SelEx model surpasses competing approaches across all evaluation settings (\textit{All}, \textit{Known}, \textit{Novel}), showcasing its adaptability in scenarios with uncertain category counts. This adaptability is further underscored by the fact that our model's performance improves, particularly for novel categories within the Stanford Cars dataset. This improvement can be attributed to the model's hierarchical nature, which does not strictly depend on an exact category count. Specifically, the model likely allocates additional categories to novel classes, given the higher estimated count of novel categories. This allocation effectively mitigates the general tendency of Generalized Category Discovery (GCD) methods to favor known categories, thereby enhancing performance for novel classes.

\noindent \textbf{Time-Complexity Analysis.} 
The time-complexity of our method is influenced by the variant of K-means in our semi-supervised approach. 
With GPU-based K-means, which loads the entire dataset's embedding in memory, the complexity is $\mathcal{O}(TK)$, where $K$ is the number of categories, and $T$ is the iteration count.
Practical timing of the GPU variant is provided in \cref{tab:time}.
\begin{table}[h!]
\caption{Time and Computational complexity of SelEx}
\vspace{0em}
\centering
\resizebox{0.85\linewidth}{!}{%
\setlength{\tabcolsep}{4pt}{\small
\begin{tabular}{lcccccc}
\toprule
&\multicolumn{3}{c}{\textbf{Accuracy on CUB}}&\multicolumn{3}{c}{\textbf{Estimated Performance}}\\ 
\cmidrule(lr){2-4} \cmidrule(lr){5-7} 
\textbf{Method}&All&Known&Novel &Epoch time & FLOPs & Params \\
\midrule
GCD (Baseline)&51.3&56.6&48.7   & 61.57 s &19.98 G &37.25 M\\
SelEx &73.6&75.3&72.8  &84.29 s  &40.43 G&37.25 M\\
\bottomrule
\end{tabular}
}
}
\vspace{0em}

\label{tab:time}
\vspace{-0.8em}
\end{table}

\begin{figure}[t]
  \centering
  \includegraphics[width=\linewidth]{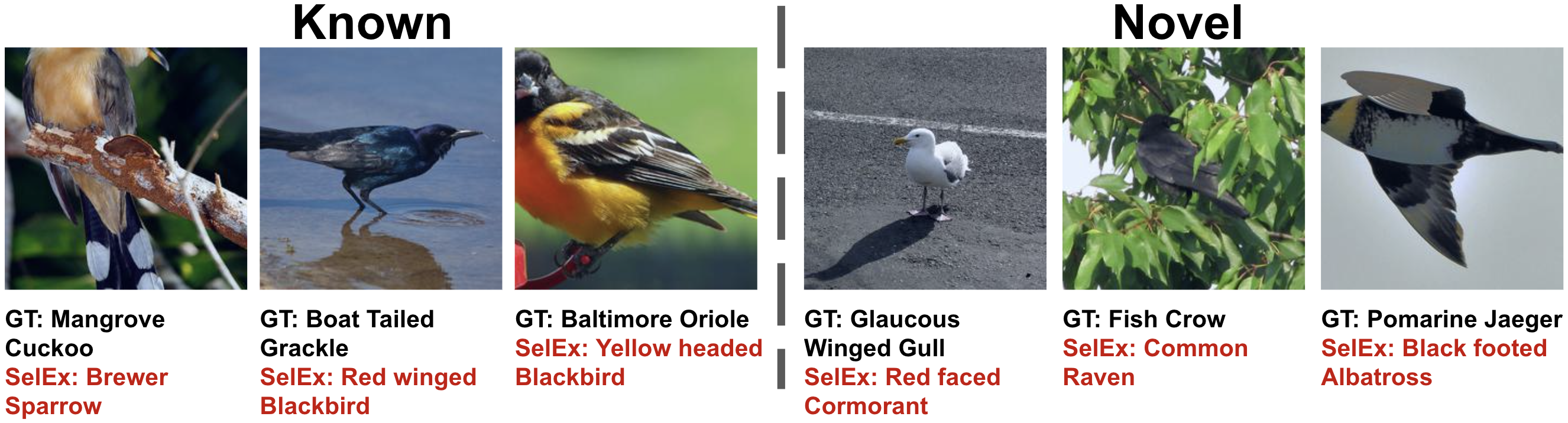}
  \vspace{-2em}
   \caption{Failure cases of our approach}
   \label{fig:fails}
   \vspace{-1.5em}
\end{figure}

\noindent \textbf{Failure Cases.}
\cref{fig:fails} highlights several instances where our model encounters challenges. Specifically, the model has difficulty distinguishing known categories when parts of the bird are occluded or reflected. For novel categories, failures typically arise due to occlusion or camouflage. These observations indicate potential areas for future work to enhance the robustness of our approach.

\vspace{-0.3em}
\section{Discussion}
\vspace{-0.8em}

\subsection{Limitations}
A significant constraint of methodologies grounded in hierarchical structures lies in their inherent assumption of dataset hierarchies. In scenarios where the dataset lacks clear hierarchical organization, this presupposition could be detrimental. While our method, SelEx, endeavors to mitigate this limitation by dynamically allocating portions of the latent dimension to represent higher hierarchical levels, it does not completely circumvent the issue. Furthermore, SelEx presupposes a balanced data distribution, a presumption that becomes particularly challenging in the context of datasets with a long-tailed distribution, such as the Herbarium dataset. This limitation highlights a critical area for further improvement and adaptation in our approach.
\subsection{Future Works}
Our work introduces SelEx as a novel approach for generalized category discovery, demonstrating promising results. The ablation studies reveal SelEx's potential applicability beyond its initial scope, particularly in few-shot and low-shot learning scenarios. Additionally, its hierarchical structure suggests utility in tasks with partially labeled data, offering a versatile framework for handling such complexities. Moreover, by incorporating label smoothing techniques to introduce uncertainty, SelEx shows resilience against noisy labels, enhancing its robustness and reliability in real-world applications. These insights indicate that SelEx possesses broad applicability across various domains within computer vision, especially in environments characterized by limited or sparse supervision. Future research will explore the extension of SelEx to a wider array of computer vision challenges, capitalizing on its adaptability and effectiveness in scenarios where supervision is minimal or data labels are imperfect.

%
%


\end{document}